\pdfoutput=1
\documentclass{article}

\ifdefined\ICLRNUMERIC
  \PassOptionsToPackage{numbers,sort&compress}{natbib}
\fi
\usepackage{iclr2027_conference,times}
\iclrfinalcopy % non-anonymous version (arXiv)
\ifdefined\ICLRNUMERIC
  \setcitestyle{numbers,square,sort&compress}
\fi

\usepackage[utf8]{inputenc}
\usepackage[T1]{fontenc}
\usepackage{hyperref}
\usepackage{url}
\usepackage{booktabs}
\usepackage{array}
\usepackage{multirow}
\usepackage{pifont}
\usepackage{makecell}
\usepackage{adjustbox}
\newcommand{\cmark}{\textcolor{green!45!black}{\ding{51}}}
\newcommand{\xmark}{\textcolor{red!70!black}{\ding{55}}}
\usepackage{amsmath}
\usepackage{amsfonts}
\usepackage{graphicx}
\usepackage{microtype}
\usepackage{xcolor}
\usepackage{colortbl}
\usepackage{enumitem}
\usepackage{capt-of}
\usepackage{needspace}
\usepackage{tikz}
\usepackage{tcolorbox}
\tcbuselibrary{breakable,skins}
\hypersetup{colorlinks=true,linkcolor=black,citecolor=blue!55!black,urlcolor=blue!55!black}

\definecolor{promptaccent}{RGB}{52,73,94}
\definecolor{promptborder}{RGB}{202,209,216}
\definecolor{promptbackground}{RGB}{248,249,251}
\definecolor{promptheader}{RGB}{237,241,245}
\newcommand{\rb}{\textsc{ReactionBench}}

\newtcolorbox[auto counter,number within=section]{promptbox}[2][]{
  enhanced jigsaw,
  breakable,
  width=\linewidth,
  colback=promptbackground,
  colframe=promptborder,
  colbacktitle=promptheader,
  coltitle=promptaccent,
  fonttitle=\footnotesize\sffamily\bfseries,
  fontupper=\small,
  title={\MakeUppercase{Prompt}~\thetcbcounter\hfill #2},
  title after break={\MakeUppercase{Prompt}~\thetcbcounter\ (continued)\hfill #2},
  label={#1},
  boxrule=0.45pt,
  leftrule=1.8pt,
  arc=1mm,
  outer arc=1mm,
  boxsep=0pt,
  left=2.4mm,
  right=2.4mm,
  top=1.7mm,
  bottom=1.7mm,
  toptitle=1mm,
  bottomtitle=1mm,
  before skip=8pt plus 2pt,
  after skip=8pt plus 2pt,
  before upper={\setlength{\parindent}{0pt}\setlength{\parskip}{0.55\baselineskip}}
}

\title{ReactionBench: \\A Human-Anchored Benchmark for Modeling Reactions to Video}
\author{%
Yibo Zhao$^{1,2,*}$\quad Ao Qu$^{2,*}$\quad Xuan Jiang$^{2}$\quad Jiahao Xu$^{3}$\quad Keane Ong$^{4}$\\
{\bfseries Hang Jiang$^{5}$\quad Zhaofeng Wu$^{2}$\quad Dingyi Zhuang$^{2}$\quad Yihong Tang$^{6}$}\\
{\bfseries Kaichen Zhou$^{2}$\quad Jinhua Zhao$^{2}$\quad Paul Pu Liang$^{2}$}\\[4pt]
{\normalfont\footnotesize $^{1}$University of Illinois Urbana-Champaign\quad $^{2}$Massachusetts Institute of Technology}\\
{\normalfont\footnotesize $^{3}$Independent Researcher\quad $^{4}$National University of Singapore}\\
{\normalfont\footnotesize $^{5}$Northeastern University\quad $^{6}$McGill University}\\[2pt]
{\normalfont\footnotesize $^{*}$Equal contribution.\quad \texttt{\{yibozhao,qua\}@mit.edu}}
}

\begin{document}
\maketitle
% \iclrfinalcopy prints "Published as a conference paper at ICLR 2027"; this version is a preprint.
\lhead{Preprint}

\begin{abstract}
Understanding what happens in a video is not sufficient for socially responsive multimodal systems: such systems also aim to relate what a person observes to how that person responds. Existing video benchmarks largely evaluate events, temporal reasoning, or people depicted in the video, while affective and viewer-response datasets often focus on emotion labels, aggregate responses, or prediction from the stimulus alone. This leaves an underexplored question: can current multimodal models ground an observed viewer reaction in the stimulus context associated with it? We introduce \rb{}, a human-anchored benchmark for reaction-specific video understanding. Built from a candidate pool of 31,834 synchronized stimulus–reaction clips spanning 433 hours and eight domains, \rb{} evaluates three complementary abilities: ranking the intensity of a viewer’s reactions within a video, matching a reaction to its temporally aligned stimulus, and explaining the concrete stimulus cues linked to that reaction. Human annotation provides 14,894 ratings over 3,257 moments, and the frozen evaluation comprises 250 intensity profiles, 112 matching items, and 200 rationale items with 447 blind-written human references.
Across seven multimodal models, reaction-specific grounding remains challenging: the best model reaches 50.9\% accuracy on four-way stimulus–reaction matching, while GPT-5 shows no detectable loss under reaction substitution. Performance is also sensitive to evaluation protocol.
\\[3pt]
\textbf{Project page:} \url{https://yibozh.github.io/ReactionBench-page/}

\end{abstract}

\section{Introduction}

Socially responsive multimodal systems aim to connect video events to how viewers respond to them. A laugh follows a visual gag; a pause is meaningful only against the same viewer’s surrounding behavior. Evaluating this relation requires paired stimulus–reaction evidence and tasks that test reaction strength, stimulus correspondence, and the cues that explain the response.

Existing benchmarks cover parts of this problem. General video benchmarks prioritize event, temporal, or knowledge understanding~\citep{li2024mvbench,fu2025videomme,mangalam2023egoschema,he2024mmworld}, while human-centered benchmarks commonly target actions, affect categories, or mental-state questions~\citep{gu2018ava,jiang2020dfew,zadeh2019socialiq,jin2024mmtomqa}. Reaction datasets, in turn, often provide limited stimulus diversity, coarse labels, or no evaluation protocol for current multimodal models~\citep{sun2021eev,li2023reactionet,jia2024inducedsentiment}. These resources either study people depicted in a video, predict or aggregate viewer responses, or model isolated affective signals; none jointly evaluates whether a model can rank, temporally align, and explain an observed viewer reaction against its associated stimulus.

\begin{figure*}[t]
  \centering
  \includegraphics[width=\textwidth]{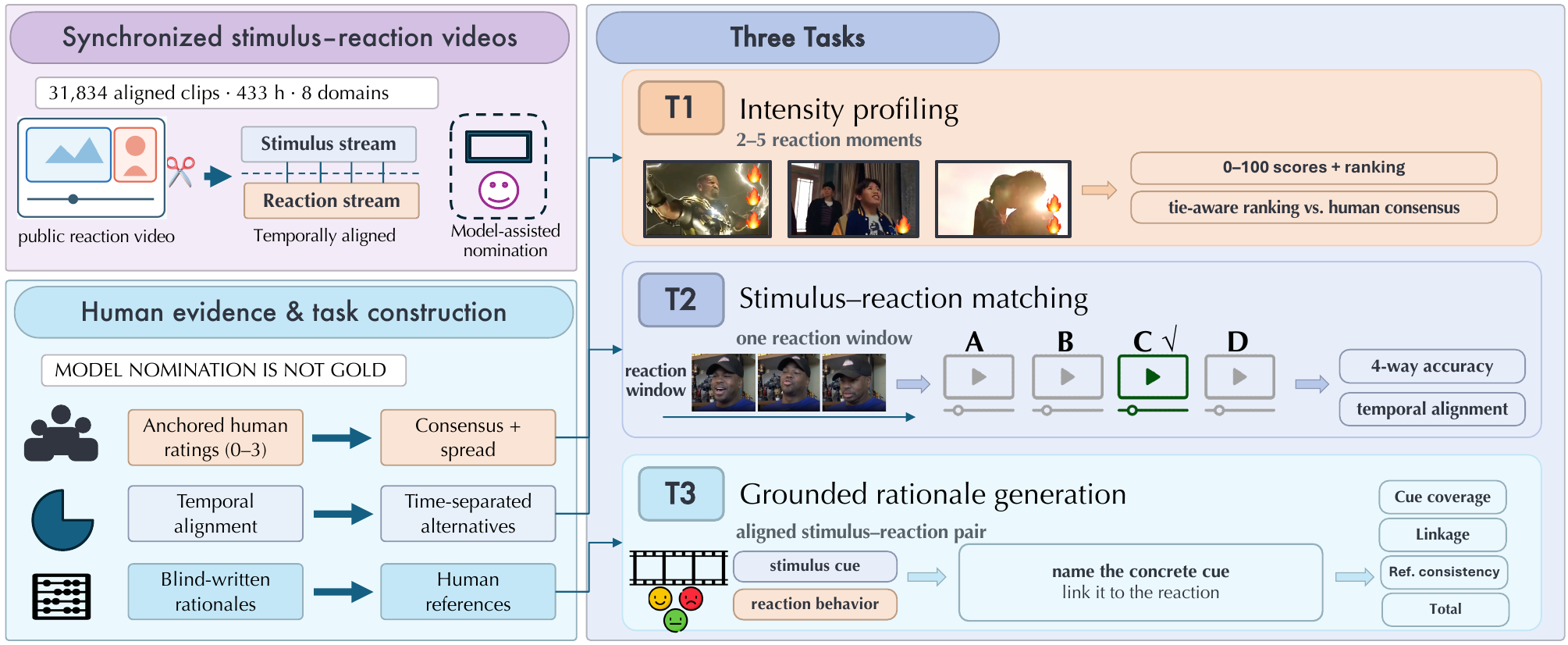}
  \caption{\textbf{Overview of \rb.} Built from synchronized in-the-wild stimulus--reaction videos, \rb{} evaluates reaction-specific video understanding: whether multimodal models can connect a stimulus to an observed viewer reaction. Three tasks probe this relation: scoring the intensity of moments within one reaction video (IP), matching a reaction to its aligned stimulus (SRM), and explaining a reaction from the concrete stimulus cue (GRG). Intensity scores and rationale references are human-authored; SRM answers come from temporal alignment. Ref. consistency checks for contradictions with rationale references.}
  \label{fig:overview}
\end{figure*}

We present \textbf{\rb}, built from synchronized in-the-wild stimulus--reaction videos and organized around three tasks (Figure~\ref{fig:overview}): \textbf{Intensity Profiling (IP)} scores marked moments from one reaction video; \textbf{Stimulus--Reaction Matching (SRM)} selects the temporally aligned stimulus from four same-video candidates; and \textbf{Grounded Rationale Generation (GRG)} produces a cue-based explanation scored against blind-written human references. Model-assisted nomination selects candidates for annotation; task labels come from human ratings, human-written references, and temporal alignment.

Our study asks two questions: how well current multimodal models understand and ground observed viewer reactions across the three tasks (RQ2), and which annotation, input, and supervision choices affect how that capability is measured (RQ1, RQ3, RQ4).
% We ask how well current multimodal models ground observed viewer reactions (RQ2), and how annotation, input, and supervision choices affect their measurement (RQ1, RQ3, RQ4).

\textbf{Contributions.}\quad \textbf{(i) \rb{}:} a human-anchored benchmark for grounding an observed viewer reaction in its stimulus, over three tasks that each state their input, supervision source, and scoring rule: 250 videos with 642 intensity moments, 112 same-video matching items, and 200 rationale items carrying 447 human references. \textbf{(ii) A seven-model evaluation} showing substantial headroom in reaction understanding: stimulus--reaction matching reaches 50.9\% at best, while GPT-5 controls show no detectable loss under reaction substitution. \textbf{(iii) Diagnostic analyses} of how scoring organization and supervision affect measured performance, and of which reaction evidence models use.

\section{Related Work}
\label{sec:related}

\begin{table*}[t]
\centering
\caption{\textbf{Comparison between \rb{} and existing video benchmarks and datasets.} \emph{Obs.\ response} is the recorded human response signal; \emph{Int.}/\emph{Align.}/\emph{Rat.} denote within-episode intensity ranking, alignment of a reaction to its stimulus, and grounded rationale. Modalities: T text, A audio, V video, P physiological. Counts are those stated in each resource's own paper; \mbox{---} marks an unstated quantity ($^{s}$~stimulus videos, from 27 participants and 532 reaction trials). A starred mark is a related but distinct ability: EEV and MuSe-Reaction regress reaction strength rather than ranking human-anchored moments, HumanVBench compares intensity across individuals, StimuVAR selects emotion-triggered frames, and DAR segments affect transitions along the stimulus timeline without an observed reaction. \rb{}'s \#Clips is the source collection behind its frozen evaluation sets; its code, task files, and annotations are public (\emph{annot.}), and its video is referenced by key rather than redistributed.}
\label{tab:related}
\setlength{\tabcolsep}{5.5pt}
\scriptsize
\begin{adjustbox}{width=\textwidth}
\begin{tabular}{@{}l l c c c c c l l c c c@{}}
\toprule
\multirow{2}{*}{\textbf{Type}} & \multirow{2}{*}{\textbf{Dataset / Benchmark}}
& \multicolumn{4}{c}{\textbf{Data}}
& \multicolumn{2}{c}{\textbf{Benchmark}}
& \multicolumn{4}{c}{\textbf{Reaction modeling}} \\
\cmidrule(lr){3-6}\cmidrule(lr){7-8}\cmidrule(l){9-12}
& & \textbf{\#Clips} & \textbf{Mod.} & \textbf{Source} & \textbf{Open}
  & \textbf{\#Items} & \textbf{Format}
  & \textbf{Obs.\ response} & \textbf{Int.} & \textbf{Align.} & \textbf{Rat.} \\
\midrule
\multirow{5}{*}{\makecell[l]{General video\\understanding}}
& Video-MME~\citep{fu2025videomme}            & 900   & T,A,V & Web      & yes   & 2,700  & MCQ            & ---   & \xmark & \xmark & \xmark \\
& MVBench~\citep{li2024mvbench}               & ---   & T,V   & Existing & yes   & 4,000  & MCQ            & ---   & \xmark & \xmark & \xmark \\
& CG-Bench~\citep{chen2025cgbench}            & 1,219 & T,A,V & Web      & yes   & 12,129 & MCQ+OE+Grd.    & ---   & \xmark & \xmark & \xmark \\
& VRBench~\citep{yu2025vrbench}               & 960   & T,A,V & Web      & yes   & 8,243  & MCQ+OE         & ---   & \xmark & \xmark & \xmark \\
& SeriesBench~\citep{zhang2025seriesbench}    & 1,072 & T,A,V & Web      & yes   & 29,196 & MCQ+TF+OE      & ---   & \xmark & \xmark & \xmark \\
\midrule
\multirow{5}{*}{\makecell[l]{Human-centered\\and affective}}
& HumanVBench~\citep{zhou2026humanvbench}     & 2,182  & T,A,V & Web      & yes     & 2,475   & MCQ          & ---   & \xmark$^{*}$ & \xmark & \xmark \\
& MimeQA~\citep{li2025mimeqa}                 & 101    & T,V   & Web      & yes     & 806     & OE           & ---   & \xmark & \xmark & \xmark \\
& Social Genome~\citep{mathur2025socialgenome}& 272    & T,A,V & Existing & partial & 1,486   & Trace+Ans.   & ---   & \xmark & \xmark & \cmark \\
& HumanSense~\citep{qin2026humansense}        & ---    & T,A,V & Existing & no      & 3,882   & MCQ          & ---   & \xmark & \xmark & \xmark \\
& MME-Emotion~\citep{zhang2025mmeemotion}     & 6,500  & T,A,V & Existing & ---     & 6,500   & CLS+CoT      & ---   & \xmark & \xmark & \xmark \\
\midrule
\multirow{10}{*}{\makecell[l]{Viewer response\\to a stimulus}}
& MAHNOB-HCI~\citep{soleymani2012mahnob}      & 20$^{s}$ & A,V,P & In-person & gated   & 532     & ---          & Face, gaze, physio.   & \xmark & \xmark & \xmark \\
& EEV~\citep{sun2021eev}                      & 23,574 & A,V   & Web      & partial & 36.7\,M & Regression   & Facial expr.\ (auto)  & \cmark$^{*}$ & \xmark & \xmark \\
& ReactioNet~\citep{li2023reactionet}         & 2,486  & T,A,V & Web      & request & 50,000  & CLS          & Facial behavior       & \xmark & \xmark & \xmark \\
& MuSe-Reaction~\citep{amiriparian2022muse}   & ---    & A,V   & In-the-wild & gated & ---     & Regression   & Reaction intensity    & \cmark$^{*}$ & \xmark & \xmark \\
& Seamless Inter.~\citep{agrawal2025seamless} & 64,739 & T,A,V & In-person & yes     & 16,898  & ---          & Dyadic behavior       & \xmark & \xmark & \xmark \\
& Video2Reaction~\citep{nguyen2026video2reaction} & 10,348 & T,A,V & Existing & yes & 252,462 & LDL+CLS      & Comment distribution  & \xmark & \xmark & \xmark \\
& SMILE~\citep{hyun2024smile,lee2026smilenext}& 887    & T,A,V & Existing & yes     & 887     & OE           & Laughter              & \xmark & \xmark & \cmark \\
& VCE~\citep{mazeika2022vce}                  & 61,046 & V     & Web      & yes     & 61,046  & Distribution & Reported emotion      & \xmark & \xmark & \xmark \\
& StimuVAR~\citep{guo2025stimuvar}            & 61,046 & V     & Existing & partial & ---     & CLS+OE       & --- (predicted)       & \xmark & \xmark$^{*}$ & \cmark \\
& DAR~\citep{zhang2026dar}                    & 15,087 & V     & Existing & yes     & 36,908  & Seg.+CLS+OE  & Reported emotion      & \xmark & \xmark$^{*}$ & \cmark \\
\midrule
& \textbf{\rb{} (ours)} & \textbf{31,834} & T,A,V & Web & annot. & \textbf{954} & Rank+MCQ+OE & \textbf{Reaction video} & \cmark & \cmark & \cmark \\
\bottomrule
\end{tabular}
\end{adjustbox}
\end{table*}

\paragraph{General and Temporal Video Evaluation.}
Video benchmarks cover task taxonomies, durations, modalities, temporal skills, and causal chains~\citep{li2024mvbench,fu2025videomme,fu2026videommev2,wu2024longvideobench,chen2025cgbench,zhang2025seriesbench,hu2025videommmu,zhang2026castbench,xie2026maverix}, and controls show that temporal items can be answered without temporal evidence and that audio can supply shortcuts~\citep{cores2024tvbench,shangguan2025tomato,hong2026worldsense,yang2025avut}. 
% \rb{} applies input controls to paired stimuli and observed reactions.

\paragraph{Human-Centered and Social-Affective Video Understanding.}
Human-centric, theory-of-mind, and affective benchmarks probe perception, motion, nonverbal signals, goals, beliefs, and social causality~\citep{zhou2026humanvbench,qin2026humansense,hong2025motionbench,li2025mimeqa,ong2026behavioratlas,jin2024mmtomqa,shi2025mumat,wei2026moviegraphtom,villacueva2025moments}, combine emotion recognition with explanation or localization~\citep{cheng2024emotionllama,lian2024affectgpt,lin2024e3,hu2025emobenchm,zhang2025mmeemotion}, extract an emotion's trigger~\citep{wang2024semeval,lin2025whywefeel}, grade intensity~\citep{zhou2024memobench}, and ask when affect changes~\citep{hu2026emotrans}; Social Genome further requires models to cite supporting evidence~\citep{mathur2025socialgenome}, and egoEMOTION anchors affect physiologically~\citep{jammot2025egoemotion}. Such models often answer from priors rather than the video and localize poorly in time~\citep{xing2025emotionhallucer,han2025longinsightbench}. These benchmarks primarily study people depicted \emph{in} a video. \rb{} focuses on a viewer's response \emph{to} one.

\paragraph{Stimulus-Induced Response.}
EmotionBench compares model and human responses to text-described situations~\citep{huang2024emotionbench}, while induced-sentiment and audience-reaction prediction aggregate video-triggered text comments into a population-level distribution rather than one viewer's response over time~\citep{jia2024inducedsentiment,nguyen2026video2reaction}. Table~\ref{tab:related} places \rb{} among the resources that record a viewer response: affective-computing datasets capture synchronized signals but target affect recognition rather than model evaluation~\citep{soleymani2012mahnob,amiriparian2022muse,sun2021eev,li2023reactionet}; VCE records recruited viewers' reported emotion distributions for 61,046 videos~\citep{mazeika2022vce}; StimuVAR and GRACE evaluate models but predict from the stimulus alone~\citep{guo2025stimuvar,yang2026grace}; DAR adds event-aligned affect transitions and reasoning chains over VCE's reported viewer emotions rather than observed reactions~\citep{zhang2026dar}; laughter reasoning explains a reaction without ranking intensity or localizing a trigger~\citep{hyun2024smile,lee2026smilenext}; and highlight, engagement, and reaction-generation work scores saliency or realism, not interpretation~\citep{lei2021qvhighlights,sul2023mrhisum,chang2025aha,sun2025engagement,ng2023listen,yu2025hero,agrawal2025seamless,li2025vividlistener,zhou2026musteernet,song2026react}. Preference-level benchmarks approach our ranking protocol, and models reproduce human video ratings only weakly~\citep{lian2025emoprefer,chen2026multiemobench,shrestha2026synthetic}, but none evaluates whether models can rank, temporally align, and explain an observed reaction against human-anchored references.

\paragraph{Evaluating a Subjective Target.}
Because reaction intensity is a subjective judgment, the protocol that measures it needs justification. Most evaluations report neither uncertainty nor statistical tests~\citep{bean2025constructvalidity}, while WAGIBench and VideoAutoArena build evaluator meta-evaluation into the benchmark~\citep{veerabadran2025wagibench,luo2025videoautoarena}. Annotator disagreement can reflect perspective rather than noise~\citep{ivey2025nutmeg,xu2026beyondconsensus} and absolute thresholds suit perceptual judgments poorly~\citep{wong2021crossreplication}, so we report disagreement structure alongside the coefficient. LLM judges vary with model, protocol, position, and self-preference~\citep{gu2025judgesurvey,tan2025judgebench,salinas2025judgetuning,panickssery2024selfpreference}, diverge from humans most on absolute scoring~\citep{chen2024mllmasajudge}, are weaker without video~\citep{waheed2025videojudge}, and miss human uncertainty on affect~\citep{norman2026reliability,inoshita2026emotionuncertainty}; rubrics help~\citep{liu2023geval,zheng2023mtbench} but do not replace validating the judge on the target task~\citep{bavaresco2025llmsinstead,watts2024pariksha,calderon2025alttest}.

\section{The \texorpdfstring{\rb{}}{ReactionBench} Benchmark}
\label{sec:data}

\rb{} starts from 31,834 synchronized stimulus--reaction clips totaling 433:42:49 hours across games, music, sports, live streams, movies and television, expert commentary, pop culture, and comedy. These clips form the candidate pool for three evaluation sets constructed from human ratings, temporal alignment, and blind-written rationales.

\subsection{Task Design and Definition}
\label{sec:tasks}

The three tasks use synchronized stimulus--reaction evidence with distinct inputs, outputs, and supervision sources.

\paragraph{Intensity Profiling (IP).}
For a reaction video $V$, let $\mathcal{T}_V=\{[t_{\mathrm{start},i},t_{\mathrm{end},i}]\}_{i=1}^{m}$ denote its $m\in\{2,\ldots,5\}$ marked time ranges, and let $F_1(V)$ denote the reaction frames sampled at 1~FPS. In the joint setting, the model receives $F_1(V)$ together with all marked ranges and returns a score vector $\hat{\mathbf y}_V$. We define the prediction and video-level score as
\begin{equation}
\begin{aligned}
\hat{\mathbf y}_V
  &= f_{\mathrm{IP}}\!\left(F_1(V),\mathcal{T}_V\right),
  \quad \hat y_i\in\{0,\ldots,100\},\\[-2pt]
\rho_0(V)
  &=
  \begin{cases}
    0, & \hat y_1=\cdots=\hat y_m,\\
    \rho_{\mathrm S}(\hat{\mathbf y}_V,\mathbf y_V), & \text{otherwise},
  \end{cases}
\end{aligned}
\label{eq:ip}
\end{equation}
where $\mathbf y_V$ contains the full human-consensus scores and $\rho_{\mathrm S}$ is standard Spearman correlation. The extension $\rho_0$ assigns zero to parsed constant predictions, for which standard Spearman is undefined. We average $\rho_0$ equally over videos with parsed outputs. Missing outputs are excluded and counted separately. Standard Spearman after excluding constants is a conditional sensitivity analysis (Appendix~\ref{app:robustness}).

We evaluate IP under two settings: \textbf{(a) Joint:} the model receives the full sampled video and scores all $m$ marked moments in one call; and \textbf{(b) Independent:} the model receives only the frames within one marked time range and returns one 0--100 score, with this process repeated for all $m$ ranges. Both settings use the same marked moments and scoring scale. Because they differ in both available video context and output format, their comparison measures sensitivity to the evaluation setting rather than the causal effect of context alone.

\paragraph{Stimulus--Reaction Matching (SRM).}
SRM presents the reaction and four same-video stimulus candidates in one call and requires A, B, C, or D. For item $j$, let $R_j$ be the reaction window, $\mathbf S_j=(S_{j,1},\ldots,S_{j,4})$ the candidates, and $c_j$ the temporally aligned candidate. With sampled frames,
\begin{equation}
\hat c_j=f_{\mathrm{SRM}}\!\left(F_1(R_j),F_1(\mathbf S_j)\right),\qquad
\mathrm{Acc}_{\mathrm{SRM}}=\frac{1}{N}\sum_{j=1}^{N}\mathbf{1}[\hat c_j=c_j].
\label{eq:srm}
\end{equation}
A--D map to 1--4. Error distance describes the start-time separation of an incorrect selection from the aligned window. The correct option is fixed by the synchronized recording: it is the stimulus window on screen during the reaction, so SRM labels are ground truth by construction and require no annotation.

\paragraph{Grounded Rationale Generation (GRG).}
GRG has two stages: \textbf{(a) Generation:} a model receives sampled frames from the aligned stimulus and reaction windows and writes one cue-based explanation, without seeing pre-screen text or human references. \textbf{(b) Judging:} a text-only judge compares it with the accepted human explanations. Cue and link scores are in $\{0,1,2\}$ and reference consistency is in $\{0,1\}$; their sum is the 0--5 composite. The executable rubric calls the last field \emph{faithful}; it tests contradiction with references, rather than direct factual fidelity to the video.

\subsection{Human-Anchored Annotation Pipeline}
\label{sec:pipeline}

\begin{figure*}[t]
  \centering
  \includegraphics[width=\textwidth]{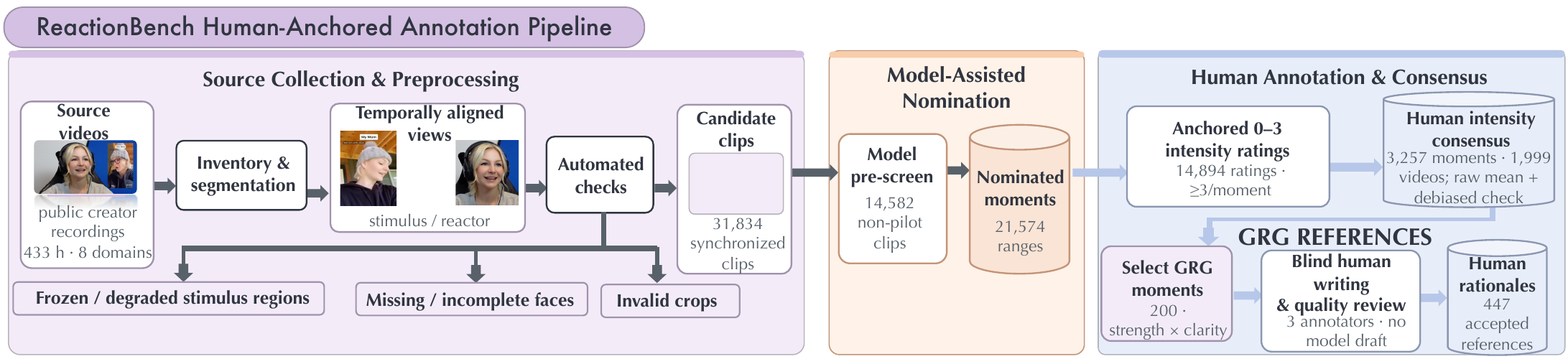}\\[1pt]
  {\small (a)}\\[6pt]
  \includegraphics[width=\textwidth]{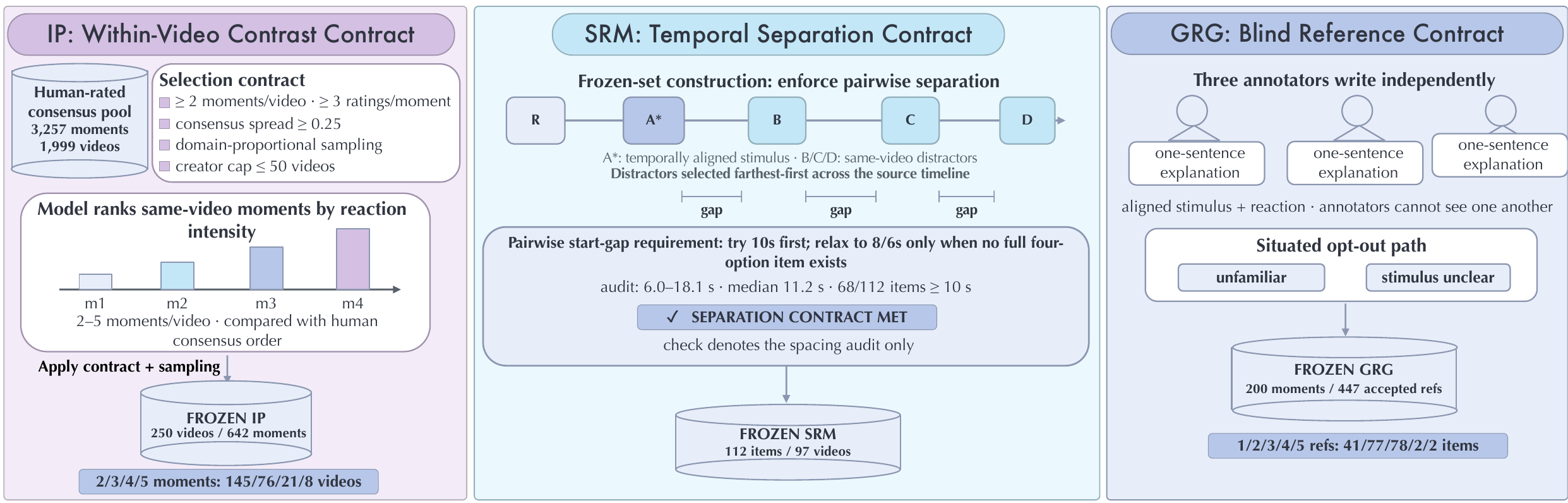}\\[1pt]
  {\small (b)}
  \caption{\textbf{From nominated moments to frozen evaluation sets.} \textbf{(a)} The \rb{} annotation pipeline: automated processing and model-based nomination reduce annotation load, while people provide intensity ratings and rationale references. \textbf{(b)} Construction of the three evaluation sets, each under an explicit contract: within-video contrast for IP, temporal separation between candidate windows for SRM, and blind-written human references for GRG.}
  \label{fig:pipeline}
  \label{fig:construction}
\end{figure*}

\paragraph{Source Collection and Preprocessing.}
The source videos are real-world reaction recordings collected from public online creator channels. Each recording contains the viewed content and the creator's response, commonly in picture-in-picture layouts. The preprocessing pipeline inventories videos and metadata, segments usable viewing intervals, and extracts temporally aligned stimulus and reactor views. Automated checks target frozen or degraded stimulus regions, missing or incomplete faces, and invalid crops. These operations produce candidate clips for annotation, as detailed in Appendix~\ref{app:source}.

\paragraph{Model-Assisted Nomination.}
A model-based pre-screen nominates reaction moments for human annotation. In a blind 50-clip pilot, a human reviewer scored nominations from both candidate selectors using the same behavioral intensity rubric as the human annotation. Among nominations from the selected Gemini 3.6 Flash configuration, 92.0\% received a human score of at least 2, compared with 68.0\% for a LibreFace-based facial-analysis selector~\citep{chang2024libreface}. Gemini 3.6 Flash was then applied to 14,582 non-pilot clips, producing 21,574 nominations. The exact prompt and output schema appear in Appendix~\ref{app:nomination}. Model scores, ranks, and explanations are used only for nomination and excluded from evaluation prompts and human consensus. Excluding these outputs prevents direct label leakage but does not remove selection effects or possible selector-family dependence: the pre-screen favors strong visible or audible responses, so the evaluated sets represent selected creator reactions rather than ordinary viewing behavior or reactions the selector missed.

\paragraph{Human Intensity Annotation.}
Human raters scored nominated moments on a behaviorally anchored 0--3 ordinal scale (Appendix~\ref{app:intensity-annotation}). The annotation pool contains 14,894 ratings on 3,257 moments from 1,999 videos. Every moment has at least three ratings, from eight raters in total. We retain all ratings and compute both raw and rater-debiased consensus using a two-way additive model. In the 250-video ranking set, debiasing changes the within-video order in 0 of 250 videos and leaves every model correlation unchanged to four decimals. We therefore report raw-mean results for readability and treat debiasing as a robustness check.

\subsection{Evaluation-Set Construction}
\label{sec:construction}

\paragraph{IP Selection.}
IP includes 250 videos with 642 moments, with 145/76/21/8 videos containing 2/3/4/5 moments. Eligibility requires at least two rated moments and consensus spread of at least 0.25. The set contains seven creators from five channels, with a cap of 50 videos per creator; 200 videos come from a source category labeled Patreon. Patreon identifies an acquisition source; the final set has a narrower distribution than the eight-domain collection (Appendix~\ref{app:annotation}).

\paragraph{SRM Construction.}
SRM uses 112 items from 97 clips, with seven creator personas from five channels. Each reaction is paired with its aligned stimulus window and three same-video alternatives, with audited start-time separation (Appendix~\ref{app:srm-construction}). A manual review confirmed that the four candidates of every item are visually distinguishable. In a Gemini 3.7 Flash native-video control with audio, replacing same-video alternatives with other-video windows increases accuracy by $+0.232$ (McNemar $p=6\times10^{-6}$); this accuracy increase is consistent with an audio rejection cue, without isolating audio from the other changes in distractor source.

\paragraph{GRG References.}
GRG contains 200 moments from 185 clips, with eight creator personas from seven channels, selected for visible reaction strength and stimulus clarity (Table~\ref{tab:eval-inventory}). 
% Three completed annotators watched the stimulus with audio and a muted reaction view, then wrote one sentence from a blank text box. Paid opt-outs recorded unfamiliarity or unclear stimuli. Quality review retained 447 references, with 1/2/3/4/5 references for 41/77/78/2/2 items. These references can depend on speech or sound unavailable to frame-only generators, so recovery scores reflect both input observability and inference.
Three annotators who completed all 200 items watched the stimulus with audio and a muted reaction view, then wrote one sentence from a blank text box. Annotators could mark unfamiliar or unclear stimuli without losing payment. Quality review retained 447 references, with 1/2/3/4/5 references for 41/77/78/2/2 items. These references can depend on speech or sound unavailable to frame-only generators, so GRG scores reflect both input observability and inference.

\section{Evaluation}
\label{sec:experiments}

This section asks whether models ground an observed reaction in its stimulus, and how measured performance changes with scoring organization and supervision.

\subsection{Experimental Settings}
\label{sec:settings}

\textbf{Models, inputs, and statistical protocol.}\quad We evaluate GPT-5~\citep{openai2025gpt5}, Gemini 3.7 Flash~\citep{googledeepmind2026gemini37flash,geminiteam2025gemini25}, Qwen3.5-9B and Qwen3.5-397B-A17B~\citep{qwen2026qwen35,qwen2026qwen35omni}, Qwen3.6-27B~\citep{qwen2026qwen3627b}, Qwen3.6-35B-A3B~\citep{qwen2026qwen3635ba3b}, and Kimi-K2.6~\citep{moonshot2026kimik26,kimiteam2026k25}. The common input uses 1~FPS extraction, a maximum side of 512 pixels, and frame caps of 64 for joint IP, 12 reaction and 6 per option for SRM, and 12 per stream for GRG. Each reported inference setting uses one run, with deterministic decoding where supported. Accuracy uses Wilson intervals~\citep{wilson1927interval}; mean differences use paired bootstrap intervals over videos or, in follow-up analyses, groups of clips from the same source recording~\citep{efron1979bootstrap}, with the original Wilcoxon tests retained as secondary distributional checks~\citep{wilcoxon1945individual}. Intervals are pointwise and conditional on the fixed annotator pool. A Gemini native-video control includes audio and is not modality-matched. Appendix~\ref{app:timestamp-audit} reports a frame-time indexing audit and a corresponding sensitivity analysis.

\textbf{GRG judging and human comparison.}\quad DeepSeek-V3.1~\citep{deepseekai2024deepseekv3} supplies the shared text-only judge. Its lack of video limits direct factual verification~\citep{waheed2025videojudge}. Three Prolific raters graded 84 anonymized generator--item pairs sampled across seven generators and three judge-score strata, using the same rubric and visible references. Judge composite scores correlate with the human mean at Spearman $\rho=0.593$ (95\% bootstrap CI $[0.43,0.72]$). Mean judge--individual agreement is 0.447, close to the mean inter-rater agreement of 0.441, supporting item-level rubric association in this stratified sample; cross-judge sensitivity is reported in Appendix~\ref{app:robustness}.

\subsection{RQ1: What Intensity Reference Do the Human Annotations Support?}
\label{sec:reliability}

Ordinal Krippendorff's $\alpha=0.357$~\citep{krippendorff2004content,krippendorff2011computing} indicates substantial disagreement. Across 27,455 within-item rater pairs, 57.1\% receive exactly the same score; 25.3\% of items are unanimous across all raters. Of one-point disagreements, 69\% occur at 1--2, 24\% at 2--3, and 7\% at 0--1. These proportions describe the selected pool of strong reactions.

We preserve continuous consensus scores and rater spread. For human--model comparisons, both the held-out rater and model are scored against the same remaining-rater consensus. The primary statistic is pairwise ordering accuracy: consensus ties are omitted and prediction ties receive half credit; scores are averaged over pairs within each rater--video unit, then over raters within each video, then equally over videos. There are 956 eligible rater--video units over 250 videos, including 536 with constant held-out-rater profiles. The video-weighted individual-rater reference is 0.607. Conditional Spearman on jointly nonconstant profiles is reported with coverage in Appendix~\ref{app:human-matched}. This reference is agreement with a fixed rater pool rather than a human ceiling.

\subsection{RQ2: How Well Do Current Multimodal Models Understand and Ground Viewer Reactions?}
\label{sec:rq2}

GPT-5 shows sensitivity to the presence of a reaction, but little detectable sensitivity to which reaction it is shown; two paired input controls separately test reaction removal and reaction substitution. On SRM, 109 shared items score 0.422 with the normal input, 0.376 with the stimulus options alone, and 0.385 when another viewer's reaction replaces the aligned one; both paired intervals include zero (Appendix~\ref{app:reaction-controls}). On GRG, 197 common items score 2.690 out of 5 with the normal input, 1.959 without the reaction, and 2.746 with another viewer's reaction: removing the reaction changes the score (interval excludes zero), while substituting a different one does not (Appendix~\ref{app:grg-controls}).

Absolute performance also leaves substantial headroom. SRM accuracy spans 0.259--0.509 against uniform-choice chance of 0.250, with unparsed responses counted as incorrect. GRG composites span 2.610--2.910 out of 5 on each model's scored items, or 2.644--2.948 on the 191 items scored for all seven. Intensity ranking against the full human consensus gives joint $\rho_0$ from 0.026 to 0.176 and independent $\rho_0$ from 0.177 to 0.329 (Table~\ref{tab:main-results}). Figure~\ref{fig:qualitative} shows one IP video and one GRG item with model outputs.

No model leads every column: Gemini leads IP joint and consistency, GPT-5 SRM and cue, Qwen3.6-27B IP independent, and Kimi-K2.6 the GRG composite; Qwen3.6-27B also exceeds Qwen3.5-397B-A17B on IP and SRM, so parameter count alone does not predict performance. Open-weight models receive higher linking scores and lower reference-consistency scores than the two proprietary models (Table~\ref{tab:rationale}), and composite intervals overlap across generators, so we report dimension-level differences. A matched LOO comparison gives Gemini joint pairwise accuracy 0.585 against 0.607 for the held-out rater on the same 249 videos (model minus human $-0.022$, 95\% CI $[-0.081,0.036]$); independent models span 0.587--0.661 on their matched sets (Appendix~\ref{app:human-matched}).

\begin{table*}[t]
  \centering
  \caption{\textbf{Zero-shot results across all three \rb{} tasks.} IP reports mean per-video extended Spearman $\rho_0$ and pooled Pearson $r$ under the joint and independent settings. SRM reports four-way accuracy on the frozen 112-item set (chance .25), with the mean/median temporal distance of incorrect selections as descriptive columns. GRG uses one shared text-only DeepSeek-V3.1 judge; Cons. denotes reference consistency. $\Delta_{\rm J}$/$\Delta_{\rm I}$ report model minus held-out-rater pairwise accuracy under the matched leave-one-rater-out (LOO) protocol (individual-rater reference .607; joint and independent inputs); $\dagger$ marks a pointwise 95\% video-bootstrap interval that excludes zero (Table~\ref{tab:human-matched}). Bold marks the best evaluated model and underline the best open-weight model. Per-model item counts, parse-failure handling, and the chance error-distance computation are documented in Appendix~\ref{app:main-table-notes}.}
  \label{tab:main-results}
  \small
  \setlength{\tabcolsep}{4pt}
  \renewcommand{\arraystretch}{1.15}
  \resizebox{\textwidth}{!}{%
  \begin{tabular}{@{}l*{12}{c}@{}}
    \toprule
    \multirow{2}{*}{\textbf{Model}} & \multicolumn{2}{c}{\textbf{IP joint}} & \multicolumn{2}{c}{\textbf{IP independent}} & \multicolumn{3}{c}{\textbf{SRM}} & \multicolumn{3}{c}{\textbf{GRG}} & \multicolumn{2}{c}{\textbf{vs.\ human (PA)}} \\
    \cmidrule(lr){2-3}\cmidrule(lr){4-5}\cmidrule(lr){6-8}\cmidrule(lr){9-11}\cmidrule(lr){12-13}
    & $\rho_0\uparrow$ & $r\uparrow$ & $\rho_0\uparrow$ & $r\uparrow$ & Acc.$\uparrow$ & Mean err. & Med. err. & Comp./5$\uparrow$ & Cue/2$\uparrow$ & Cons./1$\uparrow$ & $\Delta_{\rm J}$ & $\Delta_{\rm I}$ \\
    \midrule
    \rowcolor[gray]{0.96}\multicolumn{13}{@{}l}{\textbf{\textit{Reference}}} \\
    Chance & N/A & N/A & N/A & N/A & .250 & 28.4 & 26.4 & N/A & N/A & N/A & N/A & N/A \\
        \midrule
    \rowcolor[gray]{0.96}\multicolumn{13}{@{}l}{\textbf{\textit{Proprietary Models}}} \\
    GPT-5 & .026 & .293 & .177 & .426 & \textbf{.509} & 29.4 & 26.8 & 2.775 & \textbf{1.245} & .465 & -.091$^\dagger$ & -.019 \\
    Gemini 3.7 Flash & \textbf{.176} & \textbf{.308} & .316 & \textbf{.518} & .500 & 26.0 & 24.8 & 2.654 & .906 & \textbf{.618} & -.022 & +.051 \\
    \midrule
    \rowcolor[gray]{0.96}\multicolumn{13}{@{}l}{\textbf{\textit{Open-weight Models}}} \\
    Qwen3.5-9B & .029 & .204 & .255 & .389 & .259 & 27.2 & 26.4 & 2.825 & .900 & .205 & -.089$^\dagger$ & +.018 \\
    Qwen3.6-27B & \underline{.163} & .238 & \textbf{\underline{.329}} & \underline{.434} & .446 & 26.6 & 25.7 & 2.610 & .815 & .270 & -.023 & +.054$^\dagger$ \\
    Qwen3.6-35B-A3B & .043 & .187 & .210 & .398 & \underline{.482} & 28.9 & 26.8 & 2.775 & .950 & .275 & -.085$^\dagger$ & -.004 \\
    Qwen3.5-397B-A17B & .065 & .217 & .267 & .387 & .411 & 25.4 & 24.7 & 2.905 & .950 & .270 & -.074$^\dagger$ & +.024 \\
    Kimi-K2.6 & .086 & \underline{.241} & .233 & .430 & .375 & 26.3 & 24.5 & \textbf{\underline{2.910}} & \underline{1.085} & \underline{.295} & -.059$^\dagger$ & +.009 \\
    \bottomrule
  \end{tabular}%
  }
\end{table*}

% Qualitative examples: one IP video (joint vs independent) and one GRG item.
% Reaction frames are Gaussian-blurred; see figures/qualitative/README.md.
\begin{figure*}[t]
  \centering
  \scriptsize
  \setlength{\tabcolsep}{2pt}
  \begin{minipage}[t]{0.335\textwidth}
    \centering
    \textbf{(a) IP: video with three marked moments}\\[2pt]
    \includegraphics[width=0.3\linewidth]{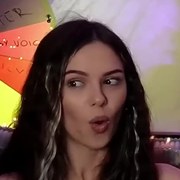}\hfill
    \includegraphics[width=0.3\linewidth]{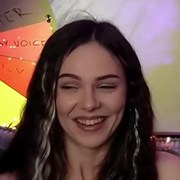}\hfill
    \includegraphics[width=0.3\linewidth]{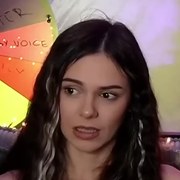}\\[-1pt]
    \makebox[\linewidth][s]{\tiny $m_1$: 29.5--33.0\,s\hfill $m_2$: 34.0--37.5\,s\hfill $m_3$: 55.5--58.5\,s}\\[3pt]
    \begin{tabular}{@{}lrrrr@{}}
      \toprule
      & $m_1$ & $m_2$ & $m_3$ & $\rho_0$ \\
      \midrule
      Human consensus & 1.50 & \textbf{2.00} & 1.25 & --- \\
      Gemini, joint & 50 & 25 & \textbf{65} & $-1.00$ \\
      Gemini, indep. & 22 & \textbf{30} & 20 & $+1.00$ \\
      Qwen3.6-27B, joint & 15 & \textbf{75} & 25 & $+0.50$ \\
      Qwen3.6-27B, indep. & 35 & \textbf{65} & 15 & $+1.00$ \\
      \bottomrule
    \end{tabular}
  \end{minipage}\hfill
  \begin{minipage}[t]{0.645\textwidth}
    \centering
    \textbf{(b) GRG: aligned stimulus window (left) and reaction (right)}\\[2pt]
    \includegraphics[height=0.142\linewidth]{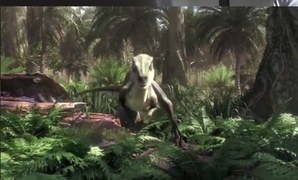}\hfill
    \includegraphics[height=0.142\linewidth]{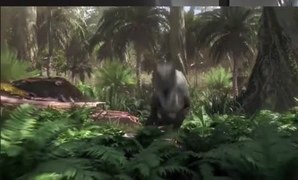}\hfill
    \includegraphics[height=0.142\linewidth]{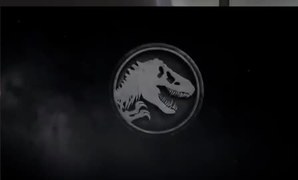}\hfill
    \includegraphics[height=0.142\linewidth]{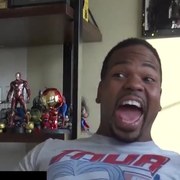}\\[3pt]
    \begin{tabular}{@{}p{0.125\linewidth}p{0.665\linewidth}c@{}}
      \toprule
      Source & Text & Cue/Link/Cons. \\
      \midrule
      Human refs. & \emph{``A dinosaur is preparing to pounce.''} \; \emph{``The cue is the dinosaur coming closer to the screen, which triggers this reaction of fearfulness.''} \; \emph{``Viewer was amazed while watching a Jurassic trailer.''} & --- \\
      GPT-5 & When the dinosaur suddenly lunges at the camera from the jungle and the screen whips forward into the Jurassic logo, it startles him. & 2 / 2 / 1 \\
      Kimi-K2.6 & The viewer's excited reaction was caused by the velociraptor suddenly lunging directly at the camera in an attack motion. & 2 / 2 / 0 \\
      Qwen3.6-27B & The viewer's reaction was caused by the sudden appearance of the Jurassic Park logo on the screen. & 0 / 0 / 0 \\
      \bottomrule
    \end{tabular}
  \end{minipage}
  \caption{\textbf{Qualitative examples.} \textbf{(a)} One IP video; the same models invert or recover the human order depending on scoring organization. Gemini's joint profile ranks $m_3$ highest ($\rho_0=-1$), whereas its independent per-moment scores match the consensus order ($\rho_0=1$); Qwen3.6-27B improves from $0.5$ to $1$. \textbf{(b)} One GRG item with its blind-written references and three of seven generator outputs, scored by the shared DeepSeek-V3.1 judge. Six generators locate the lunge; Qwen3.6-27B attributes the reaction to the logo that follows it (item identifiers in Appendix~\ref{app:robustness}).}
  \label{fig:qualitative}
\end{figure*}

\subsection{RQ3: How Does Scoring Organization Change Intensity Ranking?}
\label{sec:insights}
\label{sec:rq3}

\begin{figure*}[t]
  \centering
  \includegraphics[width=\textwidth]{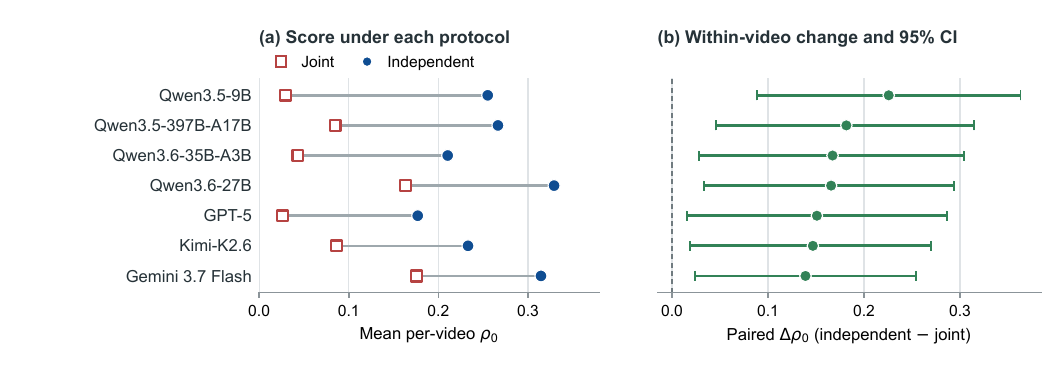}
  \caption{\textbf{Independent per-moment scoring improves intensity ranking for every model.} \textbf{(a)} Mean per-video extended correlation $\rho_0$ under joint and independent scoring on each model's paired video set; connecting lines show the within-model change. \textbf{(b)} Paired mean change $\Delta\rho_0=\rho_{0,\mathrm{independent}}-\rho_{0,\mathrm{joint}}$ with pointwise 95\% paired-bootstrap confidence intervals ($n=226$--250 videos per model). Constant predictions contribute zero correlation. All seven changes are positive (0.139--0.226), and every interval excludes zero.}
  \label{fig:elicitation}
\end{figure*}

IP performance depends substantially on the scoring protocol. Figure~\ref{fig:elicitation} compares the joint and independent IP settings on paired videos. Independent scoring improves mean $\rho_0$ by 0.139--0.226 across the seven models; every pointwise bootstrap interval excludes zero. The largest gain is Qwen3.5-9B's $+0.226$, while Qwen3.6-27B has the highest independent score ($\rho_0=0.329$); gain and final ranking therefore convey different information. The original Wilcoxon tests give $p\leq0.035$. Excluding videos with constant predictions preserves positive differences. Excluding the five videos with incorrect post-subsampling time indexing leaves all seven paired means positive, with six of seven mean-difference intervals above zero (Appendix~\ref{app:timestamp-audit}).

To further examine this protocol difference, we evaluate Qwen3.5-9B with grouped-window joint scoring: the independent condition's frames are grouped by moment, with all scores returned together. On 246 common videos, full-video joint, grouped joint, and independent achieve $\rho_0=0.039$, 0.172, and 0.278. Grouping improves on full-video joint by 0.133 ($[0.026,0.240]$), and independent improves on grouped joint by 0.105 ($[0.015,0.200]$). Grouped-window scoring also avoids the constant profiles independent scoring produces: 0 of 246 grouped outputs are constant, against 56 of 250 independent profiles. This grouped-window joint scoring isolates part of the protocol difference; frame selection, context, instructions, and output organization still co-vary (Appendix~\ref{app:grouped-joint}).

\subsection{RQ4: How Does Supervised Adaptation Change Scores and Outputs?}
\label{sec:rq4}

% Generated by scripts/export_review_tables.py from frozen finetune_paired_*.json caches.
\begin{table}[t]
\centering
\caption{\textbf{Supervised adaptation of Qwen3.5-9B.} All runs use the same local base, 250 IP videos, and 112 SRM items. IP reports full-consensus $\rho_0$, paired mean differences and 95\% bootstrap intervals; the seed row gives mean $\pm$ SD. Constants count joint profiles. Appendix~\ref{app:reproducibility} gives tests, source-overlap sensitivity analyses, and the training time-index limitation.}
\label{tab:adaptation}
\footnotesize
\setlength{\tabcolsep}{2pt}
\begin{tabular}{lrlrlrr}
\toprule
& \multicolumn{2}{c}{IP joint} & \multicolumn{2}{c}{IP independent} & Constant & SRM \\
\cmidrule(lr){2-3}\cmidrule(lr){4-5}
Run & $\rho_0$ & $\Delta\rho_0$ [95\% CI] & $\rho_0$ & $\Delta\rho_0$ [95\% CI] & (joint) & Acc. \\
\midrule
Local base & 0.024 & --- & 0.253 & --- & 0/250 & 0.304 \\
Intensity mixture, seed 0 & 0.135 & 0.112 $[-0.014,0.233]$ & 0.373 & 0.120 $[0.017,0.220]$ & 136/250 & 0.402 \\
Mixture, seeds 0--2 & 0.105 & $0.081\pm0.030$ & 0.357 & $0.104\pm0.018$ & 152$\pm$36 & --- \\
Independent only & 0.104 & 0.081 $[-0.009,0.170]$ & 0.373 & 0.119 $[0.025,0.216]$ & 4/250 & --- \\
Joint only & 0.144 & 0.121 $[0.004,0.240]$ & 0.211 & -0.042 $[-0.143,0.062]$ & 114/250 & --- \\
Matching only & 0.016 & -0.008 $[-0.030,0.010]$ & 0.224 & -0.030 $[-0.081,0.020]$ & 0/250 & 0.411 \\
Multi-task & 0.157 & 0.134 $[0.021,0.243]$ & 0.393 & 0.140 $[0.047,0.229]$ & 153/250 & 0.402 \\
\bottomrule
\end{tabular}
\end{table}

Supervised adaptation also changes measured scores. We study rank-16 LoRA adaptation of Qwen3.5-9B~\citep{hu2022lora}, comparing each adapter against the same base model under the local evaluation harness on 250 IP videos and 112 SRM items (Table~\ref{tab:adaptation}). Training and evaluation clips share source recordings, and joint training used the original frame-time instruction, whose frame-to-second mapping is incorrect after subsampling, so these results characterize the evaluated training configuration rather than transfer to unseen recordings. Appendix~\ref{app:reproducibility} reports the split audit and sensitivity analyses on 219 IP videos and 91 SRM items after excluding shared parent clips, where the multi-task gains are 0.133 joint and 0.159 independent.

The intensity mixture, trained on both joint and independent IP examples, increases joint $\rho_0$ by $0.081\pm0.030$ and independent $\rho_0$ by $0.104\pm0.018$ across three seeds. Joint mean-difference intervals include zero for each seed, while independent intervals exclude zero for two seeds. Under the matched LOO protocol, seed 0 reaches 0.681 independent pairwise accuracy, 0.074 above the individual-rater reference ($[0.031,0.116]$). Format-specific training shows a similar dependence on the evaluation protocol: independent-only supervision increases independent $\rho_0$ by 0.119 (95\% CI $[0.025,0.216]$), with only 4/250 constant joint profiles, whereas joint-only training increases joint $\rho_0$ by 0.121 but produces 114/250 constant joint profiles and decreases independent $\rho_0$ by 0.042. Every intensity-trained adapter except joint-only keeps the independent-minus-joint gap at or above the base model's 0.230. These gains can also coincide with output contraction. For intensity-mixture seed 0, the 114 videos with nonconstant adapted outputs improve from 0.096 to 0.296, a conditional gain of 0.201 compared with 0.112 over all 250 videos; both arms use the same videos in each comparison (Appendix~\ref{app:adaptation-table-notes}). Constrained-format training has been associated with reduced diversity~\citep{yun2025priceofformat,karouzos2026collapse}, and here contraction and ranking gains co-occur.

On SRM, matching-only adaptation raises accuracy from 0.304 to 0.411 (McNemar $p=0.036$), removing 17 parse failures. Its gain is only 0.011 on the 95 base-parseable items ($p=1.000$), while intensity-only adaptation reaches 0.402 with the same 0.011 gain on that parsed subset. This concentration on format-failure items is consistent with response-format stabilization~\citep{chu2025sft,wang2026testtime}. Multi-task adaptation increases joint $\rho_0$ by 0.134 and independent $\rho_0$ by 0.140, with positive mean-difference intervals. Of the joint gain, 0.030 comes from replacing predictions with constant outputs and the remaining 0.104 from improved ranking on 97 nonconstant videos (0.138 to 0.405). Under independent scoring, the nonconstant subset improves from 0.328 to 0.540, while constant profiles contribute $-0.015$ to the all-video gain. Thus, supervision improves ranking despite substantial output collapse. Under the matched LOO protocol, multi-task adaptation reaches 0.692 independent pairwise accuracy, 0.085 above the individual-rater reference ($[0.040,0.128]$), while its SRM accuracy of 0.402 matches the intensity-only adapter (Appendix~\ref{app:reproducibility}).

\section{Conclusion}

\rb{} tests whether multimodal models connect an observed viewer reaction to its stimulus. Across seven models, reaction-specific grounding remains challenging: SRM peaks at 50.9\%, and GPT-5 substitution controls show no detectable sensitivity to reaction identity. Human–model comparisons show no universal gap. Intensity ranking is protocol-dependent: independent scoring improves every model, while grouped-window scoring recovers part of the joint gap for Qwen3.5-9B. Adaptation improves ranking but can collapse score distinctions. We therefore recommend reporting elicitation format and constant-prediction rate, and validating judges on the target rubric. Task files, references, prompts, predictions, and code are released at \url{https://github.com/MIT-MI/ReactionBench} with an opt-out procedure.

\subsection*{AI use statement}

Generative AI was used as an object of evaluation, a candidate-nomination tool, and a text-based rationale judge, as described in Sections~\ref{sec:pipeline} and~\ref{sec:experiments}. No model-generated intensity score or rationale serves as a human reference; SRM supervision is temporal alignment. Generative AI tools also assisted with manuscript drafting and editing, and partial code generation. The authors take responsibility for the final content and reported evidence.

\subsection*{Ethics statement}

Appendix~\ref{app:ethics} describes research oversight and data governance for
human participants and identifiable people in publicly posted videos. Human annotation was carried out by a rater pool and by
Prolific participants; per-study compensation rates and median completion times are
disclosed there. The human-participant research has completed institutional ethics review.
The source videos are publicly posted reaction recordings; the release pseudonymizes participant
and creator identifiers, separates private annotation metadata from benchmark
labels, does not redistribute video, and documents a creator opt-out and takedown path. \rb{} is intended to
evaluate research systems that connect observable events to observable reactions
under a disclosed protocol. It is explicitly not intended for inferring private
mental states, diagnosing emotion or health, determining deception, ranking people
by emotional expressiveness, or supporting surveillance or consequential decisions
about individuals. Known limits on population coverage, measurement validity, and
research oversight are stated in Appendix~\ref{sec:limitations}.

\subsection*{Reproducibility statement}

Exact executable prompts, media ordering, input and output budgets, parsing rules,
scoring definitions, and the statistical protocol are given in
Appendix~\ref{app:prompts}. Task construction, the intensity-set selection record,
the matching separation contract, and the rationale-collection procedure are in
Appendix~\ref{app:source}. Full results under both tie conventions, the paired
scoring-protocol analysis, the superseded matching diagnostics, the native-video
control, and the judge-sensitivity analysis are in
Appendix~\ref{app:robustness}; adaptation splits, configurations, per-run results,
and the cost ledger are in Appendix~\ref{app:reproducibility}. Task files preserve
canonical item identifiers, source keys, time windows, option order, human
references, and consensus fields, and randomized set construction and bootstrap
analysis use recorded seeds. Code, frozen task files, de-identified annotations, and every
model output are released at \url{https://github.com/MIT-MI/ReactionBench}, together with
scripts that recompute the reported numbers.

\newpage
\ifdefined\ICLRNUMERIC
  \bibliographystyle{unsrtnat}
\else
  \bibliographystyle{iclr2027_conference}
\fi
\bibliography{references}

@inproceedings{zhou2026humanvbench,
  title     = {{HumanVBench}: Probing Human-Centric Video Understanding in {MLLMs} with Automatically Synthesized Benchmarks},
  author    = {Zhou, Ting and Chen, Daoyuan and Jiao, Qirui and Ding, Bolin and Li, Yaliang and Shen, Ying},
  booktitle = {Proceedings of the IEEE/CVF Conference on Computer Vision and Pattern Recognition (CVPR)},
  pages     = {4494--4504},
  month     = jun,
  year      = {2026},
  url       = {https://openaccess.thecvf.com/content/CVPR2026/html/Zhou_HumanVBench_Probing_Human-Centric_Video_Understanding_in_MLLMs_with_Automatically_Synthesized_CVPR_2026_paper.html}
}

@article{wei2026moviegraphtom,
  title   = {{MovieGraph-ToM}: Evaluating Long-Range Theory of Mind in Large Language Models via Implicit Social-Causal Graphs},
  author  = {Wei, Tingjiang and Ni, Qin and Gao, Rong and Wang, Yingying and He, Liang},
  journal = {Proceedings of the AAAI Conference on Artificial Intelligence},
  volume  = {40},
  number  = {40},
  pages   = {33827--33835},
  year    = {2026},
  doi     = {10.1609/aaai.v40i40.40674},
  url     = {https://ojs.aaai.org/index.php/AAAI/article/view/40674}
}

@article{shi2025mumat,
  title   = {{MuMA-ToM}: Multi-modal Multi-Agent Theory of Mind},
  author  = {Shi, Haojun and Ye, Suyu and Fang, Xinyu and Jin, Chuanyang and Isik, Leyla and Kuo, Yen-Ling and Shu, Tianmin},
  journal = {Proceedings of the AAAI Conference on Artificial Intelligence},
  volume  = {39},
  number  = {2},
  pages   = {1510--1519},
  year    = {2025},
  doi     = {10.1609/aaai.v39i2.32142},
  url     = {https://ojs.aaai.org/index.php/AAAI/article/view/32142}
}

@inproceedings{veerabadran2025wagibench,
  title     = {Benchmarking Egocentric Multimodal Goal Inference for Assistive Wearable Agents},
  author    = {Veerabadran, Vijay and Xiao, Fanyi and Kamra, Nitin and Matias, Pedro and Chen, Joy and Drooff, Caley and Roads, Brett and Williams, Riley J. and Henderson, Ethan and Zhao, Xuanyi and Carlberg, Kevin and Tighe, Joseph and Ridgeway, Karl},
  booktitle = {Advances in Neural Information Processing Systems},
  volume    = {38},
  year      = {2025},
  note      = {Datasets and Benchmarks Track},
  doi       = {10.52202/085713-0846},
  url       = {https://proceedings.neurips.cc/paper_files/paper/2025/hash/23ab960082db936f874b171822e0d097-Abstract-Datasets_and_Benchmarks_Track.html}
}

@inproceedings{gu2018ava,
  title     = {{AVA}: A Video Dataset of Spatio-Temporally Localized Atomic Visual Actions},
  author    = {Gu, Chunhui and Sun, Chen and Ross, David A. and Vondrick, Carl and Pantofaru, Caroline and Li, Yeqing and Vijayanarasimhan, Sudheendra and Toderici, George and Ricco, Susanna and Sukthankar, Rahul and Schmid, Cordelia and Malik, Jitendra},
  booktitle = {Proceedings of the IEEE/CVF Conference on Computer Vision and Pattern Recognition (CVPR)},
  pages     = {6047--6056},
  month     = jun,
  year      = {2018},
  url       = {https://openaccess.thecvf.com/content_cvpr_2018/html/Gu_AVA_A_Video_CVPR_2018_paper.html}
}

@inproceedings{jiang2020dfew,
  title     = {{DFEW}: A Large-Scale Database for Recognizing Dynamic Facial Expressions in the Wild},
  author    = {Jiang, Xingxun and Zong, Yuan and Zheng, Wenming and Tang, Chuangao and Xia, Wanchuang and Lu, Cheng and Liu, Jiateng},
  booktitle = {Proceedings of the 28th ACM International Conference on Multimedia},
  pages     = {2881--2889},
  year      = {2020},
  url       = {https://arxiv.org/abs/2008.05924}
}

@inproceedings{zadeh2019socialiq,
  title     = {{Social-IQ}: A Question Answering Benchmark for Artificial Social Intelligence},
  author    = {Zadeh, Amir and Chan, Michael and Liang, Paul Pu and Tong, Edmund and Morency, Louis-Philippe},
  booktitle = {Proceedings of the IEEE/CVF Conference on Computer Vision and Pattern Recognition (CVPR)},
  pages     = {8807--8817},
  month     = jun,
  year      = {2019},
  url       = {https://openaccess.thecvf.com/content_CVPR_2019/html/Zadeh_Social-IQ_A_Question_Answering_Benchmark_for_Artificial_Social_Intelligence_CVPR_2019_paper.html}
}

@inproceedings{zhang2026castbench,
  title     = {{CaST-Bench}: Benchmarking Causal Chain-Grounded Spatio-Temporal Reasoning for Video Question Answering},
  author    = {Zhang, Mingfang and Pan, Jingjing and Kumar, Ashutosh and Saini, Rajat and Erdogan, Mustafa and Yang, Hsuan-Kung and Kang, Caixin and Huang, Yifei and Sato, Yoichi and Kong, Quan},
  booktitle = {Proceedings of the IEEE/CVF Conference on Computer Vision and Pattern Recognition (CVPR)},
  pages     = {31856--31866},
  month     = jun,
  year      = {2026},
  url       = {https://openaccess.thecvf.com/content/CVPR2026/html/Zhang_CaST-Bench_Benchmarking_Causal_Chain-Grounded_Spatio-Temporal_Reasoning_for_Video_Question_Answering_CVPR_2026_paper.html}
}

@article{xie2026maverix,
  title   = {{MAVERIX}: Multimodal Audio-Visual Evaluation and Recognition IndeX},
  author  = {Xie, Liuyue and Kuthiala, Avik and Wei, George Z. and Zheng, Ce and Bal, Ananya and Dabhi, Mosam and Wen, Liting and Rustagi, Taru and Lai, Ethan and Khyalia, Sushil and Choudhury, Rohan and Ziyadi, Morteza and Zhang, Xu and Yang, Hao and Jeni, Laszlo A.},
  journal = {Proceedings of the AAAI Conference on Artificial Intelligence},
  volume  = {40},
  number  = {32},
  pages   = {27090--27098},
  year    = {2026},
  doi     = {10.1609/aaai.v40i32.39923},
  url     = {https://ojs.aaai.org/index.php/AAAI/article/view/39923}
}

@inproceedings{fu2025videomme,
  title     = {{Video-MME}: The First-Ever Comprehensive Evaluation Benchmark of Multi-modal {LLMs} in Video Analysis},
  author    = {Fu, Chaoyou and Dai, Yuhan and Luo, Yongdong and Li, Lei and Ren, Shuhuai and Zhang, Renrui and Wang, Zihan and Zhou, Chenyu and Shen, Yunhang and Zhang, Mengdan and Chen, Peixian and Li, Yanwei and Lin, Shaohui and Zhao, Sirui and Li, Ke and Xu, Tong and Zheng, Xiawu and Chen, Enhong and Shan, Caifeng and He, Ran and Sun, Xing},
  booktitle = {Proceedings of the IEEE/CVF Conference on Computer Vision and Pattern Recognition (CVPR)},
  pages     = {24108--24118},
  month     = jun,
  year      = {2025},
  url       = {https://openaccess.thecvf.com/content/CVPR2025/html/Fu_Video-MME_The_First-Ever_Comprehensive_Evaluation_Benchmark_of_Multi-modal_LLMs_in_CVPR_2025_paper.html}
}

@inproceedings{li2024mvbench,
  title     = {{MVBench}: A Comprehensive Multi-modal Video Understanding Benchmark},
  author    = {Li, Kunchang and Wang, Yali and He, Yinan and Li, Yizhuo and Wang, Yi and Liu, Yi and Wang, Zun and Xu, Jilan and Chen, Guo and Luo, Ping and Wang, Limin and Qiao, Yu},
  booktitle = {Proceedings of the IEEE/CVF Conference on Computer Vision and Pattern Recognition (CVPR)},
  pages     = {22195--22206},
  month     = jun,
  year      = {2024},
  url       = {https://openaccess.thecvf.com/content/CVPR2024/html/Li_MVBench_A_Comprehensive_Multi-modal_Video_Understanding_Benchmark_CVPR_2024_paper.html}
}

@inproceedings{mangalam2023egoschema,
  title     = {{EgoSchema}: A Diagnostic Benchmark for Very Long-form Video Language Understanding},
  author    = {Mangalam, Karttikeya and Akshulakov, Raiymbek and Malik, Jitendra},
  booktitle = {Advances in Neural Information Processing Systems},
  volume    = {36},
  year      = {2023},
  note      = {Datasets and Benchmarks Track},
  url       = {https://openreview.net/forum?id=JVlWseddak}
}

@inproceedings{he2024mmworld,
  title     = {{MMWorld}: Towards Multi-discipline Multi-faceted World Model Evaluation in Videos},
  author    = {He, Xuehai and Feng, Weixi and Zheng, Kaizhi and Lu, Yujie and Zhu, Wanrong and Li, Jiachen and Fan, Yue and Wang, Jianfeng and Li, Linjie and Yang, Zhengyuan and Lin, Kevin and Wang, William Yang and Wang, Lijuan and Wang, Xin Eric},
  booktitle = {International Conference on Learning Representations (ICLR)},
  year      = {2025},
  url       = {https://arxiv.org/abs/2406.08407}
}

@inproceedings{luo2025videoautoarena,
  title     = {{VideoAutoArena}: An Automated Arena for Evaluating Large Multimodal Models in Video Analysis through User Simulation},
  author    = {Luo, Ziyang and Wu, Haoning and Li, Dongxu and Ma, Jing and Kankanhalli, Mohan and Li, Junnan},
  booktitle = {Proceedings of the Computer Vision and Pattern Recognition Conference (CVPR)},
  pages     = {8461--8474},
  month     = jun,
  year      = {2025},
  url       = {https://openaccess.thecvf.com/content/CVPR2025/html/Luo_VideoAutoArena_An_Automated_Arena_for_Evaluating_Large_Multimodal_Models_in_CVPR_2025_paper.html}
}

@inproceedings{lin2024e3,
  title     = {{$E^3$}: Exploring Embodied Emotion Through A Large-Scale Egocentric Video Dataset},
  author    = {Lin, Wang and Feng, Yueying and Han, Wenkang and Jin, Tao and Zhao, Zhou and Wu, Fei and Yao, Chang and Chen, Jingyuan},
  booktitle = {Advances in Neural Information Processing Systems},
  volume    = {37},
  year      = {2024},
  note      = {Datasets and Benchmarks Track},
  doi       = {10.52202/079017-3753},
  url       = {https://proceedings.neurips.cc/paper_files/paper/2024/hash/d611d5c0251d9680f869c5d2c46c6fcd-Abstract-Datasets_and_Benchmarks_Track.html}
}

@inproceedings{cheng2024emotionllama,
  title     = {{Emotion-LLaMA}: Multimodal Emotion Recognition and Reasoning with Instruction Tuning},
  author    = {Cheng, Zebang and Cheng, Zhi-Qi and He, Jun-Yan and Sun, Jingdong and Wang, Kai and Lin, Yuxiang and Lian, Zheng and Peng, Xiaojiang and Hauptmann, Alexander G.},
  booktitle = {Advances in Neural Information Processing Systems},
  volume    = {37},
  year      = {2024},
  doi       = {10.52202/079017-3518},
  url       = {https://proceedings.neurips.cc/paper_files/paper/2024/hash/c7f43ada17acc234f568dc66da527418-Abstract-Conference.html}
}

@inproceedings{jia2024inducedsentiment,
  title     = {Infer Induced Sentiment of Comment Response to Video: A New Task, Dataset and Baseline},
  author    = {Jia, Qi and Fan, Baoyu and Xu, Cong and Liu, Lu and Jin, Liang and Du, Guoguang and Guo, Zhenhua and Zhao, Yaqian and Huang, Xuanjing and Li, Rengang},
  booktitle = {Advances in Neural Information Processing Systems},
  volume    = {37},
  year      = {2024},
  note      = {Datasets and Benchmarks Track},
  doi       = {10.52202/079017-3295},
  url       = {https://proceedings.neurips.cc/paper_files/paper/2024/hash/bbf090d264b94d29260f5303efea868c-Abstract-Datasets_and_Benchmarks_Track.html}
}

@inproceedings{jammot2025egoemotion,
  title     = {{egoEMOTION}: Egocentric Vision and Physiological Signals for Emotion and Personality Recognition in Real-world Tasks},
  author    = {Jammot, Matthias and Braun, Bj{\"o}rn and Streli, Paul and Wampfler, Rafael and Holz, Christian},
  booktitle = {Advances in Neural Information Processing Systems},
  volume    = {38},
  year      = {2025},
  note      = {Datasets and Benchmarks Track},
  doi       = {10.52202/085713-4255},
  url       = {https://proceedings.neurips.cc/paper_files/paper/2025/hash/b9784f4c3685bebeacd387c4e680b5b2-Abstract-Datasets_and_Benchmarks_Track.html}
}

@inproceedings{huang2024emotionbench,
  title     = {Apathetic or Empathetic? Evaluating {LLMs}' Emotional Alignments with Humans},
  author    = {Huang, Jen-tse and Lam, Man Ho and Li, Eric John and Ren, Shujie and Wang, Wenxuan and Jiao, Wenxiang and Tu, Zhaopeng and Lyu, Michael R.},
  booktitle = {Advances in Neural Information Processing Systems},
  volume    = {37},
  year      = {2024},
  doi       = {10.52202/079017-3077},
  url       = {https://proceedings.neurips.cc/paper_files/paper/2024/hash/b0049c3f9c53fb06f674ae66c2cf2376-Abstract-Conference.html}
}

@inproceedings{ivey2025nutmeg,
  title     = {{NUTMEG}: Separating Signal From Noise in Annotator Disagreement},
  author    = {Ivey, Jonathan and Gauch, Susan and Jurgens, David},
  booktitle = {Proceedings of the 2025 Conference on Empirical Methods in Natural Language Processing},
  pages     = {2874--2887},
  address   = {Suzhou, China},
  publisher = {Association for Computational Linguistics},
  month     = nov,
  year      = {2025},
  doi       = {10.18653/v1/2025.emnlp-main.144},
  url       = {https://aclanthology.org/2025.emnlp-main.144/}
}

@inproceedings{tan2025judgebench,
  title     = {{JudgeBench}: A Benchmark for Evaluating {LLM}-Based Judges},
  author    = {Tan, Sijun and Zhuang, Siyuan and Montgomery, Kyle and Tang, William Yuan and Cuadron, Alejandro and Wang, Chenguang and Popa, Raluca Ada and Stoica, Ion},
  booktitle = {The Thirteenth International Conference on Learning Representations},
  year      = {2025},
  url       = {https://openreview.net/forum?id=G0dksFayVq}
}

@inproceedings{salinas2025judgetuning,
  title     = {Tuning {LLM} Judge Design Decisions for 1/1000 of the Cost},
  author    = {Salinas, David and Swelam, Omar and Hutter, Frank},
  booktitle = {Proceedings of the 42nd International Conference on Machine Learning},
  pages     = {52728--52744},
  volume    = {267},
  series    = {Proceedings of Machine Learning Research},
  publisher = {PMLR},
  month     = jul,
  year      = {2025},
  url       = {https://proceedings.mlr.press/v267/salinas25a.html}
}

@inproceedings{watts2024pariksha,
  title     = {{PARIKSHA}: A Large-Scale Investigation of Human-{LLM} Evaluator Agreement on Multilingual and Multi-Cultural Data},
  author    = {Watts, Ishaan and Gumma, Varun and Yadavalli, Aditya and Seshadri, Vivek and Swaminathan, Manohar and Sitaram, Sunayana},
  booktitle = {Proceedings of the 2024 Conference on Empirical Methods in Natural Language Processing},
  pages     = {7900--7932},
  address   = {Miami, Florida, USA},
  publisher = {Association for Computational Linguistics},
  month     = nov,
  year      = {2024},
  doi       = {10.18653/v1/2024.emnlp-main.451},
  url       = {https://aclanthology.org/2024.emnlp-main.451/}
}

@inproceedings{panickssery2024selfpreference,
  title     = {{LLM} Evaluators Recognize and Favor Their Own Generations},
  author    = {Panickssery, Arjun and Bowman, Samuel R. and Feng, Shi},
  booktitle = {Advances in Neural Information Processing Systems},
  volume    = {37},
  year      = {2024},
  doi       = {10.52202/079017-2197},
  url       = {https://proceedings.neurips.cc/paper_files/paper/2024/hash/7f1f0218e45f5414c79c0679633e47bc-Abstract-Conference.html}
}

@inproceedings{zheng2023mtbench,
  title     = {Judging {LLM}-as-a-Judge with {MT-Bench} and Chatbot Arena},
  author    = {Zheng, Lianmin and Chiang, Wei-Lin and Sheng, Ying and Zhuang, Siyuan and Wu, Zhanghao and Zhuang, Yonghao and Lin, Zi and Li, Zhuohan and Li, Dacheng and Xing, Eric P. and Zhang, Hao and Gonzalez, Joseph E. and Stoica, Ion},
  booktitle = {Advances in Neural Information Processing Systems},
  volume    = {36},
  year      = {2023},
  note      = {Datasets and Benchmarks Track},
  doi       = {10.52202/075280-2020},
  url       = {https://proceedings.neurips.cc/paper_files/paper/2023/hash/91f18a1287b398d378ef22505bf41832-Abstract-Datasets_and_Benchmarks.html}
}

@inproceedings{liu2023geval,
  title     = {{G-Eval}: {NLG} Evaluation using {GPT}-4 with Better Human Alignment},
  author    = {Liu, Yang and Iter, Dan and Xu, Yichong and Wang, Shuohang and Xu, Ruochen and Zhu, Chenguang},
  booktitle = {Proceedings of the 2023 Conference on Empirical Methods in Natural Language Processing},
  pages     = {2511--2522},
  address   = {Singapore},
  publisher = {Association for Computational Linguistics},
  month     = dec,
  year      = {2023},
  doi       = {10.18653/v1/2023.emnlp-main.153},
  url       = {https://aclanthology.org/2023.emnlp-main.153/}
}

@article{sun2021eev,
  title   = {{EEV}: A Large-Scale Dataset for Studying Evoked Expressions from Video},
  author  = {Sun, Jennifer J. and Liu, Ting and Cowen, Alan S. and Schroff, Florian and Adam, Hartwig and Prasad, Gautam},
  journal = {arXiv preprint arXiv:2001.05488},
  year    = {2021},
  url     = {https://arxiv.org/abs/2001.05488}
}

@inproceedings{li2023reactionet,
  title     = {{ReactioNet}: Learning High-Order Facial Behavior from Universal Stimulus-Reaction by Dyadic Relation Reasoning},
  author    = {Li, Xiaotian and Wang, Taoyue and Zhao, Geran and Zhang, Xiang and Kang, Xi and Yin, Lijun},
  booktitle = {Proceedings of the IEEE/CVF International Conference on Computer Vision (ICCV)},
  month     = oct,
  year      = {2023},
  doi       = {10.1109/ICCV51070.2023.01899},
  url       = {https://openaccess.thecvf.com/content/ICCV2023/papers/Li_ReactioNet_Learning_High-Order_Facial_Behavior_from_Universal_Stimulus-Reaction_by_Dyadic_ICCV_2023_paper.pdf}
}

@article{guo2025stimuvar,
  title   = {{StimuVAR}: Spatiotemporal Stimuli-aware Video Affective Reasoning with Multimodal Large Language Models},
  author  = {Guo, Yuxiang and Siddiqui, Faizan and Zhao, Yang and Chellappa, Rama and Lo, Shao-Yuan},
  journal = {arXiv preprint arXiv:2409.00304},
  year    = {2024},
  note    = {Accepted by the International Journal of Computer Vision},
  url     = {https://arxiv.org/abs/2409.00304}
}

@inproceedings{sul2023mrhisum,
  title     = {{Mr.\ HiSum}: A Large-scale Dataset for Video Highlight Detection and Summarization},
  author    = {Sul, Jinhwan and Han, Jihoon and Lee, Joonseok},
  booktitle = {Advances in Neural Information Processing Systems (NeurIPS) Track on Datasets and Benchmarks},
  year      = {2023},
  url       = {https://proceedings.neurips.cc/paper_files/paper/2023/hash/7f880e3a325b06e3601af1384a653038-Abstract-Datasets_and_Benchmarks.html}
}

@inproceedings{jin2024mmtomqa,
  title     = {{MMToM-QA}: Multimodal Theory of Mind Question Answering},
  author    = {Jin, Chuanyang and Wu, Yutong and Cao, Jing and Xiang, Jiannan and Kuo, Yen-Ling and Hu, Zhiting and Ullman, Tomer and Torralba, Antonio and Tenenbaum, Joshua B. and Shu, Tianmin},
  booktitle = {Proceedings of the 62nd Annual Meeting of the Association for Computational Linguistics (ACL)},
  year      = {2024},
  url       = {https://arxiv.org/abs/2401.08743}
}

@inproceedings{wang2024semeval,
  title     = {{SemEval-2024} Task 3: Multimodal Emotion Cause Analysis in Conversations},
  author    = {Wang, Fanfan and Ma, Heqing and Yu, Jianfei and Xia, Rui and Cambria, Erik},
  booktitle = {Proceedings of the 18th International Workshop on Semantic Evaluation (SemEval-2024)},
  pages     = {2039--2050},
  year      = {2024},
  url       = {https://aclanthology.org/2024.semeval-1.277/}
}

@inproceedings{ng2023listen,
  title     = {Can Language Models Learn to Listen?},
  author    = {Ng, Evonne and Subramanian, Sanjay and Klein, Dan and Kanazawa, Angjoo and Darrell, Trevor and Ginosar, Shiry},
  booktitle = {Proceedings of the IEEE/CVF International Conference on Computer Vision (ICCV)},
  year      = {2023},
  url       = {https://arxiv.org/abs/2308.10897}
}

@article{agrawal2025seamless,
  title   = {Seamless Interaction: Dyadic Audiovisual Motion Modeling and Large-Scale Dataset},
  author  = {Agrawal, Vasu and Akinyemi, Akinniyi and Alvero, Kathryn and others},
  journal = {arXiv preprint arXiv:2506.22554},
  year    = {2025},
  url     = {https://arxiv.org/abs/2506.22554}
}

@inproceedings{lei2021qvhighlights,
  title     = {{QVHighlights}: Detecting Moments and Highlights in Videos via Natural Language Queries},
  author    = {Lei, Jie and Berg, Tamara L. and Bansal, Mohit},
  booktitle = {Advances in Neural Information Processing Systems (NeurIPS)},
  year      = {2021},
  url       = {https://arxiv.org/abs/2107.09609}
}

@article{yu2025hero,
  title   = {{HERO}: Human Reaction Generation from Videos},
  author  = {Yu, Chengjun and Zhai, Wei and Yang, Yuhang and Cao, Yang and Zha, Zheng-Jun},
  journal = {arXiv preprint arXiv:2503.08270},
  year    = {2025},
  url     = {https://arxiv.org/abs/2503.08270}
}

@article{fu2026videommev2,
  title   = {{Video-MME-v2}: Towards the Next Stage in Benchmarks for Comprehensive Video Understanding},
  author  = {Fu, Chaoyou and Yuan, Haozhi and Dong, Yuhao and Zhang, Yi-Fan and Shen, Yunhang and Hu, Xiaoxing and Li, Xueying and Su, Jinsen and Long, Chengwu and Xie, Xiaoyao and Xie, Yongkang and Zheng, Xiawu and Yang, Xue and Cao, Haoyu and Wu, Yunsheng and Liu, Ziwei and Sun, Xing and Shan, Caifeng and He, Ran},
  journal = {arXiv preprint arXiv:2604.05015},
  year    = {2026},
  url     = {https://arxiv.org/abs/2604.05015}
}

@inproceedings{wu2024longvideobench,
  title     = {{LongVideoBench}: A Benchmark for Long-context Interleaved Video-Language Understanding},
  author    = {Wu, Haoning and Li, Dongxu and Chen, Bei and Li, Junnan},
  booktitle = {Advances in Neural Information Processing Systems},
  year      = {2024},
  note      = {Datasets and Benchmarks Track},
  url       = {https://proceedings.neurips.cc/paper_files/paper/2024/hash/329ad516cf7a6ac306f29882e9c77558-Abstract-Datasets_and_Benchmarks_Track.html}
}

@inproceedings{shangguan2025tomato,
  title     = {{TOMATO}: Assessing Visual Temporal Reasoning Capabilities in Multimodal Foundation Models},
  author    = {Shangguan, Ziyao and Li, Chuhan and Ding, Yuxuan and Zheng, Yanan and Zhao, Yilun and Fitzgerald, Tesca and Cohan, Arman},
  booktitle = {International Conference on Learning Representations (ICLR)},
  year      = {2025},
  url       = {https://arxiv.org/abs/2410.23266}
}

@inproceedings{chen2025cgbench,
  title     = {{CG-Bench}: Clue-grounded Question Answering Benchmark for Long Video Understanding},
  author    = {Chen, Guo and Liu, Yicheng and Huang, Yifei and He, Yuping and Pei, Baoqi and Xu, Jilan and Wang, Yali and Lu, Tong and Wang, Limin},
  booktitle = {International Conference on Learning Representations (ICLR)},
  year      = {2025},
  url       = {https://arxiv.org/abs/2412.12075}
}

@article{cores2024tvbench,
  title   = {Lost in Time: A New Temporal Benchmark for {VideoLLMs}},
  author  = {Cores, Daniel and Dorkenwald, Michael and Mucientes, Manuel and Snoek, Cees G. M. and Asano, Yuki M.},
  journal = {arXiv preprint arXiv:2410.07752},
  year    = {2024},
  url     = {https://arxiv.org/abs/2410.07752}
}

@inproceedings{hong2026worldsense,
  title     = {{WorldSense}: Evaluating Real-world Omnimodal Understanding for Multimodal {LLMs}},
  author    = {Hong, Jack and Yan, Shilin and Cai, Jiayin and Jiang, Xiaolong and Hu, Yao and Xie, Weidi},
  booktitle = {International Conference on Learning Representations (ICLR)},
  year      = {2026},
  url       = {https://arxiv.org/abs/2502.04326}
}

@inproceedings{yang2025avut,
  title     = {Audio-centric Video Understanding Benchmark without Text Shortcut},
  author    = {Yang, Yudong and Zhuang, Jimin and Sun, Guangzhi and Tang, Changli and Li, Yixuan and Li, Peihan and Jiang, Yifan and Li, Wei and Ma, Zejun and Zhang, Chao},
  booktitle = {Proceedings of the 2025 Conference on Empirical Methods in Natural Language Processing (EMNLP)},
  year      = {2025},
  url       = {https://aclanthology.org/2025.emnlp-main.333/}
}

@inproceedings{mathur2025socialgenome,
  title     = {Social Genome: Grounded Social Reasoning Abilities of Multimodal Models},
  author    = {Mathur, Leena and Qian, Marian and Liang, Paul Pu and Morency, Louis-Philippe},
  booktitle = {Proceedings of the 2025 Conference on Empirical Methods in Natural Language Processing (EMNLP)},
  pages     = {24868--24891},
  year      = {2025},
  url       = {https://aclanthology.org/2025.emnlp-main.1264/}
}

@inproceedings{qin2026humansense,
  title     = {{HumanSense}: From Multimodal Perception to Empathetic Context-Aware Responses through Reasoning {MLLMs}},
  author    = {Qin, Zheng and Zheng, Ruobing and Wang, Yabing and Li, Tianqi and Yuan, Yi and Chen, Jingdong and Wang, Le},
  booktitle = {Proceedings of the AAAI Conference on Artificial Intelligence},
  year      = {2026},
  url       = {https://arxiv.org/abs/2508.10576}
}

@article{hu2025emobenchm,
  title   = {{EmoBench-M}: Benchmarking Emotional Intelligence for Multimodal Large Language Models},
  author  = {Hu, He and You, Lianzhong and Xu, Hongbo and Wang, Qianning and Yu, Fei Richard and Ma, Fei and Cheng, Zebang and Lian, Zheng and Zhou, Yucheng and Cui, Laizhong},
  journal = {arXiv preprint arXiv:2502.04424},
  year    = {2025},
  url     = {https://arxiv.org/abs/2502.04424}
}

@article{zhang2025mmeemotion,
  title   = {{MME-Emotion}: A Holistic Evaluation Benchmark for Emotional Intelligence in Multimodal Large Language Models},
  author  = {Zhang, Fan and Cheng, Zebang and Deng, Chong and Li, Haoxuan and Lian, Zheng and Chen, Qian and Liu, Huadai and Wang, Wen and others},
  journal = {arXiv preprint arXiv:2508.09210},
  year    = {2025},
  url     = {https://arxiv.org/abs/2508.09210}
}

@inproceedings{lin2025whywefeel,
  title     = {Why We Feel: Breaking Boundaries in Emotional Reasoning with Multimodal Large Language Models},
  author    = {Lin, Yuxiang and Sun, Jingdong and Cheng, Zhi-Qi and Wang, Jue and Liang, Haomin and Cheng, Zebang and Dong, Yifei and He, Jun-Yan and Peng, Xiaojiang and Hua, Xian-Sheng},
  booktitle = {Proceedings of the IEEE/CVF Conference on Computer Vision and Pattern Recognition (CVPR) Workshops},
  year      = {2025},
  note      = {Workshop on New Trends in Expressive Discovery},
  url       = {https://arxiv.org/abs/2504.07521}
}

@article{hu2026emotrans,
  title   = {{EmoTrans}: A Benchmark for Understanding, Reasoning, and Predicting Emotion Transitions in Multimodal {LLMs}},
  author  = {Hu, He and Weng, Tengjin and Cheng, Zebang and Wang, Yu and Luo, Jiachen and Schuller, Bj{\"o}rn and Lian, Zheng and Cui, Laizhong},
  journal = {arXiv preprint arXiv:2604.23348},
  year    = {2026},
  url     = {https://arxiv.org/abs/2604.23348}
}

@article{zhou2024memobench,
  title   = {{MEMO-Bench}: A Multiple Benchmark for Text-to-Image and Multimodal Large Language Models on Human Emotion Analysis},
  author  = {Zhou, Yingjie and Zhang, Zicheng and Cao, Jiezhang and Jia, Jun and Jiang, Yanwei and Wen, Farong and Liu, Xiaohong and Min, Xiongkuo and Zhai, Guangtao},
  journal = {arXiv preprint arXiv:2411.11235},
  year    = {2024},
  url     = {https://arxiv.org/abs/2411.11235}
}

@article{xing2025emotionhallucer,
  title   = {{EmotionHallucer}: Evaluating Emotion Hallucinations in Multimodal Large Language Models},
  author  = {Xing, Bohao and Liu, Xin and Zhao, Guoying and Liu, Chengyu and Fu, Xiaolan and K{\"a}lvi{\"a}inen, Heikki},
  journal = {arXiv preprint arXiv:2505.11405},
  year    = {2025},
  url     = {https://arxiv.org/abs/2505.11405}
}

@article{villacueva2025moments,
  title   = {{MoMentS}: A Comprehensive Multimodal Benchmark for Theory of Mind},
  author  = {Villa-Cueva, Emilio and Ahmed, S M Masrur and Chevi, Rendi and Cruz, Jan Christian Blaise and Elzeky, Kareem and Cristobal, Fermin and others},
  journal = {arXiv preprint arXiv:2507.04415},
  year    = {2025},
  url     = {https://arxiv.org/abs/2507.04415}
}

@article{han2025longinsightbench,
  title   = {{LongInsightBench}: A Comprehensive Benchmark for Evaluating Omni-Modal Models on Human-Centric Long-Video Understanding},
  author  = {Han, ZhaoYang and Lin, Qihan and Liang, Hao and Chen, Bowen and Liu, Zhou and Zhang, Wentao},
  journal = {arXiv preprint arXiv:2510.17305},
  year    = {2025},
  url     = {https://arxiv.org/abs/2510.17305}
}

@article{nguyen2026video2reaction,
  title   = {{Video2Reaction}: Mapping Video to Audience Reaction Distribution in the Wild},
  author  = {Nguyen, Trang and Zhang, Sidong and Shankar, Shiv and Jagatap, Gauri and Chandran, Deepak and Fanelli, Andrea and Fiterau, Madalina},
  journal = {arXiv preprint arXiv:2607.06875},
  year    = {2026},
  url     = {https://arxiv.org/abs/2607.06875}
}

@article{shrestha2026synthetic,
  title   = {Multimodal Large Language Models as Synthetic Participants in Video-Based Studies: An Evaluation},
  author  = {Shrestha, Prabal and Jiang, Bohan and Xue, Haoning and Liu, Huan and Zhou, Xinyi},
  journal = {arXiv preprint arXiv:2606.07541},
  year    = {2026},
  url     = {https://arxiv.org/abs/2606.07541}
}

@article{yang2026grace,
  title   = {{GRACE}: Boosting Video {MLLMs} with Grounded Action-Centric Evidence for Viewer Sentiment Prediction},
  author  = {Yang, Ruoxuan and Chen, Tieyuan and Huang, Xiaofeng and Yin, Haibing and Wang, Jun and Chen, Xiping and others},
  journal = {arXiv preprint arXiv:2606.16198},
  year    = {2026},
  url     = {https://arxiv.org/abs/2606.16198}
}

@article{lian2025emoprefer,
  title   = {{EmoPrefer}: Can Large Language Models Understand Human Emotion Preferences?},
  author  = {Lian, Zheng and Sun, Licai and Chen, Lan and Chen, Haoyu and Cheng, Zebang and Zhang, Fan and others},
  journal = {arXiv preprint arXiv:2507.04278},
  year    = {2025},
  url     = {https://arxiv.org/abs/2507.04278}
}

@article{chen2026multiemobench,
  title   = {{MultiEmo-Bench}: Multi-label Visual Emotion Analysis for Multi-modal Large Language Models},
  author  = {Chen, Tianwei and Furusawa, Takuya and Hirakawa, Yuki and Shimizu, Ryotaro and Fan, Mo and Wada, Takashi},
  journal = {arXiv preprint arXiv:2605.14635},
  year    = {2026},
  url     = {https://arxiv.org/abs/2605.14635}
}

@article{song2026react,
  title   = {{REACT} 2026: The Fourth Multiple Appropriate Facial Reaction Generation
             Challenge: Personalised {MAFRG} and Appropriate {EEG} Reaction Prediction},
  author  = {Song, Siyang and Spitale, Micol and Wu, Zijian and Kong, Xiangyu and Luo, Cheng and Palmero, Cristina and others},
  journal = {arXiv preprint arXiv:2606.07935},
  year    = {2026},
  url     = {https://arxiv.org/abs/2606.07935}
}

@article{zhou2026musteernet,
  title   = {{MuSteerNet}: Human Reaction Generation from Videos via Observation-Reaction Mutual Steering},
  author  = {Zhou, Yuan and Li, Yongzhi and Dai, Yanqi and Zhu, Xingyu and Tan, Yi and Xu, Qingshan and Zhu, Beier and Hong, Richang and Zhang, Hanwang},
  journal = {arXiv preprint arXiv:2603.20187},
  year    = {2026},
  url     = {https://arxiv.org/abs/2603.20187}
}

@article{li2025vividlistener,
  title   = {{VividListener}: Expressive and Controllable Listener Dynamics Modeling for Multi-Modal Responsive Interaction},
  author  = {Li, Shiying and Qi, Xingqun and Yang, Bingkun and Weile, Chen and Tian, Zezhao and Sun, Muyi and Liu, Qifeng and Zhang, Man and Sun, Zhenan},
  journal = {arXiv preprint arXiv:2504.21718},
  year    = {2025},
  url     = {https://arxiv.org/abs/2504.21718}
}

@inproceedings{chang2025aha,
  title     = {{AHA} -- Predicting What Matters Next: Online Highlight Detection Without Looking Ahead},
  author    = {Chang, Aiden and De Melo, Celso and Lukin, Stephanie M.},
  booktitle = {Advances in Neural Information Processing Systems},
  year      = {2025},
  url       = {https://arxiv.org/abs/2509.16421}
}

@article{sun2025engagement,
  title   = {Engagement Prediction of Short Videos with Large Multimodal Models},
  author  = {Sun, Wei and Cao, Linhan and Cao, Yuqin and Zhang, Weixia and Wen, Wen and Zhang, Kaiwei and Chen, Zijian and Lu, Fangfang and Min, Xiongkuo and Zhai, Guangtao},
  journal = {arXiv preprint arXiv:2508.02516},
  year    = {2025},
  url     = {https://arxiv.org/abs/2508.02516}
}

@article{gu2025judgesurvey,
  title   = {A Survey on {LLM}-as-a-Judge},
  author  = {Gu, Jiawei and Jiang, Xuhui and Shi, Zhichao and Tan, Hexiang and Zhai, Xuehao and Xu, Chengjin and others},
  journal = {arXiv preprint arXiv:2411.15594},
  year    = {2025},
  url     = {https://arxiv.org/abs/2411.15594}
}

@inproceedings{bavaresco2025llmsinstead,
  title     = {{LLM}s instead of Human Judges? A Large Scale Empirical Study across 20 {NLP} Evaluation Tasks},
  author    = {Bavaresco, Anna and Bernardi, Raffaella and Bertolazzi, Leonardo and Elliott, Desmond and Fern{\'a}ndez, Raquel and Gatt, Albert and others},
  booktitle = {Proceedings of the 63rd Annual Meeting of the Association for Computational Linguistics (Volume 2: Short Papers)},
  pages     = {238--255},
  year      = {2025},
  url       = {https://aclanthology.org/2025.acl-short.20/}
}

@article{calderon2025alttest,
  title   = {The Alternative Annotator Test for {LLM}-as-a-Judge: How to Statistically Justify Replacing Human Annotators with {LLMs}},
  author  = {Calderon, Nitay and Reichart, Roi and Dror, Rotem},
  journal = {arXiv preprint arXiv:2501.10970},
  year    = {2025},
  url     = {https://arxiv.org/abs/2501.10970}
}

@article{norman2026reliability,
  title   = {Reliability without Validity: A Systematic, Large-Scale Evaluation of {LLM}-as-a-Judge Models Across Agreement, Consistency, and Bias},
  author  = {Norman, Justin D. and Rivera, Michael U. and Hughes, D. Alex},
  journal = {arXiv preprint arXiv:2606.19544},
  year    = {2026},
  url     = {https://arxiv.org/abs/2606.19544}
}

@inproceedings{chen2024mllmasajudge,
  title     = {{MLLM}-as-a-Judge: Assessing Multimodal {LLM}-as-a-Judge with Vision-Language Benchmark},
  author    = {Chen, Dongping and Chen, Ruoxi and Zhang, Shilin and Liu, Yinuo and Wang, Yaochen and Zhou, Huichi and others},
  booktitle = {Proceedings of the 41st International Conference on Machine Learning (ICML)},
  year      = {2024},
  url       = {https://arxiv.org/abs/2402.04788}
}

@article{waheed2025videojudge,
  title   = {{VideoJudge}: Bootstrapping Enables Scalable Supervision of {MLLM}-as-a-Judge for Video Understanding},
  author  = {Waheed, Abdul and Wu, Zhen and Alharthi, Dareen and Kim, Seungone and Raj, Bhiksha},
  journal = {arXiv preprint arXiv:2509.21451},
  year    = {2025},
  url     = {https://arxiv.org/abs/2509.21451}
}

@article{xu2026beyondconsensus,
  title   = {Beyond Consensus: Perspectivist Modeling and Evaluation of Annotator Disagreement in {NLP}},
  author  = {Xu, Yinuo and Jurgens, David},
  journal = {arXiv preprint arXiv:2601.09065},
  year    = {2026},
  url     = {https://arxiv.org/abs/2601.09065}
}

@inproceedings{bean2025constructvalidity,
  title     = {Measuring what Matters: Construct Validity in Large Language Model Benchmarks},
  author    = {Bean, Andrew M. and Kearns, Ryan Othniel and Romanou, Angelika and Hafner, Franziska Sofia and Mayne, Harry and Batzner, Jan and others},
  booktitle = {Advances in Neural Information Processing Systems},
  year      = {2025},
  note      = {Datasets and Benchmarks Track},
  url       = {https://arxiv.org/abs/2511.04703}
}

@article{inoshita2026emotionuncertainty,
  title   = {{LLMs} Capture Emotion Labels, Not Emotion Uncertainty: Distributional Analysis and Calibration of Human--{LLM} Judgment Gaps},
  author  = {Inoshita, Keito and Zhou, Xiaokang and Kawai, Akira and Yada, Katsutoshi},
  journal = {arXiv preprint arXiv:2604.27345},
  year    = {2026},
  url     = {https://arxiv.org/abs/2604.27345}
}

@inproceedings{wong2021crossreplication,
  title     = {Cross-replication Reliability -- An Empirical Approach to Interpreting Inter-rater Reliability},
  author    = {Wong, Ka and Paritosh, Praveen and Aroyo, Lora},
  booktitle = {Proceedings of the 59th Annual Meeting of the Association for Computational Linguistics and the 11th International Joint Conference on Natural Language Processing (Volume 1: Long Papers)},
  pages     = {7053--7065},
  year      = {2021},
  doi       = {10.18653/v1/2021.acl-long.548},
  url       = {https://aclanthology.org/2021.acl-long.548/}
}

@article{openai2025gpt5,
  title   = {{OpenAI} {GPT-5} System Card},
  author  = {{OpenAI}},
  journal = {arXiv preprint arXiv:2601.03267},
  year    = {2025},
  url     = {https://arxiv.org/abs/2601.03267}
}

@misc{googledeepmind2026gemini37flash,
  title        = {{Gemini 3.7 Flash} Model Card},
  author       = {{Google DeepMind}},
  year         = {2026},
  month        = aug,
  howpublished = {Model card},
  url          = {https://deepmind.google/models/model-cards/gemini-3-7-flash/}
}

@article{geminiteam2025gemini25,
  title   = {{Gemini 2.5}: Pushing the Frontier with Advanced Reasoning, Multimodality, Long Context, and Next Generation Agentic Capabilities},
  author  = {{Gemini Team, Google}},
  journal = {arXiv preprint arXiv:2507.06261},
  year    = {2025},
  url     = {https://arxiv.org/abs/2507.06261}
}

@misc{qwen2026qwen35,
  title        = {{Qwen3.5}: Towards Native Multimodal Agents},
  author       = {{Qwen Team}},
  year         = {2026},
  month        = feb,
  howpublished = {Release note},
  url          = {https://qwen.ai/blog?id=qwen3.5}
}

@misc{qwen2026qwen3627b,
  title        = {{Qwen3.6-27B}: Flagship-Level Coding in a 27{B} Dense Model},
  author       = {{Qwen Team}},
  year         = {2026},
  month        = apr,
  howpublished = {Release note},
  url          = {https://qwen.ai/blog?id=qwen3.6-27b}
}

@misc{qwen2026qwen3635ba3b,
  title        = {{Qwen3.6-35B-A3B}: Agentic Coding Power, Now Open to All},
  author       = {{Qwen Team}},
  year         = {2026},
  month        = apr,
  howpublished = {Release note},
  url          = {https://qwen.ai/blog?id=qwen3.6-35b-a3b}
}

@article{qwen2026qwen35omni,
  title   = {{Qwen3.5-Omni} Technical Report},
  author  = {{Qwen Team}},
  journal = {arXiv preprint arXiv:2604.15804},
  year    = {2026},
  url     = {https://arxiv.org/abs/2604.15804}
}

@misc{moonshot2026kimik26,
  title        = {{Kimi K2.6}: Advancing Open-Source Coding},
  author       = {{Moonshot AI}},
  year         = {2026},
  month        = apr,
  howpublished = {Release note},
  url          = {https://www.kimi.ai/blog/kimi-k2-6}
}

@article{kimiteam2026k25,
  title   = {{Kimi K2.5}: Visual Agentic Intelligence},
  author  = {{Kimi Team}},
  journal = {arXiv preprint arXiv:2602.02276},
  year    = {2026},
  url     = {https://arxiv.org/abs/2602.02276}
}

@article{deepseekai2024deepseekv3,
  title   = {{DeepSeek-V3} Technical Report},
  author  = {{DeepSeek-AI}},
  journal = {arXiv preprint arXiv:2412.19437},
  year    = {2024},
  url     = {https://arxiv.org/abs/2412.19437}
}

@inproceedings{hu2022lora,
  title     = {{LoRA}: Low-Rank Adaptation of Large Language Models},
  author    = {Hu, Edward J. and Shen, Yelong and Wallis, Phillip and Allen-Zhu, Zeyuan and Li, Yuanzhi and Wang, Shean and Wang, Lu and Chen, Weizhu},
  booktitle = {International Conference on Learning Representations (ICLR)},
  year      = {2022},
  url       = {https://arxiv.org/abs/2106.09685}
}

@inproceedings{chu2025sft,
  title     = {{SFT} Memorizes, {RL} Generalizes: A Comparative Study of Foundation Model Post-training},
  author    = {Chu, Tianzhe and Zhai, Yuexiang and Yang, Jihan and Tong, Shengbang and Xie, Saining and Schuurmans, Dale and Le, Quoc V. and Levine, Sergey and Ma, Yi},
  booktitle = {Proceedings of the 42nd International Conference on Machine Learning (ICML)},
  year      = {2025},
  url       = {https://proceedings.mlr.press/v267/chu25c.html}
}

@article{yun2025priceofformat,
  title   = {The Price of Format: Diversity Collapse in {LLMs}},
  author  = {Yun, Longfei and An, Chenyang and Wang, Zilong and Peng, Letian and Shang, Jingbo},
  journal = {arXiv preprint arXiv:2505.18949},
  year    = {2025},
  url     = {https://arxiv.org/abs/2505.18949}
}

@article{karouzos2026collapse,
  title   = {Where does output diversity collapse in post-training?},
  author  = {Karouzos, Constantinos and Tan, Xingwei and Aletras, Nikolaos},
  journal = {arXiv preprint arXiv:2604.16027},
  year    = {2026},
  url     = {https://arxiv.org/abs/2604.16027}
}

@article{wang2026testtime,
  title   = {Test-time {RL} Alignment Exposes Task Familiarity Artifacts in {LLM} Benchmarks},
  author  = {Wang, Kun and Heckel, Reinhard},
  journal = {arXiv preprint arXiv:2603.12875},
  year    = {2026},
  url     = {https://arxiv.org/abs/2603.12875}
}

@book{krippendorff2004content,
  title     = {Content Analysis: An Introduction to Its Methodology},
  author    = {Krippendorff, Klaus},
  edition   = {2},
  publisher = {Sage},
  address   = {Thousand Oaks, CA},
  year      = {2004},
  isbn      = {9780761915454}
}

@misc{krippendorff2011computing,
  title  = {Computing {Krippendorff's} Alpha-Reliability},
  author = {Krippendorff, Klaus},
  year   = {2011},
  note   = {Departmental Papers (ASC), Annenberg School for Communication, University of Pennsylvania},
  url    = {https://repository.upenn.edu/asc_papers/43/}
}

@article{wilson1927interval,
  title   = {Probable Inference, the Law of Succession, and Statistical Inference},
  author  = {Wilson, Edwin B.},
  journal = {Journal of the American Statistical Association},
  volume  = {22},
  number  = {158},
  pages   = {209--212},
  year    = {1927},
  doi     = {10.1080/01621459.1927.10502953}
}

@article{mcnemar1947test,
  title   = {Note on the Sampling Error of the Difference Between Correlated Proportions or Percentages},
  author  = {McNemar, Quinn},
  journal = {Psychometrika},
  volume  = {12},
  number  = {2},
  pages   = {153--157},
  year    = {1947},
  doi     = {10.1007/BF02295996}
}

@article{wilcoxon1945individual,
  title   = {Individual Comparisons by Ranking Methods},
  author  = {Wilcoxon, Frank},
  journal = {Biometrics Bulletin},
  volume  = {1},
  number  = {6},
  pages   = {80--83},
  year    = {1945},
  doi     = {10.2307/3001968}
}

@article{efron1979bootstrap,
  title   = {Bootstrap Methods: Another Look at the Jackknife},
  author  = {Efron, B.},
  journal = {The Annals of Statistics},
  volume  = {7},
  number  = {1},
  pages   = {1--26},
  year    = {1979},
  doi     = {10.1214/aos/1176344552}
}

@article{spearman1904proof,
  title   = {The Proof and Measurement of Association between Two Things},
  author  = {Spearman, C.},
  journal = {The American Journal of Psychology},
  volume  = {15},
  number  = {1},
  pages   = {72--101},
  year    = {1904},
  doi     = {10.2307/1412159}
}

@inproceedings{hyun2024smile,
  title     = {{SMILE}: Multimodal Dataset for Understanding Laughter in Video with Language Models},
  author    = {Hyun, Lee and Sung-Bin, Kim and Han, Seungju and Yu, Youngjae and Oh, Tae-Hyun},
  booktitle = {Findings of the Association for Computational Linguistics: NAACL 2024},
  pages     = {1149--1167},
  year      = {2024},
  url       = {https://aclanthology.org/2024.findings-naacl.73/}
}

@inproceedings{lee2026smilenext,
  title     = {{SMILE-Next}: Teaching Large Language Models to Detect, Classify, and Reason about Laughter},
  author    = {Jung-Mok, Lee and Sung-Bin, Kim and Chang, Joohyun and Hyun, Lee and Oh, Tae-Hyun},
  booktitle = {Proceedings of the 64th Annual Meeting of the Association for Computational Linguistics (ACL)},
  year      = {2026},
  url       = {https://arxiv.org/abs/2605.28084}
}

@inproceedings{chang2024libreface,
  title     = {{LibreFace}: An Open-Source Toolkit for Deep Facial Expression Analysis},
  author    = {Chang, Di and Yin, Yufeng and Li, Zongjian and Tran, Minh and Soleymani, Mohammad},
  booktitle = {Proceedings of the IEEE/CVF Winter Conference on Applications of Computer Vision (WACV)},
  year      = {2024},
  url       = {https://arxiv.org/abs/2308.10713}
}

@article{lian2024affectgpt,
  title   = {{AffectGPT}: Dataset and Framework for Explainable Multimodal Emotion Recognition},
  author  = {Lian, Zheng and Sun, Haiyang and Sun, Licai and Yi, Jiangyan and Liu, Bin and Tao, Jianhua},
  journal = {arXiv preprint arXiv:2407.07653},
  year    = {2024},
  url     = {https://arxiv.org/abs/2407.07653}
}

@inproceedings{zhang2025seriesbench,
  title     = {{SeriesBench}: A Benchmark for Narrative-Driven Drama Series Understanding},
  author    = {Zhang, Chenkai and Lei, Yiming and Liu, Zeming and Leng, Haitao and Liu, ShaoGuo and Gao, Tingting and Liu, Qingjie and Wang, Yunhong},
  booktitle = {Proceedings of the IEEE/CVF Conference on Computer Vision and Pattern Recognition (CVPR)},
  pages     = {28995--29004},
  year      = {2025},
  url       = {https://openaccess.thecvf.com/content/CVPR2025/html/Zhang_SeriesBench_A_Benchmark_for_Narrative-Driven_Drama_Series_Understanding_CVPR_2025_paper.html}
}

@article{soleymani2012mahnob,
  title   = {A Multimodal Database for Affect Recognition and Implicit Tagging},
  author  = {Soleymani, Mohammad and Lichtenauer, Jeroen and Pun, Thierry and Pantic, Maja},
  journal = {IEEE Transactions on Affective Computing},
  volume  = {3},
  number  = {1},
  pages   = {42--55},
  year    = {2012},
  doi     = {10.1109/T-AFFC.2011.25}
}

@inproceedings{amiriparian2022muse,
  title     = {{MuSe} 2022 Challenge: Multimodal Humour, Emotional Reactions, and Stress},
  author    = {Amiriparian, Shahin and Christ, Lukas and K{\"o}nig, Andreas and Me{\ss}ner, Eva-Maria and Cowen, Alan and Cambria, Erik and Schuller, Bj{\"o}rn W.},
  booktitle = {Proceedings of the 30th ACM International Conference on Multimedia},
  year      = {2022},
  doi       = {10.1145/3503161.3551792}
}

@inproceedings{li2025mimeqa,
  title     = {{MimeQA}: Towards Socially-Intelligent Nonverbal Foundation Models},
  author    = {Li, Hengzhi and Tjandrasuwita, Megan and Fung, Yi R. and Solar-Lezama, Armando and Liang, Paul Pu},
  booktitle = {Advances in Neural Information Processing Systems},
  year      = {2025},
  note      = {Datasets and Benchmarks Track},
  url       = {https://arxiv.org/abs/2502.16671}
}

@inproceedings{ong2026behavioratlas,
  title     = {Human Behavior Atlas: Benchmarking Unified Psychological and Social Behavior Understanding},
  author    = {Ong, Keane and Dai, Wei and Li, Carol and Feng, Dewei and Li, Hengzhi and Wu, Jingyao and Cheong, Jiaee and Mao, Rui and Mengaldo, Gianmarco and Cambria, Erik and Liang, Paul Pu},
  booktitle = {International Conference on Learning Representations (ICLR)},
  year      = {2026},
  url       = {https://arxiv.org/abs/2510.04899}
}

@inproceedings{yu2025vrbench,
  title     = {{VRBench}: A Benchmark for Multi-Step Reasoning in Long Narrative Videos},
  author    = {Yu, Jiashuo and Wu, Yue and Chu, Meng and Ren, Zhifei and Huang, Zizheng and Chu, Pei and Zhang, Ruijie and He, Yinan and Li, Qirui and Li, Songze and others},
  booktitle = {Proceedings of the IEEE/CVF International Conference on Computer Vision (ICCV)},
  year      = {2025},
  url       = {https://arxiv.org/abs/2506.10857}
}

@article{hu2025videommmu,
  title   = {{Video-MMMU}: Evaluating Knowledge Acquisition from Multi-Discipline Professional Videos},
  author  = {Hu, Kairui and Wu, Penghao and Pu, Fanyi and Xiao, Wang and Zhang, Yuanhan and Yue, Xiang and Li, Bo and Liu, Ziwei},
  journal = {arXiv preprint arXiv:2501.13826},
  year    = {2025},
  url     = {https://arxiv.org/abs/2501.13826}
}

@inproceedings{hong2025motionbench,
  title     = {{MotionBench}: Benchmarking and Improving Fine-grained Video Motion Understanding for Vision Language Models},
  author    = {Hong, Wenyi and Cheng, Yean and Yang, Zhuoyi and Wang, Weihan and Wang, Lefan and Gu, Xiaotao and Huang, Shiyu and Dong, Yuxiao and Tang, Jie},
  booktitle = {Proceedings of the IEEE/CVF Conference on Computer Vision and Pattern Recognition (CVPR)},
  year      = {2025},
  url       = {https://arxiv.org/abs/2501.02955}
}

@article{zhang2026dar,
  title   = {Benchmarking Dynamic Affective Reasoning: A Viewer-Centric Video Emotion Dataset},
  author  = {Zhang, Zhiyan and Song, Peipei and Hu, Jinpeng and Jia, Jingyang and Yang, Xun and Chang, Xiaojun},
  journal = {arXiv preprint arXiv:2607.10238},
  year    = {2026},
  url     = {https://arxiv.org/abs/2607.10238}
}

@inproceedings{mazeika2022vce,
  title     = {How Would The Viewer Feel? Estimating Wellbeing From Video Scenarios},
  author    = {Mazeika, Mantas and Tang, Eric and Zou, Andy and Basart, Steven and Chan, Jun Shern and Song, Dawn and Forsyth, David and Steinhardt, Jacob and Hendrycks, Dan},
  booktitle = {Advances in Neural Information Processing Systems},
  year      = {2022},
  url       = {https://arxiv.org/abs/2210.10039}
}

\clearpage
\appendix
\section*{Appendix}

\section{Source Video Collection, Provenance, and Preprocessing}
\label{app:source}

This appendix traces the source videos, candidate metadata, human annotations, and final task instances. Supervision comes from human intensity ratings, human-written rationale references, and temporal alignment for SRM.

\subsection{Source Acquisition and Collection Scope}

The preprocessing pipeline yields 31,834 synchronized stimulus--reaction clips totaling 433:42:49 hours across eight broad domains: games, music, sports, live streams, movies and television, expert commentary, general pop culture, and comedy. Source discovery begins from manually maintained creator/channel and playlist manifests. The collection pipeline inventories platform metadata and available subtitles, downloads source media, and records canonical identifiers so that derived artifacts can be joined without using filenames as identity. Processing is organized by content domain because reaction layouts and usable viewing intervals vary across creator formats.

Each usable recording contains both the content being watched and the creator's response, commonly in a picture-in-picture layout. Domain-specific preprocessing identifies viewing intervals, separates the stimulus and reactor regions, and retains their temporal alignment. Automated quality checks flag frozen or degraded stimulus regions, missing or incomplete faces, invalid crops, and synchronization failures. The pipeline is designed to be deterministic, resumable, and auditable: completed artifacts are not silently overwritten, and failures are retained in logs for review.

Automated preprocessing also produces face and action-unit traces, separated audio, scene descriptions, and other derived metadata. These fields support collection-level inspection but are \emph{auxiliary metadata}, not gold labels. Benchmark supervision is restricted to human intensity ratings and human-written rationale references, plus the disclosed temporal-alignment answer used to construct SRM.

\begin{center}
  \begin{minipage}{\textwidth}
  \centering
  \captionof{table}{Source-to-benchmark accounting. Units differ across stages and are not a single one-to-one funnel: a video can produce several nominated ranges, ratings, or references. Automated artifacts nominate or describe candidates but do not supply human gold labels.}
  \label{tab:appendix-funnel}
  \small
  \setlength{\tabcolsep}{5pt}
  \begin{tabular}{p{0.20\textwidth}p{0.17\textwidth}p{0.17\textwidth}p{0.36\textwidth}}
    \toprule
    Stage & Videos / items & Human evidence & Benchmark role \\
    \midrule
    Synchronized clip collection & 31,834 synchronized clips; 433:42:49 h & None required & Candidate pool and provenance for task construction. \\
    Non-pilot pre-screen & 14,582 clips; 21,574 nominated ranges & 50-clip blinded pilot for selector choice & Reduces annotation load. Model scores, ranks, and explanations are excluded from benchmark prompts and consensus. \\
    Human intensity annotations & 1,999 videos; 3,257 moments; 14,894 ratings & At least three ratings per moment from eight raters & Supplies the 0--3 intensity evidence used for consensus, disagreement analysis, and task selection. \\
    IP set & 250 videos; 642 moments & Human intensity ratings and leave-one-rater-out reference & Frozen joint and independent intensity evaluation on identical moments. \\
    SRM set & 112 items; 97 clips & Temporal alignment under the audited separation contract & Frozen four-way evaluation set. A separate 150-item pool supports protocol analyses only. \\
    GRG set & 200 items; 447 accepted references & Blind-written references from three completed annotators & Supports rationale generation and rubric-based judging; an 84-pair study assesses item-level judge--human rubric association. \\
    \bottomrule
  \end{tabular}
  \end{minipage}
\end{center}

\subsection{Candidate Nomination and the Leakage Boundary}
\label{app:nomination}

The pre-screen is used only to propose ranges likely to contain observable reaction changes. In a source-hidden 50-clip pilot, a human annotator scored the merged nominations with the same 0--3 anchors later used for human intensity annotation. Among nominations from Gemini 3.6 Flash, 92.0\% received a human score of at least 2, compared with 68.0\% for the facial-analysis selector, which uses LibreFace action-unit and expression outputs~\citep{chang2024libreface}. Gemini 3.6 Flash was therefore applied to 14,582 non-pilot clips and produced 21,574 candidate ranges.

Selection changes the target distribution: the nomination prompt favors strong visible or audible responses and rejects clips without a proposed score of at least 2. Excluding model-generated scores and explanations from human labeling prevents direct label leakage, but does not remove this selection effect or possible selector-family dependence. The evaluated sets represent selected creator reactions, not ordinary viewing behavior or reactions the selector missed.

The selected pre-screen used Gemini 3.6 Flash with audio, 1~FPS sampling, low media resolution, temperature zero, 2--4 second windows, and at most five nominations per clip. The 50-clip pilot used the same prompt template with at most two nominations.

\begin{promptbox}[prompt:prescreen]{Model-assisted candidate nomination}
You are screening a reaction video for the single strongest moments of reviewer reaction.

The clip is \textless{}duration\textgreater{} seconds long. You can see the reviewer and hear the audio.

Rate reaction strength on this behaviourally anchored scale:

0 = calm viewing: no visible or audible reaction.

1 = mild: slight smile, nod, small head movement.

2 = clear: audible laugh, exclamation, raised brows with a forward lean.

3 = peak: scream, jump, hands to face, recoiling backwards.

Nominate at most five moments, strongest first. Each nominated moment must be between 2 and 4 seconds long and the moments must not overlap. Give start and end as seconds from the start of the clip, as decimal numbers.

For each nomination give the anchored score (0--3) and one sentence of evidence that cites what you actually observed. Name the visual cue, the audio cue, or both.

If no moment in this clip reaches 2 on the scale, return an empty list. Do not pad the list to reach five; nominating a weak moment is worse than nominating nothing.
\end{promptbox}

The API response schema requires a \texttt{nominations} list with \texttt{start\_s}, \texttt{end\_s}, \texttt{score}, and \texttt{evidence} for each proposed range.

\Needspace{0.45\textheight}
\section{Human Annotation and Gold-Set Construction}
\label{app:annotation}

\subsection{Evidence Roles}

Model-based pre-screening is used only for candidate nomination and never supplies gold scores, consensus labels, rationale references, or evaluated inputs. Human raters provide intensity supervision; temporal alignment defines SRM answers but is not treated as a semantic-attribution judgment; and blind-written human rationales provide situated GRG references rather than universal causal explanations. Model judges are used only for rubric-based scoring and judge-sensitivity analysis, not as human gold or an absolute measure of explanation quality.

% \begin{center}
%   \begin{minipage}{\textwidth}
%   \centering
%   \captionof{table}{Evidence-role ledger. The benchmark explicitly prevents selection tools, temporal metadata, and model judges from being described as interchangeable forms of human gold.}
%   \label{tab:evidence-roles}
%   \small
%   \setlength{\tabcolsep}{5pt}
%   \begin{tabular}{p{0.17\textwidth}p{0.23\textwidth}p{0.23\textwidth}p{0.22\textwidth}}
%     \toprule
%     Evidence source & Permitted role & Excluded role & Affected tasks \\
%     \midrule
%     Pre-screen model & Candidate nomination and annotation triage & Gold score, consensus, rationale reference, or evaluated input & IP--GRG selection only \\
%     Intensity raters & Anchored intensity scores and rater-level uncertainty & Psychological diagnosis or demographic ground truth & IP; selection for SRM/GRG \\
%     Temporal alignment & Defines the candidate window aligned with the reaction & Semantic-attribution judgment & SRM construction \\
%     Rationale annotators & Blind-written, situated rationale references and opt-out signals & Universal causal explanation or familiarity-independent gold & GRG \\
%     Model judges & Rubric scoring and judge-sensitivity diagnostics & Human gold or an absolute measure of explanation utility & GRG \\
%     \bottomrule
%   \end{tabular}
%   \end{minipage}
% \end{center}

\subsection{Human Annotation of Reaction Intensity}
\label{app:intensity-annotation}

The interface shows the complete reaction clip and marks every candidate range on a shared timeline (Figure~\ref{fig:ui-intensity}; the read-only monitoring dashboard appears in Figure~\ref{fig:ui-admin-scoring}). Raters score all marked moments, including weak or absent reactions, relative to the same viewer's surrounding behavior. The following anchors remain visible during annotation:

\begin{center}
  \begin{minipage}{\textwidth}
  \centering
  \captionof{table}{The operational intensity rubric.}
  \label{tab:intensity-anchors}
  \small
  \begin{tabular}{clp{0.50\columnwidth}}
    \toprule
    Score & Name & Behavioral anchor \\
    \midrule
    0 & Calm & No visible or audible reaction. \\
    1 & Mild & Slight smile, nod, or small head movement. \\
    2 & Clear & Audible laugh or exclamation, or raised brows with a forward lean. \\
    3 & Peak & Scream, jump, hands to face, or recoiling backwards. \\
    \bottomrule
  \end{tabular}
  \end{minipage}
\end{center}

The human intensity annotation pool contains 14,894 ratings over 3,257 moments from 1,999 videos, with at least three scores per moment and eight raters overall.

\begin{center}
  \begin{minipage}{\textwidth}
  \centering
  \captionof{table}{Human intensity agreement diagnostics.}
  \label{tab:intensity-agreement}
  \small
  \begin{tabular}{lr}
    \toprule
    Quantity & Current value \\
    \midrule
    Ordinal Krippendorff's $\alpha$~\citep{krippendorff2004content,krippendorff2011computing} & .357 \\
    Nominal Krippendorff's $\alpha$ & .235 \\
    Item-level unanimity (all raters agree) & .253 \\
    Exact agreement over 27,455 rater pairs & .571 \\
Independent two-draw agreement from pooled marginals & .446 \\
    Unanimous pass/fail decisions & .417 \\
    0--1 share of one-step disagreements & 6.97\% \\
    1--2 share of one-step disagreements & 68.91\% \\
    2--3 share of one-step disagreements & 24.12\% \\
    \bottomrule
  \end{tabular}
  \end{minipage}
\end{center}

Item-level unanimity weights each of the 3,257 moments equally. Pairwise exact agreement weights each of the 27,455 within-moment unordered rater pairs equally, so items with more raters contribute more pairs. The independent two-draw quantity uses pooled rating marginals and is not a chance baseline for item unanimity. Recomputing nominal and ordinal Krippendorff's alpha from the coincidence matrix reproduces 0.235 and 0.357, respectively.

Consensus is the raw mean of all available rater scores for a moment. As a robustness check, we fit an additive rater--item model and subtract estimated rater effects before recomputing consensus. Estimated effects range from $-0.300$ to $0.214$; the mean absolute change in consensus is 0.045 and the maximum is 0.057. In the 250-video IP set, debiasing changes 0 of 250 within-video orders and leaves every model correlation unchanged to four decimals. We therefore retain raw means for interpretability.

\subsection{Evaluation-Set Composition}

Tables~\ref{tab:eval-inventory} and~\ref{tab:creator-inventory} describe each frozen test set separately. We join SRM and GRG candidate IDs to the normalized construction metadata and retain the frozen IP inventory. SRM contains seven creator personas from five channels; GRG contains eight from seven channels. Their Patreon-source shares are 72/112 and 61/200 items, compared with 200/250 IP videos. Category labels describe the collection inventory: Patreon is an acquisition source, while the other labels name broad content groups. The item counts therefore describe the evaluated sample without assigning a genre to Patreon material.

% Generated by scripts/complete_review_checks.py.
\begin{center}
\begin{minipage}{\textwidth}
\centering
\captionof{table}{Composition of the frozen evaluation sets. SRM/GRG items are joined by candidate ID to normalized construction metadata; IP uses its frozen task inventory. Evaluation units are videos for IP and reaction moments for SRM/GRG; category and creator counts weight those units equally. Same-channel personas are merged for channel counts. Inventory categories mix content labels with the Patreon acquisition source, so they are not a uniform genre taxonomy.}
\label{tab:eval-inventory}
\small
\setlength{\tabcolsep}{3pt}
\begin{tabular*}{\textwidth}{@{\extracolsep{\fill}}lrrr@{}}
\toprule
Inventory & IP & SRM & GRG \\
\midrule
Evaluation units & 250 & 112 & 200 \\
Distinct clips & 250 & 97 & 185 \\
Creator personas & 7 & 7 & 8 \\
Channels & 5 & 5 & 7 \\
Comedy / internet culture & 10 & 2 & 62 \\
General pop culture & 24 & 7 & 40 \\
Games & 16 & 31 & 37 \\
Patreon (acquisition source) & 200 & 72 & 61 \\
\bottomrule
\end{tabular*}
\end{minipage}
\end{center}

% Generated by scripts/complete_review_checks.py.
\begin{center}
\begin{minipage}{\textwidth}
\centering
\captionof{table}{Per-creator evaluation counts, with consistent anonymous persona identifiers across tasks. IP counts videos; SRM and GRG count reaction moments. A zero denotes no retained items from that persona in the task.}
\label{tab:creator-inventory}
\small
\setlength{\tabcolsep}{3pt}
\begin{tabular*}{\textwidth}{@{\extracolsep{\fill}}lrrr@{}}
\toprule
Persona & IP & SRM & GRG \\
\midrule
P1 & 50 & 4 & 0 \\
P2 & 50 & 9 & 3 \\
P3 & 16 & 31 & 37 \\
P4 & 50 & 32 & 38 \\
P5 & 50 & 27 & 20 \\
P6 & 9 & 1 & 47 \\
P7 & 0 & 0 & 6 \\
P8 & 0 & 0 & 1 \\
P9 & 25 & 8 & 48 \\
\bottomrule
\end{tabular*}
\end{minipage}
\end{center}

\subsection{IP Selection and Human Reference}

A video is eligible for IP if it contains at least two rated moments, its consensus spread is at least 0.25, and every selected moment has at least three ratings. The rankable pool contains 572 videos. We select 250 by source-category-proportional systematic sampling with a per-creator cap of 50 videos; the exact composition is disclosed below.

\begin{center}
  \begin{minipage}{\textwidth}
  \centering
  \captionof{table}{Composition of the frozen IP evaluation set. The human value is a pool-specific reference computed by comparing each available rater with the remaining-rater consensus within the same video.}
  \label{tab:t1-inventory}
  \small
  \begin{tabular}{lr}
    \toprule
    IP inventory & Count \\
    \midrule
    Videos / moments & 250 / 642 \\
    Videos with 2 / 3 / 4 / 5 moments & 145 / 76 / 21 / 8 \\
    Comedy / general-pop / game / Patreon & 10 / 24 / 16 / 200 \\
    Four most frequent creators & 50 each \\
    Remaining creator counts & 25 / 16 / 9 \\
    Nonconstant leave-one-rater-out pairs & 420 \\
    Conditional nonconstant LOO Spearman $\rho$ & .649 \\
Videos contributing to that conditional statistic & 166 \\
Constant-inclusive matched LOO units / videos & 956 / 250 \\
    \bottomrule
  \end{tabular}
  \end{minipage}
\end{center}

Tables~\ref{tab:eval-inventory} and~\ref{tab:creator-inventory} give the task distributions and per-creator counts. The seven IP personas belong to five channels. The 0.649 statistic conditions on both rater and remaining consensus being nonconstant; it is not the all-video human baseline. Its 420 units cover 166 videos. With all eligible constant human profiles retained, there are 956 LOO units over all 250 videos; the matched comparison is defined in Appendix~\ref{app:human-matched}.

\subsection{SRM Construction and Separation Contract}
\label{app:srm-construction}

The frozen SRM set follows a deterministic construction contract:
\begin{enumerate}[leftmargin=*,itemsep=2pt]
  \item The correct option is the stimulus segment nearest the reaction's temporally aligned moment;
  \item The three distractors come from the same source video;
  \item Every pair of option start times differs by at least a required gap, taken from the ladder 10, 8, 6 seconds: the builder tries 10 first and relaxes only when no full four-option item exists at that level, recording the effective gap per item;
  \item Description-token Jaccard similarity is below 0.5 for every selected option pair, so options must also be describably distinct and not only temporally apart;
  \item Distractors are selected greedily farthest-first to spread them across the source timeline;
  \item Option clips are capped at 6 seconds and must hold at least 2 seconds of real video, verified by media probing;
  \item Candidate order is shuffled deterministically from the item seed; and
  \item The reaction view spans two seconds before through one second after the annotated range.
\end{enumerate}

\paragraph{Collision Audit on the Frozen Set.}
The frozen set contains 112 items over 97 clips. Recomputing the minimum pairwise start gap directly from the task file gives a range of 6.0--18.1 seconds with a median of 11.2 and no item below 6.0. By that audit, 68 items separate every candidate pair by at least 10 seconds, 27 by 8 to 10 seconds, and 17 by 6 to 8 seconds, so the relaxed rungs of the ladder are used for a minority of items. Correct-answer positions are distributed 32/21/23/36 over A/B/C/D. Seventy items use the original stimulus track and 42 use a re-cut window. A human review of all 112 items checked distinguishability after replaying candidates against the reaction one or two times each. Blind agreement, trigger uniqueness, and reaction necessity remain unvalidated; delayed reactions and multiple plausible attributions are possible.

Model accuracies on this set are measured against the temporal-alignment answer. In the cross-video control (Appendix~\ref{app:xv}), the native-video model gains $+.232$ when distractors come from other videos. The result supports same-video candidates as a way to reduce source-level audio mismatch, but does not show that the within-video setting is free of temporal audio matching.

\subsection{GRG Blind Rationale Collection}

For each item, the interface presents the synchronized stimulus with audio and a muted reaction view over the same interval. The text box starts blank. Annotators are asked: ``Why did the reviewer react? Write ONE sentence naming the cue in the stimulus.'' The instructions emphasize a specific event, action, line, or sound rather than the reactor's facial expression. ``I'm not familiar with this content'' and ``Stimulus is unclear / mis-timed'' are kept as distinct opt-outs. Figure~\ref{fig:ui-rationale} shows the collection interface and Figure~\ref{fig:ui-admin-rationale} its monitoring panel.

Three annotators completed all 200 items; two others stopped after four items and their abandoned work is not treated as a completed panel. One completed annotator opted out on 51\% of items. Of that annotator's 54 unclear judgments, 31 still received references from both other annotators, showing why familiarity and stimulus clarity should remain situated annotations rather than automatic item deletion. Quality review yields 447 accepted references over 200 items, with 41/77/78/2/2 items containing 1/2/3/4/5 references.

\subsection{Human Anchor for Rationale Judging}

The validation study contains 84 anonymized model--item pairs: seven generators by three judge-score strata by four items per stratum. Three Prolific raters who did not author the references graded each explanation using the cue/link/reference-consistency rubric, with references visible and generator identity hidden. The interface also offered the stimulus and reaction videos (Figure~\ref{fig:ui-anchor-rubric}). The model judge is text-only. Spearman correlation with the human mean is 0.593; judge--individual and inter-rater mean correlations are 0.447 and 0.441. These item-level associations apply to judge-stratified sampling with shared references. Population calibration, grader equivalence, and generator ordering require separate evidence.

\subsection{Annotation and Monitoring Interfaces}
\label{app:interfaces}

Figures~\ref{fig:ui-intensity}--\ref{fig:ui-anchor-rubric} show the web interfaces behind the three human-annotation stages and the monitoring dashboards used to track them. Annotation is keyboard-driven with mandatory playback of every marked moment. The monitoring views are read-only: they display coverage, score distributions, agreement diagnostics, and low-effort flags, and cannot edit any rating.

\begin{center}
  \begin{minipage}{\textwidth}
  \centering
  \includegraphics[width=\textwidth]{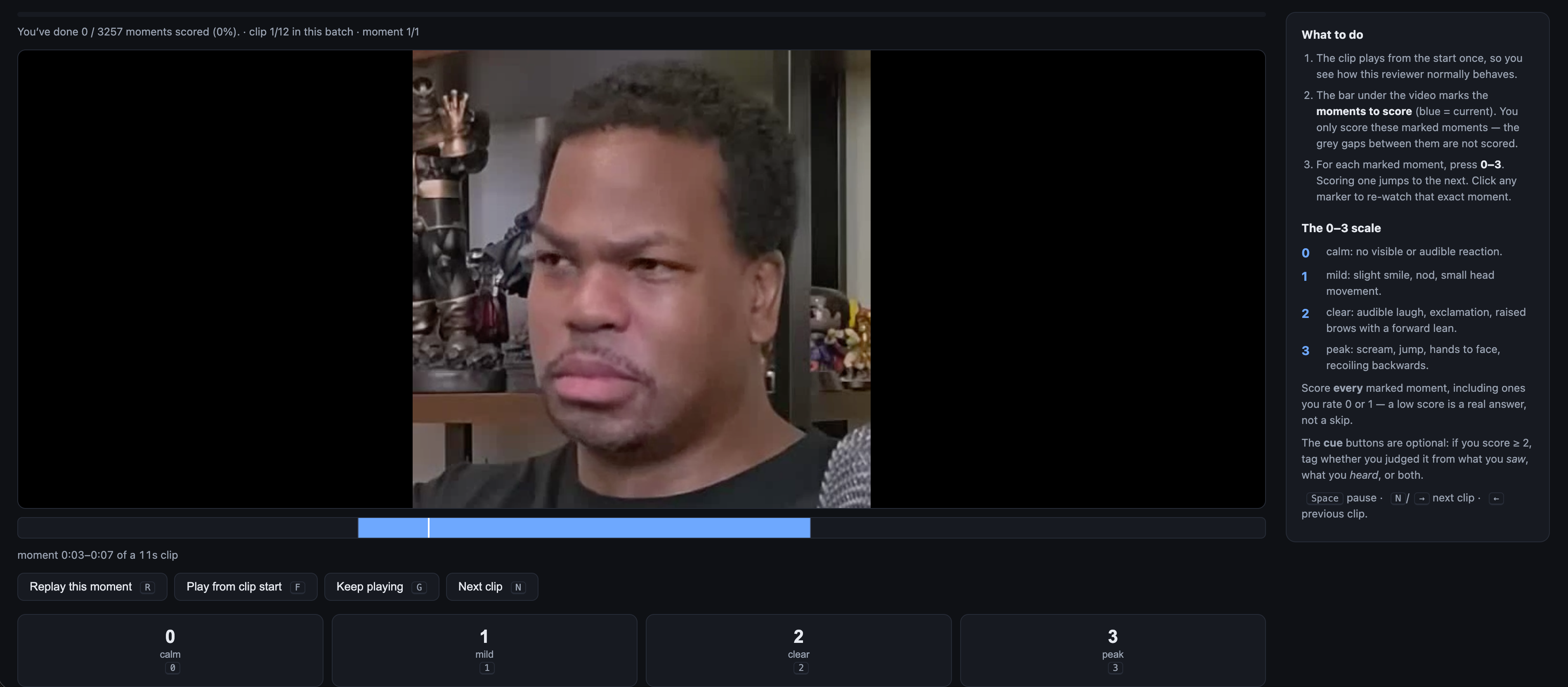}
  \captionof{figure}{Intensity-annotation interface. The complete reaction clip plays with every candidate moment marked on the shared timeline (blue = current); the behaviorally anchored 0--3 scale of Table~\ref{tab:intensity-anchors} remains visible at right, and raters must score every marked moment.}
  \label{fig:ui-intensity}
  \end{minipage}
\end{center}

\begin{center}
  \begin{minipage}{\textwidth}
  \centering
  \includegraphics[width=\textwidth]{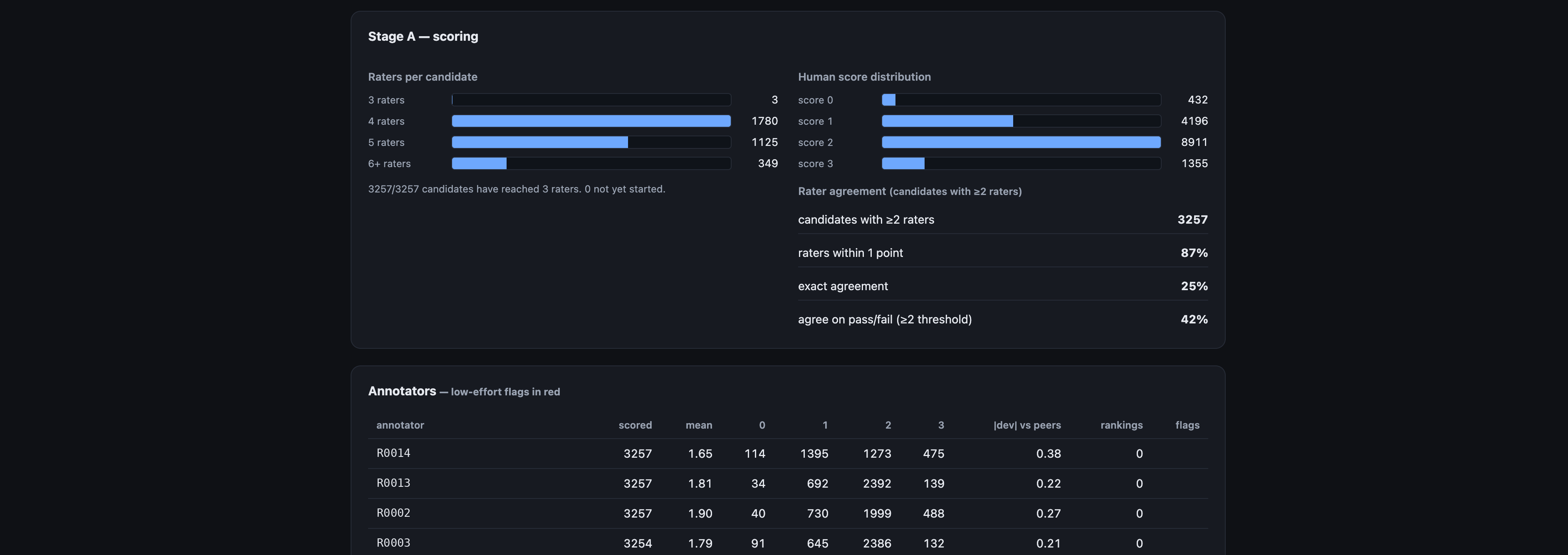}
  \captionof{figure}{Read-only monitoring dashboard for intensity annotation: rater counts per candidate, the pooled 0--3 score distribution, agreement diagnostics, and a per-annotator screen with low-effort flags.}
  \label{fig:ui-admin-scoring}
  \end{minipage}
\end{center}

\begin{center}
  \begin{minipage}{\textwidth}
  \centering
  \includegraphics[width=\textwidth]{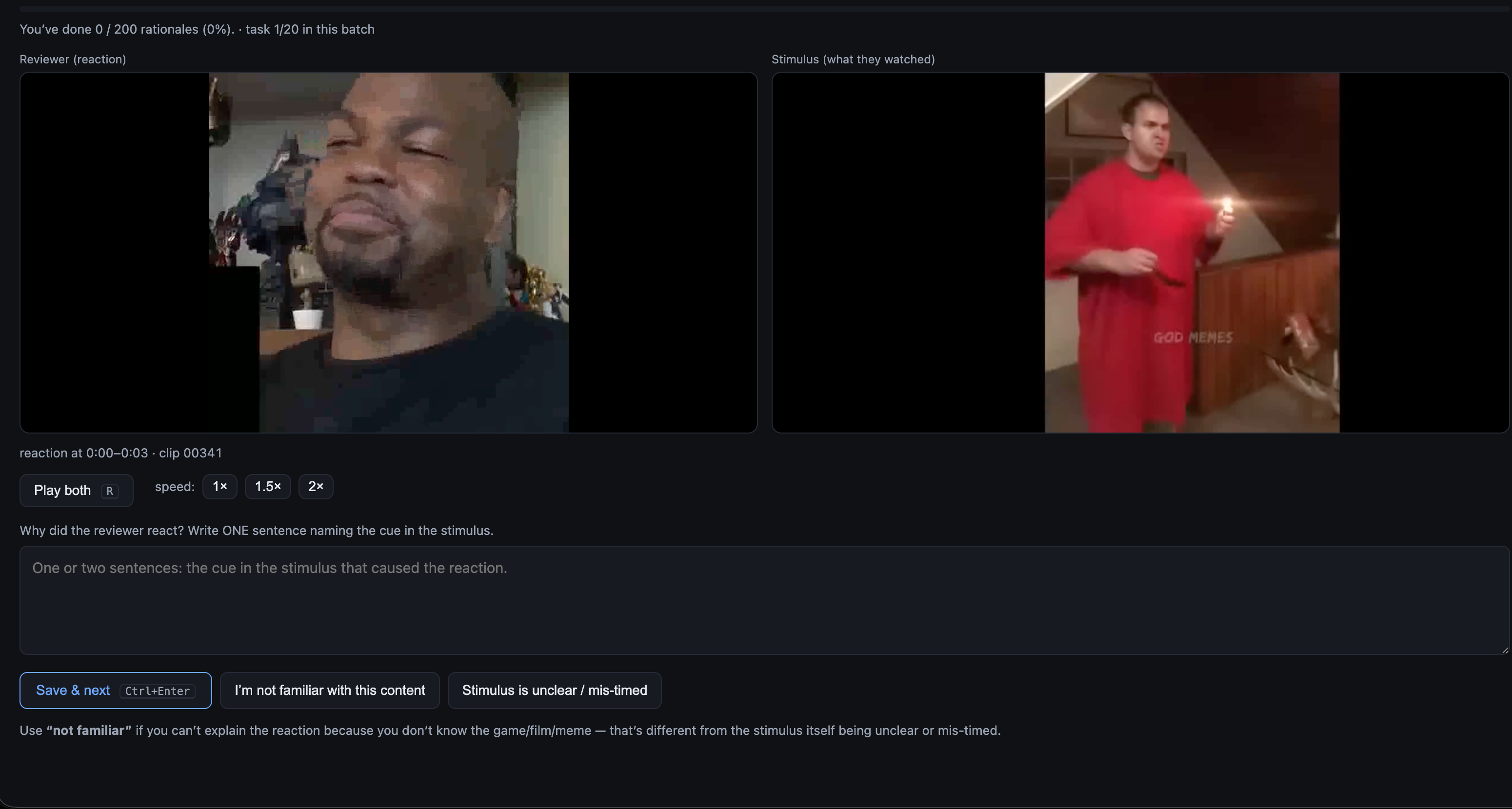}
  \captionof{figure}{Rationale-collection interface (GRG). The reviewer reaction and the stimulus play side by side over the same interval; the annotator answers ``Why did the reviewer react?'' in one sentence naming the cue in the stimulus, with ``not familiar'' and ``stimulus unclear'' recorded as distinct opt-outs.}
  \label{fig:ui-rationale}
  \end{minipage}
\end{center}

\begin{center}
  \begin{minipage}{\textwidth}
  \centering
  \includegraphics[width=\textwidth]{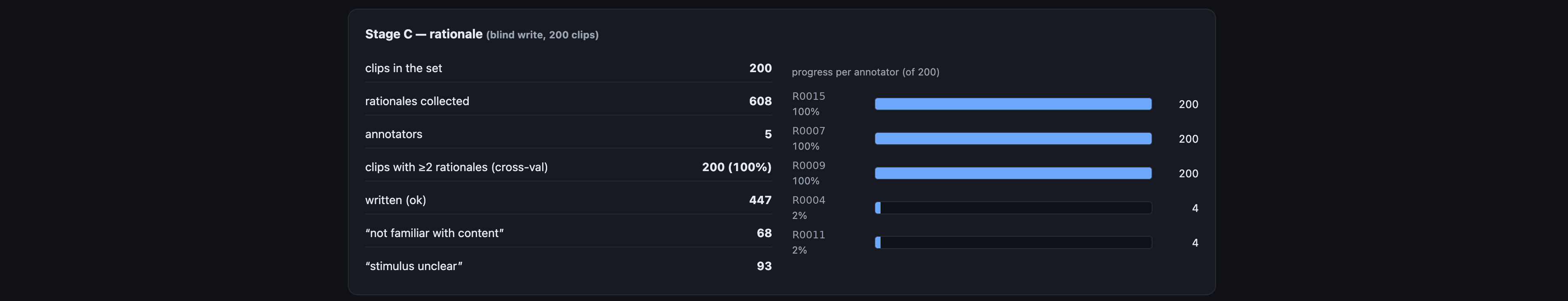}
  \captionof{figure}{Monitoring panel for rationale collection: clip coverage, collected and accepted reference counts, opt-out counts, and per-annotator progress. The two annotators who stopped after four items are visible; their work is excluded from the completed panel.}
  \label{fig:ui-admin-rationale}
  \end{minipage}
\end{center}

\begin{center}
  \begin{minipage}{\textwidth}
  \centering
  \includegraphics[width=\textwidth]{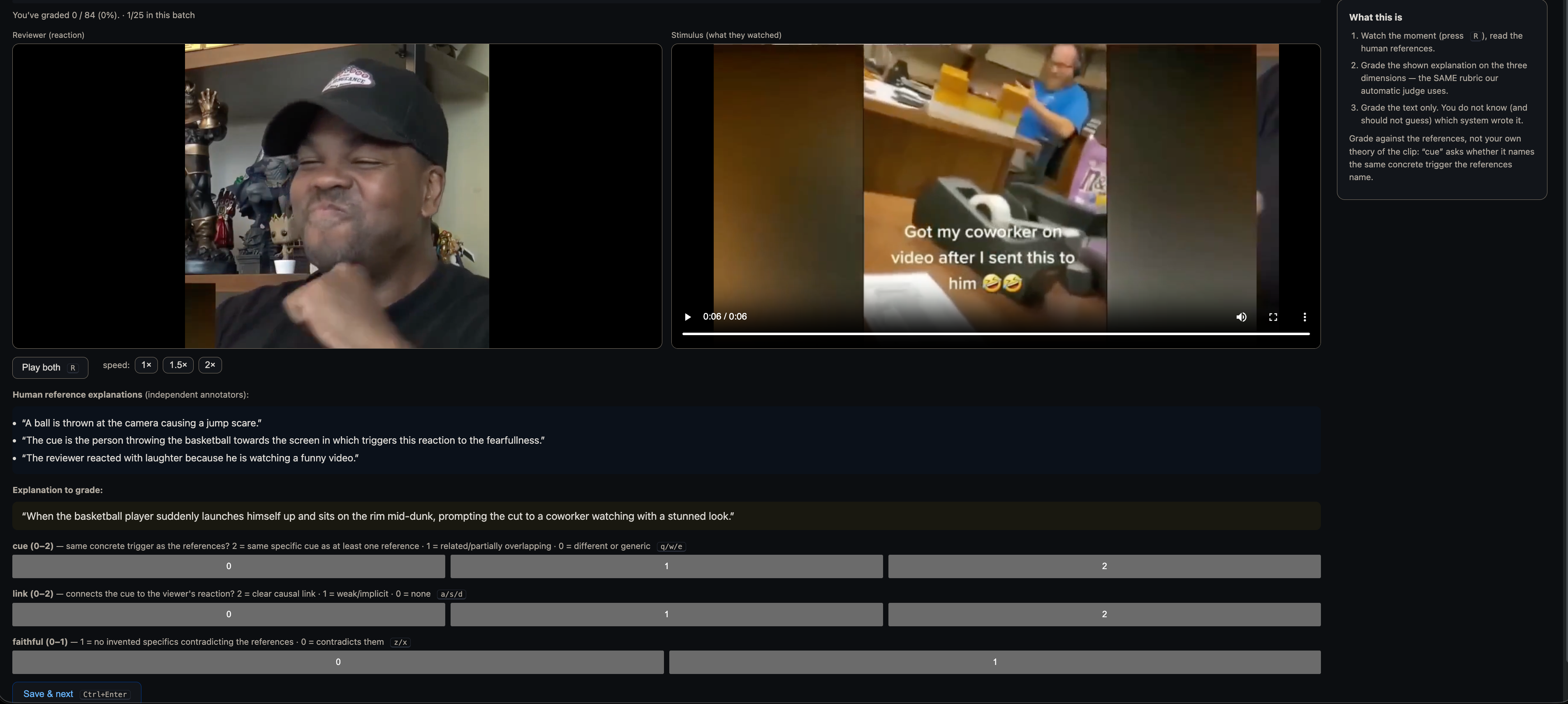}
  \captionof{figure}{Grading interface for the 84-item judge-anchor study. The rater watches the reaction and stimulus, reads the accepted human references, and grades one anonymized explanation on the same cue/link/faithfulness rubric as the model judge; the instructions state that the rater does not know, and should not guess, which system wrote the explanation.}
  \label{fig:ui-anchor-rubric}
  \end{minipage}
\end{center}

\section{Exact Task and Evaluation Protocols}
\label{app:prompts}

This section records the protocols that produced the reported model outputs. The legacy frame-time instruction is retained verbatim for reproducibility; its correction is documented in Appendix~\ref{app:timestamp-audit}. Prompt text is reproduced verbatim except that dynamic item fields are represented by angle-bracket placeholders. The media labels and answer suffixes are separate content parts in the API request, in the order shown below.

\subsection{Common Visual Input and Decoding}

Frames are extracted deterministically at 1~FPS, resized so the longer side is at most 512 pixels, and encoded as JPEG at the pipeline's quality setting 4. If a clip exceeds its frame budget, frames are sampled evenly across the available sequence. A degenerate sub-second window falls back to its first available frame. The sampled-frame protocol is the common visual input for all seven models. After uniform subsampling, the frame sequence no longer has a one-frame-per-second correspondence; the executed joint IP prompt did not account for this change.

\begin{center}
  \begin{minipage}{\textwidth}
  \centering
  \captionof{table}{Executable input and output budgets. The native-video control keeps audio, whereas the side-by-side SRM setting is explicitly rendered without audio.}
  \label{tab:protocol-budgets}
  \small
  \setlength{\tabcolsep}{4pt}
  \begin{tabular}{p{0.15\textwidth}p{0.17\textwidth}p{0.14\textwidth}p{0.45\textwidth}}
    \toprule
    Protocol & Visual budget & Output budget & Media/content-part order \\
    \midrule
    IP joint & 64 frames for the full reaction clip & 320 tokens & Prompt with moment ranges; reaction frames; ``JSON scores:'' \\
    IP independent & 12 frames for one marked window & 8 tokens & Prompt; window frames; ``Score (0-100):'' \\
    SRM joint & 12 reaction; 6 per option & 48 tokens & Prompt; reaction label/frames; A--D labels/frames; answer suffix \\
    SRM per-option setting & 12 reaction; 6 for one option & 8 tokens & Prompt; reaction label/frames; option label/frames; score suffix \\
    SRM side-by-side setting & one synchronized silent native video & 8 tokens & Prompt; reaction-left/stimulus-right video; score suffix \\
    GRG generation & 12 stimulus and 12 reaction & 80 tokens & Prompt; stimulus label/frames; reaction label/frames; explanation suffix \\
    GRG judging & Text only & 60 tokens & Rubric prompt containing accepted references and one model hypothesis \\
    \bottomrule
  \end{tabular}
  \end{minipage}
\end{center}

Temperature is zero where supported. GPT-5 uses minimal reasoning effort; Gemini uses its low thinking setting; Tinker-served and local Qwen models request direct, no-thinking rendering where supported. Local Hugging Face inference uses greedy decoding. Each reported inference setting has one run, with the recorded retry/resume behavior; deterministic decoding is requested where supported.

Table~\ref{tab:protocol-budgets} gives the requested answer budgets passed to the model wrapper. GPT-5's wrapper sets the completion cap, including reasoning, to the larger of 1,024 tokens and the requested budget plus 512. The Gemini wrapper adds 512 tokens for Flash or 2,048 for Pro; its fallback when thinking configuration is unsupported adds 4,096. Tinker and local inference use the requested generation cap directly. SRM reaction-input follow-ups request 64 tokens; this leaves GPT-5's effective 1,024-token cap unchanged from the original 48-token request.

\begin{center}
  \begin{minipage}{\textwidth}
  \centering
  \captionof{table}{Model-access taxonomy. ``Open weights'' describes weight availability, not where inference was executed; most open-weight benchmark runs were grant-billed through a hosted service.}
  \label{tab:model-access}
  \small
  \setlength{\tabcolsep}{4pt}
  \begin{tabular}{p{0.19\textwidth}p{0.16\textwidth}p{0.17\textwidth}p{0.39\textwidth}}
    \toprule
    Model & Weight access & Serving path & Protocol note \\
    \midrule
    GPT-5 & Proprietary API & OpenAI API & Minimal reasoning; low-detail image parts. \\
    Gemini 3.7 Flash & Proprietary API & Vertex AI & Low thinking; only model used for the native-video control. \\
    Qwen3.5-9B & Open weights & Tinker; local HF for adaptation & Direct/no-thinking template; identical local harness for base and LoRA. \\
    Qwen3.6-27B & Open weights & Tinker & Direct/no-thinking rendering where supported. \\
    Qwen3.6-35B-A3B & Open weights & Tinker & Direct/no-thinking rendering where supported. \\
    Qwen3.5-397B-A17B & Open weights & Tinker & Direct/no-thinking rendering where supported. \\
    Kimi-K2.6 & Open weights & Tinker & Direct/no-thinking rendering where supported. \\
    DeepSeek-V3.1 & Open weights & Tinker, text-only & Shared GRG judge from a family outside the generator set; neutrality is not assumed. \\
    \bottomrule
  \end{tabular}
  \end{minipage}
\end{center}

\subsection{IP Joint Prompt}

\begin{promptbox}[prompt:t1-joint]{IP joint scoring}
You will see frames of a REACTION VIDEO (a viewer watching content), sampled at 1 frame per second: frame N corresponds to second N. Below are \textless{}$k$\textgreater{} moments from this video, each a time range. Rate the INTENSITY of the viewer's reaction in each moment on a 0-100 scale (0 = calm baseline watching, 100 = peak reaction such as screaming, jumping, hands on face). Use the full scale to differentiate the moments. Output ONLY a JSON object mapping moment ids to integer scores, e.g. \{``m0'': 40, ``m1'': 85\}. Do NOT write any analysis or explanation --- the JSON object must be your entire answer.

Moments:

\hspace*{1em}m0: seconds \textless{}start\textgreater{} to \textless{}end\textgreater{}

\hspace*{1em}\ldots
\end{promptbox}

The prompt is followed by the reaction frames and the text ``JSON scores:''. Moment identifiers are assigned in list order and times are rendered to one decimal place. The native-video control replaces only the opening frame description with ``a REACTION VIDEO (a viewer watching content)'', inserts the uploaded video, and retains the moment list and output contract.

\subsection{IP Independent Prompt}

\begin{promptbox}[prompt:t1-independent]{IP independent scoring}
You will see frames (1 per second) of a viewer's reaction during ONE moment of a video they are watching. Rate the INTENSITY of the viewer's reaction on a 0-100 scale (0 = calm baseline watching, 100 = peak reaction such as screaming, jumping, hands on face). Reply with a single integer only.
\end{promptbox}

The prompt is followed by frames cropped to the marked moment window and the text ``Score (0-100):''. Each moment is a separate model call. Joint and independent evaluation use the same frozen 642 moments; their paired comparison is computed on the model-specific intersection of successfully parsed videos.

\subsection{SRM Joint Four-Way Prompt}

\begin{promptbox}[prompt:t2-joint]{SRM joint four-way choice}
A viewer is watching a video and reacting to it. First you will see frames (1 per second) of the VIEWER'S REACTION. Then four candidate STIMULUS segments (A, B, C, D) from the video they were watching, also as frames. Exactly one segment is what the viewer is reacting to; the others come from different times in the same video. Which segment triggered this reaction? Reply with ONLY a single letter (A, B, C, or D) and nothing else.
\end{promptbox}

The request then contains ``VIEWER'S REACTION:'', reaction frames, four repetitions of ``SEGMENT A:'' through ``SEGMENT D:'' followed by their frames, and finally ``Answer (single letter A/B/C/D):''. The parser accepts an initial verdict letter or the last standalone uppercase A--D (optionally preceded by ``segment''). Unparsed output is counted as incorrect.

\subsection{GRG Generation Prompt}

\begin{promptbox}[prompt:t3-generation]{GRG generation}
A viewer is watching a video. You will see two sets of frames sampled at 1 frame per second over the same time window: first the CONTENT the viewer was watching, then the VIEWER'S REACTION to it. In ONE sentence, state what specifically in the content caused this reaction --- name the concrete moment, action, event, or line, not a generic description of the content. Reply with that single sentence only.
\end{promptbox}

The request then contains ``CONTENT (stimulus):'', the stimulus frames, ``VIEWER'S REACTION:'', the reaction frames, and ``Your one-sentence explanation:''. Pre-screen explanations and human rationales are not included in the generator request.

\subsection{GRG Judge Prompt}

\begin{promptbox}[prompt:t3-judge]{GRG judge rubric}
You are grading a model-written explanation of why a viewer reacted during a specific video moment. \textless{}$n$\textgreater{} human annotator(s) watched that moment and independently wrote these reference explanations:

\textless{}human references\textgreater{}

Model's explanation: ``\textless{}model explanation\textgreater{}''

Score the model's explanation on three dimensions:

- cue (0-2): does it identify the same concrete trigger in the content as the references? 2 = same specific cue as at least one reference; 1 = related or partially overlapping cue; 0 = different cue or only a generic description.

- link (0-2): does it connect that cue to the viewer's reaction? 2 = clear causal link; 1 = weak or implicit; 0 = no link.

- faithful (0-1): 1 = free of invented specifics that contradict the references; 0 = contradicts them.

Reply ONLY with a JSON object, e.g. \{``cue'': 2, ``link'': 1, ``faithful'': 1\}.
\end{promptbox}

The field named faithful is interpreted as reference consistency: the judge checks contradictions with the reference text, without seeing the video. A correct but unmentioned cue can receive little cue credit, while an unsupported addition can escape the contradiction check. The JSON parser requires all three keys within their permitted ranges. Missing or invalid judgements remain unscored; aggregate tables disclose coverage. DeepSeek-V3.1 provides the shared comparison used for all seven generators.

\subsection{Follow-up Input Controls}
\label{app:followup-prompts}

Normal SRM and GRG controls use the generation prompts above. Shuffled-reaction controls keep those prompts and the stimulus parts, replacing only the reaction frames with the fixed donor's frames. Stimulus-only controls omit the reaction label and frames and use the following instructions. The option labels and final answer suffix in SRM, and stimulus label and final explanation suffix in GRG, retain their original positions. Neither control receives option descriptions, model nomination text, or gold references.

\begin{promptbox}[prompt:srm-stimulus-only]{SRM stimulus-only instruction}
A viewer was watching one of these four STIMULUS segments (A, B, C, D). The viewer's reaction is not provided. Choose the segment most likely to be the aligned stimulus. Reply with ONLY a single letter (A, B, C, or D).
\end{promptbox}

\begin{promptbox}[prompt:grg-stimulus-only]{GRG stimulus-only instruction}
A viewer was watching this CONTENT (stimulus). The viewer's reaction is not provided. In ONE sentence, identify a concrete moment, action, event, or line in the content that could explain a viewer's reaction. Reply with that single sentence only.
\end{promptbox}

Grouped-window IP uses the same frames as independent scoring: the original marked windows, each capped at 12 frames. Each group begins with \texttt{MOMENT m0:}, \texttt{MOMENT m1:}, and so on, followed by its frames. The final suffix is \texttt{JSON scores:}; the requested output budget is 320 tokens.

\begin{promptbox}[prompt:ip-grouped]{Grouped-window joint IP instruction}
You will see explicitly grouped moments from ONE viewer's reaction video. Rate the INTENSITY in each moment on a 0-100 scale (0 = calm baseline watching, 100 = peak reaction such as screaming, jumping, hands on face). Use the full scale to differentiate the moments. Output ONLY a JSON object mapping moment ids to integer scores, e.g. \{``m0'': 40, ``m1'': 85\}.
\end{promptbox}

Corrected full-video IP replaces the legacy frame--second statement with an ordered mapping from displayed frame positions to their extraction-grid timestamps. All marked moment ranges and the score scale are retained. The implementation and input hashes accompany each follow-up run; the corrected formatter is shared with the new training-data builder.

\subsection{Preliminary SRM Evaluation Settings}

These prompts were executed on the separate 150-item protocol-analysis option set. Their results diagnose presentation and tie handling only; they are not benchmark scores.

\paragraph{Per-Option Likelihood.}
\begin{promptbox}[prompt:t2-e2]{Preliminary SRM per-option setting}
A viewer is watching a video. You will see frames (1 per second) of the VIEWER'S REACTION, then frames of ONE candidate segment from the video they were watching. Rate from 1 to 10 how likely it is that THIS segment is what the viewer is reacting to at this moment. Reply with a single integer only.
\end{promptbox}
The prompt is followed by ``VIEWER'S REACTION:'', reaction frames, ``CANDIDATE SEGMENT:'', candidate frames, and ``Score (1-10):''. Each of four options is scored in a separate call.

\paragraph{Side-by-Side Native Video.}
\begin{promptbox}[prompt:t2-v3]{Preliminary SRM side-by-side setting}
You will see one video with two panels playing on the same timeline. LEFT: a viewer's reaction while watching a video. RIGHT: one candidate segment from the video they were watching. If the right side is the true trigger, the viewer's facial and body reactions on the left should line up in time with events on the right. Rate from 1 to 10 how likely it is that the RIGHT side is what the viewer is reacting to. Reply with a single integer only.
\end{promptbox}
The request contains one silent synchronized video with reaction on the left and a candidate stimulus on the right, followed by ``Score (1-10):''.

\subsection{Scoring, Uncertainty, and Failure Handling}

\begin{center}
  \begin{minipage}{\textwidth}
  \centering
  \captionof{table}{Scoring contracts. No invalid output is replaced by a favorable score; the only explicit imputation is the IP value of zero assigned to a parsed constant prediction.}
  \label{tab:scoring-contracts}
  \small
  \setlength{\tabcolsep}{4pt}
  \begin{tabular}{p{0.13\textwidth}p{0.25\textwidth}p{0.25\textwidth}p{0.27\textwidth}}
    \toprule
    Task & Reported unit and metric & Invalid/tied output policy & Additional reporting \\
    \midrule
    IP & Extended Spearman $\rho_0$~\citep{spearman1904proof} against full consensus, averaged over videos & Parsed constants receive $\rho_0=0$; missing/unparsed outputs are excluded and counted separately & Conditional Spearman, common-coverage summaries, and a separate matched human LOO pairwise analysis \\
    SRM & Item-level four-way accuracy & Unparsed letter is incorrect & Wilson interval; temporal distance of wrong selections \\
    GRG & Item-level cue (0--2) + link (0--2) + reference consistency (0--1) & Invalid judge JSON is unscored and coverage is disclosed & Dimension means, composite mean, seeded bootstrap interval, cross-judge sensitivity, human anchor \\
    \bottomrule
  \end{tabular}
  \end{minipage}
\end{center}

Unless stated otherwise, bootstrap intervals use 10,000 resamples and seed 0~\citep{efron1979bootstrap}. Follow-up source-group analyses use 20,000 resamples, with the resampling unit specified beside each result. Original IP scoring-protocol comparisons resample paired videos and report a two-sided Wilcoxon signed-rank test~\citep{wilcoxon1945individual} on the same model-specific intersection. SRM paired variants use McNemar's test~\citep{mcnemar1947test}. These intervals quantify variation across sampled benchmark items, videos, or source groups while holding model outputs fixed; they exclude decoding variability across repeated runs.

\paragraph{Inference Target and Multiple Comparisons.}
The human comparison targets the mean video-weighted difference in pairwise accuracy; its paired bootstrap intervals are primary for that estimand. Wilcoxon signed-rank tests retained from the original protocol and adaptation analyses test a different distributional hypothesis and can disagree with a mean-difference interval, especially with many zeros and ties. All reported 95\% intervals are pointwise, without familywise multiplicity adjustment. Nonsignificance is not an equivalence result. Video resampling holds the rater pool fixed and does not account for drawing new annotators. Five channel-level groups provide limited information about new-source generalization.

\section{Complete Results and Robustness Analyses}
\label{app:robustness}

This section reports all completed secondary analyses that affect interpretation of the central claims. Experiments on a separate 150-item SRM protocol-analysis pool are isolated as design analyses and are not benchmark scores.

\subsection{Notes to the Main Results Table}
\label{app:main-table-notes}

Table~\ref{tab:main-results} uses the standard IP task-set summaries; the independent Qwen3.5-397B-A17B run has 226 valid/tied video outputs rather than 250. SRM has $n=112$ per model; error distances are in seconds and are computed only over that model's incorrect items. Unparsed SRM responses are scored as incorrect: 20 for Qwen3.5-9B and 1 for Gemini 3.7 Flash, 0 elsewhere. GRG has $n=200$ per model except Gemini ($n=191$). Chance error distances are computed from the frozen item windows: a uniform random choice is, conditional on an error, uniform over the three distractors, so the chance values are the mean and median of all 336 distractor--correct start-time distances. ``N/A'' means the metric is not defined for that row.

Figure~\ref{fig:qualitative} uses IP video 66324 (moments 66324\_c00--c02, windows 29.5--33.0, 34.0--37.5, and 55.5--58.5\,s; four raters per moment) and GRG item 51931\_c00 (window 2.88--5.88\,s). All displayed scores, rationales, references, and judge outputs are taken verbatim from the frozen prediction and reference files; frames are downsampled copies of the evaluated inputs. The items were selected for legibility from the set on which the seven generators disagree; they illustrate protocol effects and are not representative estimates.

\subsection{IP Handling of Constant Predictions}

The extended IP score $\rho_0$ assigns zero to a parsed prediction that gives every moment in a video the same score. For robustness, we also report results after excluding those videos before averaging. Tables~\ref{tab:t1-joint-full} and~\ref{tab:t1-independent-full} show why the treatment and frequency of constant predictions must be reported together.

\begin{center}
  \begin{minipage}{\textwidth}
  \centering
  \captionof{table}{Full IP joint-setting results. Bootstrap intervals are shown after excluding constant predictions; $n$ discloses parse coverage and $n_{\rm nonconst}$ is the resulting denominator.}
  \label{tab:t1-joint-full}
  \small
  \begin{tabular*}{\textwidth}{@{\extracolsep{\fill}}lccclrr@{}}
    \toprule
    \textbf{Model} & Extended $\rho_0$ & $n$ & Excl. constant $\rho$ & 95\% CI & $n_{\rm nonconst}$ & Constant \\
    \midrule
    \rowcolor[gray]{0.96}\multicolumn{7}{l}{\textbf{\textit{Proprietary Models}}} \\
    GPT-5 & .026 & 250 & .026 & $[-.083,.135]$ & 250 & 0 \\
    Gemini 3.7 Flash & .176 & 249 & .176 & $[.067,.282]$ & 249 & 0 \\
    \midrule
    \rowcolor[gray]{0.96}\multicolumn{7}{l}{\textbf{\textit{Open-weight Models}}} \\
    Qwen3.5-9B & .029 & 250 & .029 & $[-.080,.137]$ & 250 & 0 \\
    Qwen3.6-27B & .163 & 250 & .163 & $[.057,.272]$ & 250 & 0 \\
    Qwen3.6-35B-A3B & .043 & 250 & .046 & $[-.063,.157]$ & 237 & 13 \\
    Qwen3.5-397B-A17B & .065 & 250 & .065 & $[-.044,.173]$ & 250 & 0 \\
    Kimi-K2.6 & .086 & 250 & .086 & $[-.025,.197]$ & 250 & 0 \\
    \bottomrule
  \end{tabular*}
  \end{minipage}
\end{center}

\begin{center}
  \begin{minipage}{\textwidth}
  \centering
  \captionof{table}{Full IP independent-setting results. Independent scoring improves the mean correlation but also produces many constant video-level predictions when separately generated integer scores are reassembled. Assigning these predictions a score of zero retains them in the video average.}
  \label{tab:t1-independent-full}
  \small
  \begin{tabular*}{\textwidth}{@{\extracolsep{\fill}}lccclrr@{}}
    \toprule
    \textbf{Model} & Extended $\rho_0$ & $n$ & Excl. constant $\rho$ & 95\% CI & $n_{\rm nonconst}$ & Constant \\
    \midrule
    \rowcolor[gray]{0.96}\multicolumn{7}{l}{\textbf{\textit{Proprietary Models}}} \\
    GPT-5 & .177 & 250 & .204 & $[.089,.318]$ & 217 & 33 \\
    Gemini 3.7 Flash & .316 & 250 & .328 & $[.224,.430]$ & 241 & 9 \\
    \midrule
    \rowcolor[gray]{0.96}\multicolumn{7}{l}{\textbf{\textit{Open-weight Models}}} \\
    Qwen3.5-9B & .255 & 250 & .334 & $[.216,.448]$ & 191 & 59 \\
    Qwen3.6-27B & .329 & 250 & .394 & $[.285,.500]$ & 209 & 41 \\
    Qwen3.6-35B-A3B & .210 & 250 & .257 & $[.142,.368]$ & 205 & 45 \\
    Qwen3.5-397B-A17B & .267 & 226 & .331 & $[.215,.444]$ & 182 & 44 \\
    Kimi-K2.6 & .233 & 250 & .286 & $[.172,.396]$ & 204 & 46 \\
    \bottomrule
  \end{tabular*}
  \end{minipage}
\end{center}

\subsection{Paired Comparison of the IP Settings}

\begin{center}
  \begin{minipage}{\textwidth}
  \centering
  \captionof{table}{Paired change in the extended score $\Delta\rho_0=\rho_{0,\rm independent}-\rho_{0,\rm joint}$ on each model's shared videos. Intervals are paired bootstrap intervals; $p$ is the two-sided Wilcoxon signed-rank test. The positive difference persists under both treatments of constant predictions, but the experiment does not isolate whether context, response factorization, or frame selection causes the change.}
  \label{tab:t1-paired-full}
  \small
  \begin{tabular*}{\textwidth}{@{\extracolsep{\fill}}lrrrrrrrr@{}}
    \toprule
    & \multicolumn{4}{c}{Extended: constants $\rho_0=0$} & \multicolumn{4}{c}{Alternative: exclude constants} \\
    \cmidrule(lr){2-5}\cmidrule(lr){6-9}
    \textbf{Model} & $n$ & $\Delta\rho_0$ & 95\% CI & $p$ & $n$ & $\Delta\rho$ & 95\% CI & $p$ \\
    \midrule
    \rowcolor[gray]{0.96}\multicolumn{9}{l}{\textbf{\textit{Proprietary Models}}} \\
    GPT-5 & 250 & .151 & $[.015,.286]$ & .0349 & 217 & .169 & $[.021,.312]$ & .0300 \\
    Gemini 3.7 Flash & 249 & .139 & $[.024,.254]$ & .0203 & 240 & .148 & $[.032,.268]$ & .0140 \\
    \midrule
    \rowcolor[gray]{0.96}\multicolumn{9}{l}{\textbf{\textit{Open-weight Models}}} \\
    Qwen3.5-9B & 250 & .226 & $[.088,.363]$ & .0017 & 191 & .317 & $[.149,.480]$ & .0002 \\
    Qwen3.6-27B & 250 & .166 & $[.033,.294]$ & .0132 & 209 & .226 & $[.081,.373]$ & .0015 \\
    Qwen3.6-35B-A3B & 250 & .167 & $[.028,.304]$ & .0187 & 196 & .207 & $[.041,.374]$ & .0126 \\
    Qwen3.5-397B-A17B & 226 & .182 & $[.046,.315]$ & .0135 & 182 & .249 & $[.099,.400]$ & .0024 \\
    Kimi-K2.6 & 250 & .147 & $[.019,.270]$ & .0299 & 204 & .188 & $[.046,.329]$ & .0076 \\
    \bottomrule
  \end{tabular*}
  \end{minipage}
\end{center}

\subsection{Matched Human Comparison and Coverage}
\label{app:ip-robustness}
\label{app:human-matched}

The model benchmark scores in Table~\ref{tab:main-results} use all-rater consensus. Human comparison is a separate post-hoc analysis of the frozen predictions, using identical reference targets and averaging weights on each paired unit. For each video $v$ and held-out rater $r$ who scored all selected moments, let $\mathbf y_{v,-r}$ be the mean of the remaining raters. Both that rater and the model are scored against $\mathbf y_{v,-r}$. A pair of moments is eligible only when its remaining-rater consensus differs. Prediction ties receive one half and correctly ordered pairs receive one; all other pairs receive zero. Scores are averaged over eligible pairs within each rater--video unit, then over raters within a video, then equally over videos. A constant remaining-rater consensus makes the unit ineligible; a constant prediction remains eligible. Missing or unparsed model output is excluded, never treated as a constant prediction.

Among 1,073 complete rater--video units, 117 have constant remaining-rater consensus. The resulting 956 units cover all 250 videos and include 536 constant human profiles. Table~\ref{tab:human-matched} reports the paired analysis, holding the eight-rater pool fixed while resampling videos. It measures consistency with that pool, not a human upper bound. Human scores have four behavioral categories, whereas model prompts request 0--100 scores and joint prompts explicitly encourage differentiation. Linear rescaling of the human categories leaves ranks and ties unchanged.

% Generated by scripts/export_review_tables.py from validation/revision-2026-09-17/reanalysis.json.
\begin{center}
\begin{minipage}{\textwidth}
\centering
\captionof{table}{Matched human--model LOO comparison. Pairwise accuracy is averaged within rater--video units, then within videos, then equally across videos. Reference ties are omitted; prediction ties receive half credit. $\Delta$ is model minus human. Pointwise 95\% intervals resample paired videos, holding the rater pool fixed. J/I denote joint/independent inputs. The model-specific coverage is part of each comparison.}
\label{tab:human-matched}
\small
\setlength{\tabcolsep}{3pt}
\begin{tabular*}{\textwidth}{@{\extracolsep{\fill}}lrrrrrrr@{}}
\toprule
Model & Mode & Units & Videos & Human & Model & $\Delta$ & 95\% CI \\
\midrule
GPT-5 & J & 956 & 250 & 0.607 & 0.516 & -0.091 & $[-0.148,-0.033]$ \\
GPT-5 & I & 956 & 250 & 0.607 & 0.587 & -0.019 & $[-0.073,0.034]$ \\
Gemini 3.7 Flash & J & 953 & 249 & 0.607 & 0.585 & -0.022 & $[-0.081,0.036]$ \\
Gemini 3.7 Flash & I & 956 & 250 & 0.607 & 0.658 & 0.051 & $[-0.002,0.105]$ \\
Qwen3.5-9B & J & 956 & 250 & 0.607 & 0.518 & -0.089 & $[-0.147,-0.032]$ \\
Qwen3.5-9B & I & 956 & 250 & 0.607 & 0.625 & 0.018 & $[-0.028,0.065]$ \\
Qwen3.6-27B & J & 956 & 250 & 0.607 & 0.583 & -0.023 & $[-0.080,0.034]$ \\
Qwen3.6-27B & I & 956 & 250 & 0.607 & 0.661 & 0.054 & $[0.005,0.101]$ \\
Qwen3.6-35B-A3B & J & 956 & 250 & 0.607 & 0.521 & -0.085 & $[-0.139,-0.030]$ \\
Qwen3.6-35B-A3B & I & 956 & 250 & 0.607 & 0.603 & -0.004 & $[-0.055,0.045]$ \\
Qwen3.5-397B-A17B & J & 956 & 250 & 0.607 & 0.533 & -0.074 & $[-0.130,-0.017]$ \\
Qwen3.5-397B-A17B & I & 861 & 226 & 0.609 & 0.634 & 0.024 & $[-0.026,0.075]$ \\
Kimi-K2.6 & J & 956 & 250 & 0.607 & 0.548 & -0.059 & $[-0.117,-0.001]$ \\
Kimi-K2.6 & I & 956 & 250 & 0.607 & 0.616 & 0.009 & $[-0.043,0.060]$ \\
\bottomrule
\end{tabular*}
\end{minipage}
\end{center}

Gemini joint scores 0.585 against its matched human reference of 0.607 on 249 videos; the model-minus-human interval spans $[-0.081,0.036]$. The interval allows differences in either direction. Qwen3.6-27B independent has a positive pointwise difference of 0.054 ($[0.005,0.101]$), but the reference is a single-rater agreement statistic, not the best achievable human performance. This pointwise interval holds the rater pool fixed; it excludes between-pool uncertainty and multiplicity adjustment.

\paragraph{Conditional Ranking.}
Standard Spearman is reported only where both predictions and the remaining-rater consensus are nonconstant, with unit and video counts (Table~\ref{tab:human-conditional}). For Gemini joint this gives 420 units over 166 videos: human $\rho=0.557$, model $\rho=0.189$, and difference $-0.368$ ($[-0.533,-0.201]$). On exactly those units, human pairwise accuracy is 0.772 versus 0.597 for the model. The change from the overall human pairwise reference near 0.607 to 0.772 is due to eligibility, with the metric and weighting unchanged. Each model's conditional subset differs, so these results cannot supply a unified model ranking.

% Generated by scripts/export_review_tables.py from validation/revision-2026-09-17/reanalysis.json.
\begin{center}
\begin{minipage}{\textwidth}
\centering
\captionof{table}{Conditional standard Spearman where both predictions and the remaining-rater consensus are nonconstant. Coverage gives rater--video units / unique videos. Means use the same nested video weighting as Table~\ref{tab:human-matched}; intervals are for model minus human. These model-specific subsets cannot define a common leaderboard.}
\label{tab:human-conditional}
\small
\setlength{\tabcolsep}{3pt}
\begin{tabular*}{\textwidth}{@{\extracolsep{\fill}}lrrrrr@{}}
\toprule
Model & Mode & Units/videos & Human $\rho$ & Model $\rho$ & 95\% CI for $\Delta$ \\
\midrule
GPT-5 & J & 420/166 & 0.557 & 0.054 & $[-0.658,-0.343]$ \\
GPT-5 & I & 379/150 & 0.546 & 0.202 & $[-0.512,-0.179]$ \\
Gemini 3.7 Flash & J & 420/166 & 0.557 & 0.189 & $[-0.533,-0.201]$ \\
Gemini 3.7 Flash & I & 412/162 & 0.546 & 0.369 & $[-0.329,-0.024]$ \\
Qwen3.5-9B & J & 420/166 & 0.557 & 0.066 & $[-0.647,-0.333]$ \\
Qwen3.5-9B & I & 326/127 & 0.531 & 0.394 & $[-0.297,0.017]$ \\
Qwen3.6-27B & J & 420/166 & 0.557 & 0.183 & $[-0.525,-0.218]$ \\
Qwen3.6-27B & I & 353/140 & 0.504 & 0.420 & $[-0.242,0.072]$ \\
Qwen3.6-35B-A3B & J & 407/161 & 0.556 & 0.084 & $[-0.629,-0.316]$ \\
Qwen3.6-35B-A3B & I & 351/139 & 0.538 & 0.273 & $[-0.432,-0.094]$ \\
Qwen3.5-397B-A17B & J & 420/166 & 0.557 & 0.084 & $[-0.624,-0.319]$ \\
Qwen3.5-397B-A17B & I & 301/122 & 0.547 & 0.324 & $[-0.397,-0.054]$ \\
Kimi-K2.6 & J & 420/166 & 0.557 & 0.129 & $[-0.590,-0.271]$ \\
Kimi-K2.6 & I & 354/139 & 0.578 & 0.266 & $[-0.475,-0.146]$ \\
\bottomrule
\end{tabular*}
\end{minipage}
\end{center}

\paragraph{Extended Correlation Sensitivity.}
On the overall matched Gemini joint units, $\rho_0$ is 0.222 for humans and 0.166 for the model, with difference $-0.056$ ($[-0.172,0.060]$). The 0.649 human value is the rater--video-weighted standard Spearman over 420 nonconstant units. It is meaningful only as that conditional statistic. Video weighting on those units yields 0.557 over 166 videos. Neither number is directly comparable with constant-inclusive model scores against the full consensus. Bootstrapping the selected records or changing the metric without changing eligibility does not repair that asymmetry.

\paragraph{Source Dependence.}
Seven creators reduce to five channels after merging same-channel personas. Table~\ref{tab:channel-sensitivity} reports channel-specific matched accuracies and leave-one-channel-out differences. These descriptive checks characterize the collected sources; five channels provide limited evidence about broader viewer populations. The video-bootstrap intervals above hold sources and annotators fixed except for the empirical distribution induced by resampling videos.

% Generated by scripts/export_review_tables.py from validation/revision-2026-09-17/reanalysis.json.
\begin{center}
\begin{minipage}{\textwidth}
\centering
\captionof{table}{Matched pairwise accuracy by channel and leave-one-channel-out model-minus-human differences for Gemini joint and Qwen3.6-27B independent. Channel labels are anonymized. Gemini has one missing joint output; Qwen has full coverage. Five channels support descriptive sensitivity checks, not precise population-level inference.}
\label{tab:channel-sensitivity}
\small
\setlength{\tabcolsep}{3pt}
\begin{tabular*}{\textwidth}{@{\extracolsep{\fill}}lrrrrrr@{}}
\toprule
Channel & $n_J$ & Human$_J$ & Gemini J & LOCO $\Delta_J$ & Qwen I & LOCO $\Delta_I$ \\
\midrule
C1 & 100 & 0.579 & 0.554 & -0.020 & 0.628 & 0.057 \\
C2 & 16 & 0.732 & 0.605 & -0.015 & 0.794 & 0.053 \\
C3 & 99 & 0.567 & 0.612 & -0.066 & 0.663 & 0.026 \\
C4 & 9 & 0.769 & 0.567 & -0.015 & 0.637 & 0.061 \\
C5 & 25 & 0.739 & 0.597 & -0.009 & 0.704 & 0.064 \\
\bottomrule
\end{tabular*}
\end{minipage}
\end{center}

\subsection{Frame-Time Indexing Audit}
\label{app:timestamp-audit}

The original joint IP prompt says that frame $N$ corresponds to second $N$. This instruction is incorrect after uniform subsampling, because the request includes no per-frame timestamps. Five of 250 evaluation videos have more than 64 extracted frames (65, 68, 74, 77, and 91). Table~\ref{tab:timestamp-sensitivity} excludes these videos from the paired comparison. All seven mean differences remain positive; six bootstrap intervals exclude zero, while GPT-5's interval includes zero. The effect direction persists across all models, with weaker inferential support for GPT-5.

% Generated by scripts/export_review_tables.py from validation/revision-2026-09-17/reanalysis.json.
\begin{center}
\begin{minipage}{\textwidth}
\centering
\captionof{table}{Independent minus joint $\rho_0$ after excluding all five IP videos whose sampled frame indices no longer match the prompt\textquotesingle s seconds. All mean differences remain positive; the GPT-5 bootstrap interval includes zero. The secondary Wilcoxon test also does not reject zero for Kimi. This is an exclusion analysis of existing outputs, not a corrected-input rerun.}
\label{tab:timestamp-sensitivity}
\small
\setlength{\tabcolsep}{3pt}
\begin{tabular*}{\textwidth}{@{\extracolsep{\fill}}lrrrr@{}}
\toprule
Model & Videos & $\Delta\rho_0$ & 95\% mean CI & Wilcoxon $p$ \\
\midrule
GPT-5 & 245 & 0.131 & $[-0.005,0.265]$ & 0.068 \\
Gemini 3.7 Flash & 244 & 0.125 & $[0.008,0.244]$ & 0.036 \\
Qwen3.5-9B & 245 & 0.227 & $[0.086,0.367]$ & 0.002 \\
Qwen3.6-27B & 245 & 0.157 & $[0.028,0.290]$ & 0.020 \\
Qwen3.6-35B-A3B & 245 & 0.165 & $[0.027,0.305]$ & 0.021 \\
Qwen3.5-397B-A17B & 221 & 0.177 & $[0.042,0.317]$ & 0.018 \\
Kimi-K2.6 & 245 & 0.126 & $[0.003,0.253]$ & 0.062 \\
\bottomrule
\end{tabular*}
\end{minipage}
\end{center}

The defect also occurs in the original joint training: 838 of 856 oversampled rows, representing 419 of 428 unique training clips, contain subsampled nonconsecutive frame indices and the same incorrect instruction. It affects 47 of 48 joint validation rows. The revised implementation provides an ordered frame-to-time mapping using the extraction grid: file index $j$, beginning at one, maps to $(j-1)/f$ seconds at extraction rate $f$. Subsampling retains the original indices. Training and evaluation now share this formatter; these are grid timestamps, not sub-frame event annotations. Existing adapters were not retrained with this correction.

We reran Qwen3.5-9B on the five affected evaluation videos with the corrected prompt. Four per-video $\rho_0$ scores remain unchanged; one changes from $+1$ to $-1$. Replacing only these five historical predictions changes the 250-video joint mean from 0.029 to 0.021 and the independent-minus-joint mean difference from 0.226 to 0.234. This is a partial replacement sensitivity analysis, not a complete rerun or a causal estimate of the prompt correction. The original benchmark table remains frozen. A Gemini attempt failed with a provider permission error and supplies no replacement scores.

The subsequent five-case follow-up also completed for GPT-5, Qwen3.6-27B, Qwen3.6-35B-A3B, Qwen3.5-397B-A17B, and Kimi-K2.6. Table~\ref{tab:timestamp-followup} reports each partial-replacement mean. Changes range from $-0.008$ to $+0.008$ across the six accessible models, including the earlier Qwen9B run. Gemini again returned no usable predictions.

% Generated by scripts/analyze_followup.py.
\begin{center}
\begin{minipage}{\textwidth}
\centering
\captionof{table}{Corrected-timestamp sensitivity: replace the five subsampled joint-IP cases while retaining every other historical prediction. These mixed-batch means are sensitivity summaries, not full benchmark reruns or causal estimates of the prompt correction. The Qwen9B row uses the earlier five-case follow-up; the separate three-condition experiment reruns all 250 videos. Gemini returned no usable corrected predictions.}
\label{tab:timestamp-followup}
\small
\setlength{\tabcolsep}{4pt}
\begin{tabular*}{\textwidth}{@{\extracolsep{\fill}}lrrr@{}}
\toprule
Model & Corrected & Historical $\rho_0$ & Five replaced $\rho_0$ \\
\midrule
GPT-5 & 5/5 & 0.026 & 0.026 \\
Gemini 3.7 Flash & 0/5 & --- & --- \\
Qwen3.5-9B & 5/5 & 0.029 & 0.021 \\
Qwen3.6-27B & 5/5 & 0.163 & 0.167 \\
Qwen3.6-35B-A3B & 5/5 & 0.043 & 0.043 \\
Qwen3.5-397B-A17B & 5/5 & 0.065 & 0.073 \\
Kimi-K2.6 & 5/5 & 0.086 & 0.094 \\
\bottomrule
\end{tabular*}
\end{minipage}
\end{center}

\subsection{Grouped-Window Joint Scoring}
\label{app:grouped-joint}

We reran all 250 IP videos with Qwen3.5-9B under three conditions: full-video joint with corrected timestamps, grouped-window joint, and independent scoring. Grouped-window joint uses exactly the independent condition's frames for each marked window, with explicit moment labels, and requests all scores in one response. Full-video and grouped-window joint request 320 output tokens; independent requests 8 per moment. All conditions use the same Tinker backbone and decoding configuration. The study contains 1,142 successful calls: 250 per joint condition and 642 independent moment calls. Four grouped-window responses return per-moment arrays instead of scalar scores and fail parsing; all full-video and independent profiles parse. Paired analyses use the 246 common videos from 111 source groups.

% Generated by scripts/analyze_followup.py.
\begin{center}
\begin{minipage}{\textwidth}
\centering
\captionof{table}{Fresh Qwen3.5-9B IP runs. All three conditions use the same model and serving path in this follow-up. Grouped-window joint and independent use identical frames per marked window; grouped scoring returns all scores in one call. Full-video joint uses corrected frame times. Means and constant counts use the 246 common videos; $\Delta$ is relative to full-video joint, with pointwise 95\% source-group bootstrap intervals (20,000 resamples).}
\label{tab:grouped-joint}
\small
\setlength{\tabcolsep}{4pt}
\begin{tabular*}{\textwidth}{@{\extracolsep{\fill}}lrrrrl@{}}
\toprule
Input/output & Valid & Constant & $\rho_0$ & $\Delta$ & 95\% CI \\
\midrule
Full-video joint & 250 & 1 & 0.039 & --- & --- \\
Grouped-window joint & 246 & 0 & 0.172 & +0.133 & $[0.026,0.240]$ \\
Independent & 250 & 56 & 0.278 & +0.239 & $[0.106,0.373]$ \\
\bottomrule
\end{tabular*}
\end{minipage}
\end{center}

Mean $\rho_0$ is 0.039 for full-video joint, 0.172 for grouped-window joint, and 0.278 for independent scoring. Grouping the windows improves on full-video joint by 0.133 ($[0.026,0.240]$); independent improves on grouped joint by a further 0.105 ($[0.015,0.200]$). These are pointwise source-group bootstrap intervals. The intermediate condition reduces the observed protocol difference, with performance increasing at each step of the protocol change. Grouping also changes context, frame count, and localization demands, while separate calls change instructions and output organization. The result therefore distinguishes protocol components without isolating a single causal mechanism. Exact follow-up instructions appear in Appendix~\ref{app:followup-prompts}.

\subsection{Missing Outputs and Common Coverage}

Qwen3.5-397B-A17B independent scoring produces 226 complete video profiles. Each of the other 24 videos has moment-level usage entries in the call ledger, but no complete score vector in the prediction file. The ledger establishes that inference was attempted; it lacks the raw responses and failure traces needed to distinguish parsing, serving, and execution interruptions. Gemini joint has one unparsed video. For GRG, Gemini has 200 generation records, of which nine contain empty rationale strings. The judge runner skips empty strings, leaving 191 scored items with valid judge JSON. We report these as empty generations followed by absent judge scores and leave their scores unassigned.

Table~\ref{tab:common-coverage} gives means on each protocol's seven-model intersection. Table~\ref{tab:common-paired} adds paired differences from a fixed GPT-5 reference, resampling recorded source groups to retain within-recording dependence. All 21 model pairs per protocol are retained in the accompanying analysis. These pointwise intervals condition on the observed rater pool and are unadjusted for multiple comparisons.

% Generated by scripts/export_review_tables.py from validation/revision-2026-09-17/reanalysis.json.
\begin{center}
\begin{minipage}{\textwidth}
\centering
\captionof{table}{Common-item summaries across the seven zero-shot models: 249 jointly scored IP videos, 226 independently scored IP videos, and 191 GRG items with a shared-judge score for every model. These columns avoid model-specific denominators but do not test pairwise significance.}
\label{tab:common-coverage}
\small
\setlength{\tabcolsep}{3pt}
\begin{tabular*}{\textwidth}{@{\extracolsep{\fill}}lrrr@{}}
\toprule
Model & Joint $\rho_0$ & Independent $\rho_0$ & GRG / 5 \\
\midrule
GPT-5 & 0.023 & 0.157 & 2.791 \\
Gemini 3.7 Flash & 0.176 & 0.306 & 2.654 \\
Qwen3.5-9B & 0.029 & 0.288 & 2.843 \\
Qwen3.6-27B & 0.161 & 0.317 & 2.644 \\
Qwen3.6-35B-A3B & 0.044 & 0.205 & 2.780 \\
Qwen3.5-397B-A17B & 0.062 & 0.267 & 2.921 \\
Kimi-K2.6 & 0.086 & 0.214 & 2.948 \\
\bottomrule
\end{tabular*}
\end{minipage}
\end{center}

% Generated by scripts/complete_review_checks.py.
\begin{center}
\begin{minipage}{\textwidth}
\centering
\captionof{table}{Paired model-minus-GPT-5 differences on the common items for each protocol, with pointwise 95\% source-recording-group bootstrap intervals (20,000 resamples). GPT-5 is a fixed reporting reference; all 21 model pairs per protocol are retained in the analysis artifact. IP uses $\rho_0$ and GRG uses the five-point composite. Sample sizes are 249 joint IP videos, 226 independent IP videos, and 191 GRG items. Intervals hold the annotator pool fixed and are unadjusted for multiple comparisons.}
\label{tab:common-paired}
\small
\setlength{\tabcolsep}{3pt}
\begin{tabular*}{\textwidth}{@{\extracolsep{\fill}}lrrr@{}}
\toprule
Model & Joint $\Delta\rho_0$ & Independent $\Delta\rho_0$ & GRG $\Delta$/5 \\
\midrule
GPT-5 & --- & --- & --- \\
Gemini 3.7 Flash & $+0.152\;[0.014,0.274]$ & $+0.148\;[0.054,0.237]$ & $-0.136\;[-0.446,0.172]$ \\
Qwen3.5-9B & $+0.005\;[-0.081,0.097]$ & $+0.131\;[0.012,0.255]$ & $+0.052\;[-0.212,0.316]$ \\
Qwen3.6-27B & $+0.138\;[0.054,0.225]$ & $+0.160\;[0.057,0.259]$ & $-0.147\;[-0.420,0.126]$ \\
Qwen3.6-35B-A3B & $+0.021\;[-0.067,0.108]$ & $+0.047\;[-0.069,0.162]$ & $-0.010\;[-0.269,0.254]$ \\
Qwen3.5-397B-A17B & $+0.039\;[-0.058,0.130]$ & $+0.109\;[0.019,0.197]$ & $+0.131\;[-0.118,0.382]$ \\
Kimi-K2.6 & $+0.063\;[-0.029,0.149]$ & $+0.057\;[-0.048,0.152]$ & $+0.157\;[-0.080,0.393]$ \\
\bottomrule
\end{tabular*}
\end{minipage}
\end{center}

\subsection{Reaction-Input Controls for SRM}
\label{app:reaction-controls}

We ran GPT-5 under three conditions in a separate follow-up: the normal reaction-plus-options input, stimulus options alone, and the same options with another reaction. The stimulus-only instruction acknowledges that the reaction is unavailable; it does not include descriptions, nomination text, or the answer. Reaction donors use different clip IDs from the same creator, preferring the same recorded source and then the nearest reaction-frame count, with deterministic tie-breaking. This gives donors for 109 of 112 items; three without a same-creator donor are excluded from the three-way comparison. The donor assignment is fixed before scoring and does not use target labels or model outputs.

% Generated by scripts/analyze_new_controls.py.
\begin{center}
\begin{minipage}{\textwidth}
\centering
\captionof{table}{GPT-5 reaction-input controls on the 109 SRM items with same-creator donors, from 94 clips and 47 recorded source groups. Normal and stimulus-only runs cover all 112 items; this table pairs all conditions on identical items. $\Delta$ is normal minus control accuracy, with paired source-group bootstrap intervals (20,000 resamples). Donors preferentially share the source recording, then minimize reaction-frame-count difference; 67 items use same-source donors. All responses parse.}
\label{tab:srm-reaction-controls}
\small
\begin{tabular*}{\textwidth}{@{\extracolsep{\fill}}lrrrr@{}}
\toprule
Input & Correct / 109 & Accuracy & $\Delta$ & 95\% CI \\
\midrule
Normal & 46 / 109 & 0.422 & --- & --- \\
Stimulus only & 41 / 109 & 0.376 & 0.046 & $[-0.036,0.136]$ \\
Shuffled reaction & 42 / 109 & 0.385 & 0.037 & $[-0.048,0.124]$ \\
\bottomrule
\end{tabular*}
\end{minipage}
\end{center}

On all 112 items, the new normal run scores 48/112 (0.429) and stimulus-only scores 44/112 (0.393). These new calls are kept separate from the historical normal score of 57/112 (0.509); a requested model alias does not establish a fixed provider snapshot. The shared-109 comparison in Table~\ref{tab:srm-reaction-controls} uses contemporaneous runs throughout. Normal-minus-control differences are 0.046 and 0.037; their paired source-group bootstrap intervals are $[-0.036,0.136]$ and $[-0.048,0.124]$. Secondary item-level McNemar tests give $p=0.442$ and $0.503$. On the 67 items whose donor also shares the source recording, normal and shuffled accuracy are 0.373 and 0.418. Thus this model's accuracy does not yet depend on the specific reaction shown; the results neither demonstrate equivalence nor show that items are solvable without reaction evidence. Extending the control to other generators is straightforward with the released donor assignment.

\subsection{Reaction-Input Controls for GRG}
\label{app:grg-controls}

The GRG follow-up uses GPT-5 with the original generation prompt for normal and shuffled-reaction inputs. The stimulus-only condition removes reaction frames and asks for a concrete stimulus event that could explain a viewer's reaction. Stimulus frames, frame caps, and output budgets otherwise match. As in SRM, donors are fixed without using labels or predictions, require the same creator and a different clip ID, and preferentially share the recorded source. Normal and stimulus-only generation cover all 200 items; shuffled generation covers the 197 items with eligible donors. All 597 outputs are nonempty and receive valid scores from the same DeepSeek-V3.1 text judge, using the original rubric and reference set. Returned OpenAI metadata identifies every generation as \texttt{gpt-5-2025-08-07}.

% Generated by scripts/analyze_followup.py.
\begin{center}
\begin{minipage}{\textwidth}
\centering
\captionof{table}{GPT-5 GRG reaction-input controls with the shared DeepSeek-V3.1 text judge. Dimension and composite means use 197 common items from 182 clips and 105 source groups; scored counts disclose full-condition coverage. $\Delta$ is normal minus control composite, with pointwise 95\% source-group bootstrap intervals (20,000 resamples). Ref.\ denotes reference consistency.}
\label{tab:grg-controls}
\small
\setlength{\tabcolsep}{4pt}
\begin{tabular*}{\textwidth}{@{\extracolsep{\fill}}lrrrrrrl@{}}
\toprule
Input & Scored & Cue & Link & Ref. & Total/5 & $\Delta$ & 95\% CI \\
\midrule
Normal & 200 & 1.208 & 1.041 & 0.442 & 2.690 & --- & --- \\
Stimulus only & 200 & 1.112 & 0.538 & 0.310 & 1.959 & +0.731 & $[0.489,0.972]$ \\
Shuffled reaction & 197 & 1.274 & 1.071 & 0.401 & 2.746 & -0.056 & $[-0.296,0.177]$ \\
\bottomrule
\end{tabular*}
\end{minipage}
\end{center}

On the 197 common items, normal-input explanations score 2.690/5, compared with 1.959 for stimulus-only and 2.746 for shuffled reactions. Normal minus stimulus-only is 0.731, with a source-group bootstrap interval of $[0.489,0.972]$; normal minus shuffled is $-0.056$ ($[-0.296,0.177]$). Most of the stimulus-only decrease is in the linkage dimension (1.041 to 0.538). This contrast changes both reaction visibility and the corresponding instruction. The matched-prompt shuffle comparison does not establish an advantage from the correctly paired reaction. Reference-based scoring can therefore reward reaction-oriented explanations even when the supplied reaction comes from another clip. On the 120 items with same-source donors, normal and shuffled scores are 2.692 and 2.758. These controls assess this generator and rubric.

\subsection{Frozen SRM Results}

Table~\ref{tab:t2-rebuilt-full} reports the complete frozen-set run. Unparsed responses count as incorrect, giving every model the same denominator of 112 items. Error distance is the absolute difference between the start of the correct window and the start of the selected window, averaged over the incorrect items for which a choice was parsed; it describes how far a wrong answer lies from the correct window and is not comparable across models with very different accuracies.

\begin{center}
  \begin{minipage}{\textwidth}
  \centering
  \captionof{table}{Frozen SRM four-way matching, 112 items, chance .25. Intervals are Wilson intervals. The Gemini native-video row is a full-modality control and is not part of the frame-based comparison.}
  \label{tab:t2-rebuilt-full}
  \small
  \setlength{\tabcolsep}{3.4pt}
  \begin{tabular}{lrlrrr}
    \toprule
    \textbf{Model} & Acc. & 95\% CI & Unparsed & Mean err.\ (s) & Med.\ err.\ (s) \\
    \midrule
    \rowcolor[gray]{0.96}\multicolumn{6}{l}{\textbf{\textit{Proprietary Models}}} \\
    GPT-5 & .509 & $[.418,.600]$ & 0 & 29.4 & 26.8 \\
    Gemini 3.7 Flash & .500 & $[.409,.591]$ & 1 & 26.0 & 24.8 \\
    \midrule
    \rowcolor[gray]{0.96}\multicolumn{6}{l}{\textbf{\textit{Open-weight Models}}} \\
    Qwen3.6-35B-A3B & .482 & $[.392,.574]$ & 0 & 28.9 & 26.8 \\
    Qwen3.6-27B & .446 & $[.358,.539]$ & 0 & 26.6 & 25.7 \\
    Qwen3.5-397B-A17B & .411 & $[.324,.503]$ & 0 & 25.4 & 24.7 \\
    Kimi-K2.6 & .375 & $[.291,.467]$ & 0 & 26.3 & 24.5 \\
    Qwen3.5-9B & .259 & $[.187,.347]$ & 20 & 27.2 & 26.4 \\
    \midrule
    \rowcolor[gray]{0.96}\multicolumn{6}{l}{\textbf{\textit{Control and Reference}}} \\
    Gemini 3.7 Flash, native video & .545 & $[.453,.634]$ & 0 & 27.7 & 27.5 \\
    Chance & .250 & N/A & N/A & N/A & N/A \\
    \bottomrule
  \end{tabular}
  \end{minipage}
\end{center}

Per-item cultural-specificity strata are recorded in the task file but are not analyzed: 93 of the 112 items fall in one stratum and the remaining two hold 12 and 7 items, leaving little support for comparisons across strata. The spacing and description checks reduce near-duplicate alternatives. Difficulty can arise from sparse inputs, stimulus content, or the reaction--stimulus relation; the GPT-5 reaction-input controls in Appendix~\ref{app:reaction-controls} address the last distinction.

\subsection{Cross-Video Distractors and Audio-Based Rejection Cues}
\label{app:xv}

A cross-video variant keeps each item's correct stimulus window and replaces the three same-video distractors with trigger windows from other videos. This removes the pairwise-separation constraint, so the variant recovers the 68 candidates that could not form a four-option within-video item, for 180 items in total; the 112 items that share their correct option with the frozen within-video set support a paired per-model comparison. Table~\ref{tab:xv-audio} reports both.

\begin{center}
  \begin{minipage}{\textwidth}
  \centering
  \captionof{table}{Cross-video distractor control. ``Within'' and ``cross'' are accuracies on the 112 shared items; $p$ is exact McNemar on the paired items; the last column is accuracy on all 180 cross-video items. The native-video score of .777 uses different distractor sources and is not a comparable estimate of within-video matching performance.}
  \label{tab:xv-audio}
  \small
  \setlength{\tabcolsep}{4.5pt}
  \begin{tabular}{lrrrrr}
    \toprule
    \textbf{Model} & Within & Cross & $\Delta$ & $p$ & Cross (180) \\
    \midrule
    \rowcolor[gray]{0.96}\multicolumn{6}{l}{\textbf{\textit{Proprietary Models (frames)}}} \\
    GPT-5 & .509 & .429 & $-.080$ & .18 & .400 \\
    Gemini 3.7 Flash & .500 & .536 & $+.036$ & .63 & .461 \\
    \midrule
    \rowcolor[gray]{0.96}\multicolumn{6}{l}{\textbf{\textit{Open-weight Models (frames)}}} \\
    Qwen3.6-35B-A3B & .482 & .366 & $-.116$ & .06 & .278 \\
    Qwen3.6-27B & .446 & .473 & $+.027$ & .72 & .372 \\
    Qwen3.5-397B-A17B & .411 & .375 & $-.036$ & .64 & .322 \\
    Kimi-K2.6 & .375 & .411 & $+.036$ & .64 & .400 \\
    Qwen3.5-9B & .259 & .277 & $+.018$ & .86 & .250 \\
    \midrule
    \rowcolor[gray]{0.96}\multicolumn{6}{l}{\textbf{\textit{Native-video Control}}} \\
    Gemini 3.7 Flash, native video (audio) & .545 & \textbf{.777} & $+.232$ & $6\times10^{-6}$ & .761 \\
    \bottomrule
  \end{tabular}
  \end{minipage}
\end{center}

For the seven frame-based models, the changes are mixed in sign, range from $-.116$ to $+.036$, and do not reach the reported McNemar threshold. The all-180 accuracies are also lower than the shared-112 accuracies because the recovered items skew harder. On the same 112 items, the native-video model gains $+.232$ (McNemar 4/30 discordant) with cross-video distractors. This pattern is consistent with audio from other videos providing an additional rejection cue, potentially through stimulus-audio bleed in the reaction clips. Same-video distractors reduce source-level soundtrack mismatch, but they do not eliminate possible matching between bleed-through audio and the temporally aligned option. The control therefore characterizes sensitivity to distractor source rather than isolating the total effect of audio. The native-video cross-video score of .777 should not be reported as within-video matching capability.

\subsection{SRM Presentation Comparison on the Preliminary Set}

In the per-option protocol, each candidate receives an independent 1--10 likelihood score. Four quantities are necessary to interpret its apparent accuracy: strict top-1 counts a tied maximum as incorrect; tie rate reports how often the maximum is not unique; random-tie expected accuracy gives each tied maximum equal probability; and decided-only accuracy discards all ties. The last measure uses a self-selected, model-dependent subset and is not comparable without its denominator.

\begin{center}
  \begin{minipage}{\textwidth}
  \centering
  \captionof{table}{SRM setting comparison on the separate 150-item protocol-analysis option set. These values show that per-option presentation induces 52--81\% tied maxima and makes apparent performance convention-dependent. They are not \rb{} benchmark scores.}
  \label{tab:t2-presentation-diagnostic}
  \small
  \begin{tabular*}{\textwidth}{@{\extracolsep{\fill}}lrrrrr@{}}
    \toprule
    \textbf{Model} & Strict acc. & Tie rate & Random-tie exp. & Decided-only acc. & $n_{\rm decided}$ \\
    \midrule
    \rowcolor[gray]{0.96}\multicolumn{6}{l}{\textbf{\textit{Separate option frames}}} \\
    GPT-5 & .200 & .520 & .378 & .417 & 72 \\
    Qwen3.6-35B-A3B & .207 & .667 & .392 & .620 & 50 \\
    Kimi-K2.6 & .187 & .667 & .387 & .560 & 50 \\
    Gemini 3.7 Flash & .147 & .627 & .349 & .393 & 56 \\
    Qwen3.5-397B-A17B & .140 & .647 & .303 & .396 & 53 \\
    Qwen3.5-9B & .133 & .733 & .326 & .500 & 40 \\
    Qwen3.6-27B & .093 & .813 & .307 & .500 & 28 \\
    \midrule
    \rowcolor[gray]{0.96}\multicolumn{6}{l}{\textbf{\textit{Side-by-side native video}}} \\
    Gemini 3.7 Flash & .147 & .613 & .322 & .379 & 58 \\
    \bottomrule
  \end{tabular*}
  \end{minipage}
\end{center}

This comparison motivates the joint A/B/C/D setting, in which all candidates are visible in a single call and the model must return one choice. It does not establish that any model is below chance, nor does it validate that option set.

\subsection{Native-Video Control}

\begin{center}
  \begin{minipage}{\textwidth}
  \centering
  \captionof{table}{Gemini 3.7 Flash sampled-frame versus native-video controls. Native video retains audio, so the comparison is not a pure test of temporal encoding. The IP paired values use the shared 249 videos; the SRM row is an exact McNemar test on the frozen 112-item set. Both differences point the same way and neither is significant, so this remains a single-model, full-modality control.}
  \label{tab:native-video-control}
  \small
  \begin{tabular*}{\textwidth}{@{\extracolsep{\fill}}lrrrr>{\raggedright\arraybackslash}p{0.34\textwidth}@{}}
    \toprule
    Task & Frames & Native video & Paired $n$ & $p$ & Paired effect \\
    \midrule
    IP $\rho_0$ & .176 & .260 & 249 & .271 & $\Delta=.083$, 95\% CI $[-.049,.213]$ \\
    SRM accuracy & .500 & .545 & 112 & .542 & $\Delta=.045$; 24 video-only versus 19 frames-only correct \\
    \bottomrule
  \end{tabular*}
  \end{minipage}
\end{center}

Native-video input used roughly 30 times fewer counted input tokens in this control. This is a measured implementation-level efficiency observation, not a claim about total latency, provider compute, or model-family superiority.

\subsection{SRM Reporting Scope}

Per-model accuracy, intervals, parse coverage, and error distances are populated from the frozen set, and the matching-only and multi-task adaptation results on this set appear in Appendix~\ref{app:r3r4}. Results from the protocol-analysis pool remain separate from benchmark scores.

\subsection{GRG Rubric Breakdown and Judge Sensitivity}

\begin{center}
  \begin{minipage}{\textwidth}
  \centering
  \captionof{table}{GRG rubric breakdown under the shared DeepSeek-V3.1 judge. Bootstrap confidence intervals for the composite overlap across all seven models. Cue identification remains at or below 1.245/2. The 84-pair human anchor (Section~\ref{sec:settings}) supports item-level rubric association under stratified sampling, not generator ranking or natural-distribution calibration.}
  \label{tab:rationale}
  \small
  \setlength{\tabcolsep}{4pt}
  \begin{tabular}{lcccc}
    \toprule
    \textbf{Model} & Cue / 2 & Link / 2 & Cons. / 1 & Total / 5 \\
    \midrule
    \rowcolor[gray]{0.96}\multicolumn{5}{l}{\textbf{\textit{Proprietary Models}}} \\
    GPT-5 & 1.245 & 1.065 & .465 & 2.775 \\
    Gemini 3.7 Flash & .906 & 1.131 & .618 & 2.654 \\
    \midrule
    \rowcolor[gray]{0.96}\multicolumn{5}{l}{\textbf{\textit{Open-weight Models}}} \\
    Qwen3.5-9B & .900 & 1.720 & .205 & 2.825 \\
    Qwen3.6-27B & .815 & 1.525 & .270 & 2.610 \\
    Qwen3.6-35B-A3B & .950 & 1.550 & .275 & 2.775 \\
    Qwen3.5-397B-A17B & .950 & 1.685 & .270 & 2.905 \\
    Kimi-K2.6 & 1.085 & 1.530 & .295 & 2.910 \\
    \bottomrule
  \end{tabular}
  \end{minipage}
\end{center}

The shared DeepSeek-V3.1 scores provide the only directly comparable GRG column across all seven generators. Cross-judge overlap is available for five hosted open-weight generators on the same 60 items per generator:

\begin{center}
  \begin{minipage}{\textwidth}
  \centering
  \captionof{table}{Gemini-versus-GPT-5 judge sensitivity on 60 shared items per generator. Item-score Pearson correlation shows positive linear association; mean absolute differences of 1.07--1.43 points on a five-point scale show substantial calibration sensitivity. The Gemini means are descriptive and are not merged into the shared DeepSeek comparison.}
  \label{tab:t3-judge-sensitivity}
  \small
  \setlength{\tabcolsep}{4pt}
  \begin{tabular}{lrrr}
    \toprule
    Generator & Gemini total / 5 & Pearson $r$ & Mean $|\Delta|$ \\
    \midrule
    Qwen3.5-9B & 3.533 & .790 & 1.350 \\
    Qwen3.6-27B & 3.567 & .764 & 1.433 \\
    Qwen3.6-35B-A3B & 3.433 & .854 & 1.217 \\
    Qwen3.5-397B-A17B & 3.550 & .767 & 1.400 \\
    Kimi-K2.6 & 3.433 & .792 & 1.067 \\
    \bottomrule
  \end{tabular}
  \end{minipage}
\end{center}

The shared DeepSeek judge yields overlapping marginal composite intervals and limited reference-cue recovery, while absolute scores vary across judge families. Overlap alone is not a paired test of generator differences. The 84-pair human anchor study (Section~\ref{sec:settings}) tests whether these scores predict human rubric judgments: three Prolific raters, none of whom authored the reference rationales, each graded all 84 items on the judge's rubric, blind to generator identity. The judge--human-mean Spearman correlation is $0.593$ (95\% bootstrap CI $[0.43, 0.72]$); judge agreement with individual raters ($\rho$ of $0.297$--$0.602$, mean $0.447$) brackets the raters' pairwise agreement with one another ($0.398$--$0.487$, mean $0.441$); and the per-dimension human-mean correlations are $0.605$ for cue, $0.438$ for link, and $0.439$ for faithfulness. The anchor supports item-level rubric association in the judge-stratified sample. Its composites sit $0.43$ points below the human mean on the five-point scale. Similar judge--individual and inter-rater mean correlations do not establish grader equivalence; this shared-reference study neither verifies all video facts nor establishes stable generator ranks or population-level calibration.

\section{Adaptation and Reproducibility Details}
\label{app:reproducibility}

\subsection{Notes to the Adaptation Table}
\label{app:adaptation-table-notes}

The main adaptation analysis (Table~\ref{tab:adaptation}) and Tables~\ref{tab:adaptation-original}--\ref{tab:r3r4-format} use the original 250 IP videos and 112 SRM items unless explicitly stated otherwise. Provenance-filtered subsets are reported separately in Appendix~\ref{app:split-audit}. A 24-frame cap on joint training rows, against the 64-frame evaluation budget, applies to every run. Unparsed base SRM responses (17/112) are scored incorrect. The seed row reports the mean and standard deviation of three independent training seeds of the same intensity-mixture configuration; per-seed $\Delta\rho_{0,\rm joint}$ is $+.112/+.080/+.052$ and $\Delta\rho_{0,\rm indep}$ is $+.120/+.108/+.084$, so the ``$p$'' entry on that row is a lower bound: no seed attains a smaller value. All other runs are single-seed; the multi-task run is evaluated at its best checkpoint (step 200, $\approx$2.2 of 5 epochs). Mean SRM error distance is 28.0 s for the base and 29.5 s for matching-only adaptation. Results under the alternative rule that excludes constant predictions before averaging appear later in this appendix; that rule reports significant gains for some collapsed runs.

\begin{table}[t]
  \centering
  \caption{\textbf{Original full-set adaptation results.} Before provenance filtering, every run is compared with the same base model served through the same local harness on the frozen 250-video IP set under the primary convention (constant predictions receive $\rho_0=0$), with paired bootstrap intervals and two-sided Wilcoxon tests. ``Constant'' counts joint predictions that assign one score to every moment of a video and therefore carry no ranking information. Training time-indexing errors and source overlap limit generalization and format attribution (Appendix~\ref{app:split-audit}). Frame caps, per-seed results, and the alternative constant-prediction scoring rule are documented in Appendix~\ref{app:adaptation-table-notes}.}
  \label{tab:adaptation-original}
  \scriptsize
  \setlength{\tabcolsep}{1.9pt}
  \renewcommand{\arraystretch}{1.1}
  \begin{tabular}{lrlrrlrrr}
    \toprule
    & \multicolumn{3}{c}{IP joint} & \multicolumn{3}{c}{IP independent} & Constant & SRM \\
    \cmidrule(lr){2-4}\cmidrule(lr){5-7}
    Training run & $\rho_0$ & $\Delta\rho_0$ [95\% CI] & $p$ & $\rho_0$ & $\Delta\rho_0$ [95\% CI] & $p$ & (joint) & Acc. \\
    \midrule
    Base (matched harness) & .024 & --- & --- & .253 & --- & --- & 0/250 & .304 \\
    \addlinespace[2pt]
    Intensity mixture, seed 0 & .135 & $+.112$ $[-.014,.233]$ & .171 & .373 & $+.120$ $[.017,.220]$ & .052 & 136/250 & .402 \\
    Intensity mixture, mean $\pm$ sd, seeds 0--2 & .105 & $+.081\pm.030$ & $\geq.171$ & .357 & $+.104\pm.018$ & $\geq.051$ & 152$\pm$36 & --- \\
    \addlinespace[2pt]
    Independent format only, seed 0 & .104 & $+.081$ $[-.009,.170]$ & .080 & .373 & $+.119$ $[.025,.216]$ & \textbf{.020} & \textbf{4/250} & --- \\
    Joint format only, seed 0 & .144 & $+.121$ $[.004,.240]$ & .078 & .211 & $-.042$ $[-.143,.062]$ & .401 & 114/250 & --- \\
    \addlinespace[2pt]
    Matching only, 3 ep. & .016 & $-.008$ $[-.030,.010]$ & .609 & .224 & $-.030$ $[-.081,.020]$ & .247 & \textbf{0/250} & .411 \\
    Multi-task mixture, 5 ep. & .157 & $+.134$ $[.021,.243]$ & \textbf{.030} & .393 & $+.140$ $[.047,.229]$ & \textbf{.008} & 153/250 & .402 \\
    \bottomrule
  \end{tabular}
\end{table}

\subsection{Split Construction and Supervision}

The original split excluded 480 clip IDs at construction and provided 1,364 training and 155 validation clips for intensity supervision. Training targets use the rater-debiased moment effect, mapped from 0--3 to an integer 0--100 score by rounding $100s/3$. The current frozen task files contain 462 distinct clip IDs across IP, SRM, and GRG. IP and GRG are clip-ID-disjoint from the available intensity training/validation rows; SRM is not, as detailed below. Clip-ID separation alone also does not establish original-upload separation.

\begin{center}
  \begin{minipage}{\textwidth}
  \centering
  \captionof{table}{Original intensity-adaptation split, by clip ID. ``N/A'' denotes an inapplicable field.}
  \label{tab:finetune-split}
  \small
  \begin{tabular}{lrrr}
    \toprule
    Split & Videos & Moments & Multi-moment videos \\
    \midrule
    Train & 1,364 & 1,952 & 428 \\
    Validation & 155 & 213 & 48 \\
    Excluded at split construction & 480 & N/A & N/A \\
    \bottomrule
  \end{tabular}
  \end{minipage}
\end{center}

The pilot intensity run combines 1,952 independent rows with 856 joint rows produced by oversampling the 428 multi-moment training videos twice, for 2,808 training rows. Validation contains 213 independent and 48 joint rows, for 261 total. All 3,069 rows passed file-existence and answer-format validation; none exceeded the 10,000-token skip threshold. Independent windows contain 2--4 frames (median 3). Joint training rows are capped at 24 frames, a mismatch from the 64-frame evaluation budget. This cap is unchanged in the later runs, which alter the prompt part order, the schedule, or the format composition but not the frame budget, so every adapted run reported here is trained on shorter joint sequences than it is evaluated on.

\subsection{Pilot-Run Model and Optimization}

\begin{center}
  \begin{minipage}{\textwidth}
  \centering
  \captionof{table}{Executed pilot-run configuration. These are measured run records, not the earlier planning estimates.}
  \label{tab:finetune-config}
  \small
  \begin{tabular}{p{0.22\textwidth}p{0.69\textwidth}}
    \toprule
    Component & Executed configuration \\
    \midrule
    Base model & Qwen/Qwen3.5-9B; revision\newline\texttt{c202236235762e1c871ad0c}\allowbreak\texttt{cb60c8ee5ba337b9a} \\
    LoRA~\citep{hu2022lora} & Rank 16, $\alpha=32$, dropout .05; language q/k/v/o and MLP gate/up/down projections where present; vision tower and merger frozen \\
    Trainable parameters & 29,097,984 of 9,438,911,728 (0.3083\%) \\
    Precision and memory & bfloat16; gradient checkpointing; distributed data parallel on eight GPUs \\
    Batch & Per-device batch 1; gradient accumulation 4; effective batch 32 \\
    Optimizer schedule & Peak learning rate $10^{-4}$; cosine schedule; 3\% warmup; assistant-token-only loss \\
    Duration & 3 epochs; 264 optimizer steps; 11,430 s (3.18 h) \\
    Checkpoint selection & Validation loss at steps 100/200/264: .46458/.40885/.39549; final step is best \\
    Hardware & 8 $\times$ NVIDIA RTX A6000, 49,140 MiB each; driver 575.57.08; CUDA 12.9 host \\
    Software & Python 3.13.13; PyTorch 2.11.0+cu128; torchvision 0.26.0+cu128; transformers 5.15.1; PEFT 0.20.0; accelerate 1.14.0; ffmpeg 4.4.2 \\
    \bottomrule
  \end{tabular}
  \end{minipage}
\end{center}

The adapted modules cover 128 language-model projections and no vision-tower tensors. The lower-than-planned parameter count follows Qwen3.5's hybrid attention structure; the target set was not widened after observing the architecture. Validation loss was still falling at the pilot run's final step. The intensity mixture later reached a similar minimum, 0.39611 versus 0.39549; this does not isolate the effect of the longer schedule. The pilot run is treated as a protocol-mismatch control because of its prompt part order, not because it stopped early.

\subsection{Pilot-Run Paired Tests and Output Collapse}

\begin{center}
  \begin{minipage}{\textwidth}
  \centering
  \captionof{table}{Pilot-run local-base versus LoRA paired tests. The joint success criterion of $\Delta\rho_0\geq.10$ at $p<.05$ was not met. The primary analysis assigns $\rho_0=0$ to constant predictions because the model runs produced very different rates of constant output.}
  \label{tab:finetune-paired}
  \small
  \begin{tabular*}{\textwidth}{@{\extracolsep{\fill}}llrrrrrr@{}}
    \toprule
    Convention & Task & $n$ & Base & LoRA & $\Delta$ & 95\% CI & Wilcoxon $p$ \\
    \midrule
    Excl.\ constants (alt.) & IP joint & 166 & .0239 & .1291 & .1052 & $[-.0216,.2309]$ & .131 \\
    Excl.\ constants (alt.) & IP indep. & 136 & .3418 & .5262 & .1844 & $[.0840,.2927]$ & .0031 \\
    Constants $=0$ (primary) & IP joint & 250 & .0236 & .0857 & .0621 & $[-.0422,.1655]$ & .320 \\
    Constants $=0$ (primary) & IP indep. & 250 & .2534 & .3423 & .0888 & $[-.0020,.1789]$ & .117 \\
    \bottomrule
  \end{tabular*}
  \end{minipage}
\end{center}

\begin{center}
  \begin{minipage}{\textwidth}
  \centering
  \captionof{table}{Pilot-run output-range collapse. When constant predictions are excluded, the adapted model's least informative profiles leave its own denominator; the primary convention keeps them as zero-correlation outcomes.}
  \label{tab:finetune-collapse}
  \small
  \begin{tabular}{llrrr}
    \toprule
    Task & Run & Constant & Prediction SD & Distinct values \\
    \midrule
    Joint & Base & 0/250 & 27.89 & 28 \\
    Joint & LoRA & 84/250 & 9.81 & 12 \\
    Independent & Base & 51/250 & 22.97 & 16 \\
    Independent & LoRA & 86/250 & 10.80 & 10 \\
    \bottomrule
  \end{tabular}
  \end{minipage}
\end{center}

The pilot run contracts its output range. Its all-video independent improvement is smaller than the conditional nonconstant improvement, and its mean-difference interval includes zero. Both the gain and uncertainty are retained; one seed and the input mismatches limit interpretation.

\subsection{Evaluation-Aligned Intensity Adaptation (Intensity Mixture)}

The intensity mixture of Table~\ref{tab:adaptation-original} is the pilot run with exactly two changes. First, training samples use the evaluation part order, that is, text prompt, then frames, then the short answer cue, so that a training string and the corresponding evaluation string differ only in the answer. The builder imports the prompt objects from the evaluation runners and asserts string equality. This reproduces the joint frame-time instruction even after training images have been subsampled; prompt equality does not establish correct time alignment. Second, the schedule is five epochs instead of three. Everything else---the same 2,808 training rows and 261 validation rows, the same cached frames, the same rank-16 adapter, the same optimizer settings, the same node and package versions---is held at the pilot run's values. Reordering the parts changed mean sequence length by at most one token, so the intervention is on order and not on input budget, and no row exceeded the 10,000-token skip threshold.

Validation loss falls to 0.39611 at step 300, roughly 3.4 epochs, then rises to 0.4920 and 0.4939 at steps 400 and 440 while training loss continues down to 0.2157. Because the trainer selects the best validation checkpoint, the evaluated adapter is the step-300 one, not the five-epoch endpoint. The pilot run's best validation loss was 0.39549 at its final step; this numerical proximity alone does not establish statistical equivalence or adequate training. Training took 5.27 hours on eight A6000s; the two seed replicates took 5.26 and 5.29 hours and reached their minima 100 steps earlier, at 0.42641 and 0.43449, so even the location of the optimum is seed-dependent. Every adapted run reported here produced zero unparsed responses on both IP protocols.

\begin{center}
  \begin{minipage}{\textwidth}
  \centering
  \captionof{table}{The intensity mixture across three training seeds, primary convention (constant predictions receive $\rho_0=0$), paired against the local base model on all 250 IP videos. The success criterion is $\Delta\rho_{0,\rm joint}\geq.10$ with $p<.05$. Seed~0 meets the effect-size criterion alone; the across-seed mean does not, and joint mean-difference intervals include zero for all seeds. Independent mean intervals exclude zero for seeds 0 and 1, while their Wilcoxon tests do not reject at .05.}
  \label{tab:r2-seeds}
  \small
  \setlength{\tabcolsep}{3pt}
  \begin{tabular*}{\textwidth}{@{\extracolsep{\fill}}lrlrrlr@{}}
    \toprule
    Run & Joint $\rho_0$ & $\Delta\rho_0$ [95\% CI] & $p$ & Indep.\ $\rho_0$ & $\Delta\rho_0$ [95\% CI] & $p$ \\
    \midrule
    Base (local) & .0236 & --- & --- & .2534 & --- & --- \\
    Seed 0 & .1352 & $+.1116$ $[-.0139,.2326]$ & .171 & .3730 & $+.1196$ $[.0168,.2202]$ & .0516 \\
    Seed 1 & .1033 & $+.0797$ $[-.0241,.1800]$ & .227 & .3615 & $+.1080$ $[.0119,.2031]$ & .0509 \\
    Seed 2 & .0753 & $+.0517$ $[-.0627,.1641]$ & .514 & .3373 & $+.0839$ $[-.0139,.1832]$ & .136 \\
    \midrule
    Mean $\pm$ sd & .1046 $\pm$ .0300 & $+.0810\pm.0300$ & --- & .3573 $\pm$ .0182 & $+.1038\pm.0182$ & --- \\
    \bottomrule
  \end{tabular*}
  \end{minipage}
\end{center}

\subsection{Format Ablation: Which Training Format Produces the Effect}

The format ablation holds every intensity-mixture setting fixed and changes only the format composition, splitting the mixture's own rows by their format field so the ablation runs see byte-identical samples. The independent-only run trains on 1,952 rows in 0.65 hours; the joint-only run trains on 856 oversampled rows in 2.35 hours. Both select an early-stopped checkpoint for the same reason as the mixture. Validation losses are not comparable across the two runs, because each is evaluated on its own format's validation subset, where predicting a bare integer and predicting a short JSON object are not equally hard; only the within-run trace is informative, and all three runs reach their validation minimum between 3.3 and 3.7 epochs. The joint-only run received only two validation evaluations, because the evaluation interval was held at the mixture's value against a much shorter run, so its checkpoint selection is coarser than the others'.

\begin{center}
  \begin{minipage}{\textwidth}
  \centering
  \captionof{table}{Format ablation, primary convention (constant predictions receive $\rho_0=0$), paired against the same local base model on all 250 videos. The mixture's gains are numerically close to joint-only training in joint scoring and independent-only training in independent scoring. These single-seed comparisons do not establish equivalence. Constant outputs and training time-index errors also affect interpretation.}
  \label{tab:r5-ablation}
  \small
  \setlength{\tabcolsep}{3pt}
  \begin{tabular*}{\textwidth}{@{\extracolsep{\fill}}lrlrrlr@{}}
    \toprule
    Training format & Joint $\rho_0$ & $\Delta\rho_0$ [95\% CI] & $p$ & Indep.\ $\rho_0$ & $\Delta\rho_0$ [95\% CI] & $p$ \\
    \midrule
    Base (local) & .0236 & --- & --- & .2534 & --- & --- \\
    Indep.\ + joint (seed 0) & .1352 & $+.1116$ $[-.0139,.2326]$ & .171 & .3730 & $+.1196$ $[.0168,.2202]$ & .0516 \\
    Independent only & .1044 & $+.0808$ $[-.0086,.1701]$ & .0801 & .3728 & $+.1193$ $[.0247,.2157]$ & \textbf{.0195} \\
    Joint only & .1441 & $+.1205$ $[.0041,.2397]$ & .0776 & .2114 & $-.0420$ $[-.1430,.0621]$ & .401 \\
    \bottomrule
  \end{tabular*}
  \end{minipage}
\end{center}

\begin{center}
  \begin{minipage}{\textwidth}
  \centering
  \captionof{table}{Output-range collapse under the alternative exclusion rule. ``Constant'' counts videos assigned one repeated value, which carry no ranking information; $n_{\rm nonconst}$ is the denominator after these videos are excluded, while each paired difference scores the base on the same nonconstant subset selected by the adapter; the base has 250 available joint predictions. Subset size and membership vary across runs, so conditional gains do not have a common target population.}
  \label{tab:adaptation-collapse}
  \small
  \begin{tabular*}{\textwidth}{@{\extracolsep{\fill}}lrrrrrr@{}}
    \toprule
    Run & \makecell[br]{Constant\\(joint)} & \makecell[br]{Pred.\\SD} & \makecell[br]{Distinct\\values} & $n_{\rm nonconst}$ & \makecell[br]{$\Delta\rho_{\rm joint}$\\(excl.)} & $p$ \\
    \midrule
    Base (local path) & 0/250 & 27.89 & 28 & 250 & --- & --- \\
    Pilot run (3 ep., misord.) & 84/250 & 9.81 & 12 & 166 & $+.1052$ & .131 \\
    Mixture seed 0 & 136/250 & 13.26 & 13 & 114 & $+.2005$ & \textbf{.0348} \\
    Mixture seed 1 & 126/250 & 10.65 & 15 & 124 & $+.0944$ & .155 \\
    Mixture seed 2 & 193/250 & 11.94 & 16 & 57 & $+.2350$ & .125 \\
    Independent only & \textbf{4/250} & 13.75 & 26 & 246 & $+.0745$ & .108 \\
    Joint only & 114/250 & 10.99 & 20 & 136 & $+.1136$ & .175 \\
    \bottomrule
  \end{tabular*}
  \end{minipage}
\end{center}

The independent-only adapter preserves nearly all joint distinctions (4/250 constant predictions), whereas joint-only training produces 114/250 constant joint profiles. These observations associate format with output behavior, but the joint input's time-indexing defect and differences in training-row counts also vary across the formats. For mixture seed 0, the conditional comparison uses the same 114 videos for both models: $\rho$ increases from 0.0959 to 0.2964 ($\Delta=0.2005$, $p=0.0348$). It is not a comparison between a 114-video adapted mean and a 250-video base mean. Keeping constants at zero instead reports the all-video $\Delta\rho_0=0.1116$. Neither statistic should silently replace the other: one conditions on the adapter producing a ranking, while the other includes its non-discriminating outputs.

\begin{center}
  \begin{minipage}{\textwidth}
  \centering
  \captionof{table}{The elicitation gap after adaptation. Every adapted run except joint-only leaves the independent-minus-joint gap at or above the base model's, and all of those remain highly significant. Joint-only both increases joint scoring and decreases independent scoring, so a smaller protocol difference does not on its own establish improved ranking in both settings.}
  \label{tab:adaptation-gap}
  \small
  \setlength{\tabcolsep}{6pt}
  \begin{tabular}{lrrrl}
    \toprule
    Run & Joint $\rho_0$ & Indep.\ $\rho_0$ & Gap & Wilcoxon $p$ \\
    \midrule
    Base (local path) & .0236 & .2534 & .2298 & .0012 \\
    Mixture seed 0 & .1352 & .3730 & .2379 & $1.6\times10^{-5}$ \\
    Mixture seed 1 & .1033 & .3615 & .2582 & $6.0\times10^{-6}$ \\
    Mixture seed 2 & .0753 & .3373 & .2620 & $<1\times10^{-6}$ \\
    Independent only & .1044 & .3728 & .2684 & $5.4\times10^{-5}$ \\
    Joint only & .1441 & .2114 & .0673 & .286 \\
    \bottomrule
  \end{tabular}
  \end{minipage}
\end{center}

The independent-only run combines 246/250 nonconstant joint profiles with positive point-estimate changes in both settings. Its independent improvement has a positive mean-difference interval. Multi-task adaptation also improves the constant-inclusive metrics, so independent-only is not the sole run with an all-video gain.

\subsection{Matching-Only and Multi-Task Adaptation}
\label{app:r3r4}

\paragraph{Data and Disjointness.}
The matching-only run uses 93 training rows and 10 validation rows; the multi-task run combines these with intensity rows for 2,901 training and 271 validation rows. Matching targets are automatic temporal-alignment labels. The matching rows themselves have zero candidate- and clip-ID overlap with the frozen SRM tasks. However, intensity rows in the multi-task and intensity-only data share six SRM clip IDs and seven exact SRM candidate IDs; one intensity-validation clip shares one further candidate. Thus the earlier matching-only disjointness check did not establish separation for the full multi-task input. Appendix~\ref{app:split-audit} reports complete cross-task checks.

\paragraph{Matching Gains Concentrate on Previously Unparsed Items.}
On the 112-item SRM set, the local base scores .304 with 17 unparsed responses; matching-only adaptation scores .411 with none (McNemar $p=.036$). On the 95 base-parseable items the increase is only .011 ($p=1.000$). Intensity-only adaptation reaches .402 and also removes parse failures, although its training clips overlap part of SRM. All 17 unparsed base outputs consume the 48-token generation budget and contain no standalone A--D choice, so explicit-letter extraction recovers none without consulting the gold. Gains on these items are consistent with format stabilization, but the saved prose does not establish a previously correct decision. Semantic mapping to the option media or a constrained-decoding baseline remains a separate test. Mean error distance is 28.0 s for the base and 29.5 s for matching-only adaptation, measured on different error subsets; this comparison cannot establish unchanged localization on matched errors. The option distribution is also imbalanced: always selecting D scores .321, compared with uniform-choice chance .250.

\begin{center}
  \begin{minipage}{\textwidth}
  \centering
  \captionof{table}{SRM parse-coverage diagnostic on the frozen 112-item set. All comparisons are against the same local-harness base. The parsed subset restricts to the 95 items the base answers with a parseable letter; McNemar $p$ is exact.}
  \label{tab:r3r4-format}
  \small
  \setlength{\tabcolsep}{4.5pt}
  \begin{tabular}{llrrrrr}
    \toprule
    Run & Match supervision & Acc.\ (112) & Unparsed & Parsed-subset acc.\ (95) & $\Delta$ & $p$ \\
    \midrule
    Base & --- & .304 & 17 & .358 & --- & --- \\
    Intensity mixture & none & .402 & 0 & .368 & $+.011$ & 1.000 \\
    Matching only & 93 rows & .411 & 0 & .368 & $+.011$ & 1.000 \\
    Multi-task & 93 rows & .402 & 0 & .358 & $\pm.000$ & 1.000 \\
    \bottomrule
  \end{tabular}
  \end{minipage}
\end{center}

\paragraph{Multi-Task Improvements and Output Collapse.}
The multi-task adapter differs from intensity-mixture seed 0 by $+.022$ joint and $+.020$ independent (Wilcoxon $p=.69$ and $.64$), and from matching-only SRM by $-.009$ ($p=1.000$). These tests do not establish equivalence or absence of interference. The evaluated checkpoint is step 200 of 455, selected by validation loss. Its all-video joint improvement of $+.134$ contains $+.030$ from 153 constant-output videos, where base mean $\rho_0=-.049$ becomes zero, and about $+.104$ from the other 97 videos, where the paired mean rises from .138 to .405. This gain is conditional on those 97 nonconstant profiles. For independent scoring, constant-output videos contribute $-.015$, while the nonconstant subset improves from .328 to .540. Matching-only training produces no constant joint profiles; the format-dependent pattern is observed for this model and recipe, with the timestamp defect limiting causal attribution to scalar supervision.

\subsection{Matched Human Comparisons after Adaptation}

Table~\ref{tab:adaptation-human} applies the matched LOO pairwise analysis to all 250 IP videos, using 956 rater--video units. The individual-rater reference is 0.607. Multi-task pairwise accuracy is 0.573 joint and 0.692 independent; model-minus-human intervals are $[-0.070,0.002]$ and $[0.040,0.128]$, respectively. These pointwise comparisons measure agreement with the fixed annotator pool under different scoring tasks and input information, not distance to a human capability ceiling. The training timestamp defect continues to limit mechanism claims.

% Generated by scripts/export_review_tables.py from validation/revision-2026-09-17/reanalysis.json.
\begin{center}
\begin{minipage}{\textwidth}
\centering
\captionof{table}{Adaptation under matched LOO pairwise accuracy on all 250 IP videos (956 rater--video units). The video-weighted individual-rater reference is 0.607. $\Delta_H$ is model minus human with pointwise video-bootstrap intervals; $\Delta_B$ is model minus local base. This reference is agreement with a fixed rater pool, not a human capability ceiling.}
\label{tab:adaptation-human}
\small
\setlength{\tabcolsep}{3pt}
\begin{tabular*}{\textwidth}{@{\extracolsep{\fill}}lrrrrr@{}}
\toprule
Run & Mode & PA & $\Delta_H$ & 95\% CI for $\Delta_H$ & $\Delta_B$ \\
\midrule
Local base & J & 0.511 & -0.096 & $[-0.153,-0.039]$ & --- \\
Local base & I & 0.624 & 0.018 & $[-0.031,0.067]$ & --- \\
Intensity mixture, seed 0 & J & 0.563 & -0.043 & $[-0.083,-0.002]$ & 0.053 \\
Intensity mixture, seed 0 & I & 0.681 & 0.074 & $[0.031,0.116]$ & 0.057 \\
Intensity mixture, seed 1 & J & 0.553 & -0.053 & $[-0.094,-0.013]$ & 0.043 \\
Intensity mixture, seed 1 & I & 0.675 & 0.068 & $[0.028,0.108]$ & 0.051 \\
Intensity mixture, seed 2 & J & 0.535 & -0.072 & $[-0.101,-0.041]$ & 0.024 \\
Intensity mixture, seed 2 & I & 0.661 & 0.054 & $[0.013,0.096]$ & 0.037 \\
Matching only & J & 0.507 & -0.100 & $[-0.157,-0.041]$ & -0.004 \\
Matching only & I & 0.611 & 0.004 & $[-0.043,0.052]$ & -0.014 \\
Multi-task & J & 0.573 & -0.034 & $[-0.070,0.002]$ & 0.062 \\
Multi-task & I & 0.692 & 0.085 & $[0.040,0.128]$ & 0.067 \\
Independent only & J & 0.553 & -0.054 & $[-0.111,0.002]$ & 0.042 \\
Independent only & I & 0.680 & 0.073 & $[0.030,0.115]$ & 0.056 \\
Joint only & J & 0.570 & -0.036 & $[-0.075,0.004]$ & 0.060 \\
Joint only & I & 0.603 & -0.003 & $[-0.039,0.033]$ & -0.021 \\
\bottomrule
\end{tabular*}
\end{minipage}
\end{center}

\subsection{Cross-Task Split and Source-Lineage Audit}
\label{app:split-audit}

The local frozen SRM task file is byte-identical to the evaluation Git snapshot. Cross-checking every intensity and matching training row against it finds six overlapping training clip IDs, corresponding to seven SRM items, plus one validation clip corresponding to one further SRM item. The matching-only rows remain clip-ID-disjoint; the overlaps arise in intensity supervision used by other adapters. IP and GRG have no direct clip-ID overlap. The original 480-ID exclusion list therefore cannot be described as excluding every current evaluation clip.

A second check follows 21,791 recorded child-to-parent recut relationships. Against the union of available training and validation rows, 31 of 250 IP videos, 20 of 97 SRM clips, and 16 of 185 GRG clips share a parent clip. Mapping these parents to the original segment-file paths gives the same intersections. However, a parent clip can itself be one segment of a longer recording. The zero-shot models were not trained in this adaptation study; the audit limits the adapted-model generalization claims.

Removing the 31 IP videos with shared parent clips leaves 219 videos. Table~\ref{tab:adaptation-parent-excluded} reports paired differences on this sensitivity subset. Mean gains keep their sign and magnitude: multi-task intervals exclude zero in both modes, as do the independent intervals of independent-only and mixture seeds 0 and 1 and the joint interval of joint-only. Selection occurred after the original runs and uses provenance alone. These subsets are parent-clip-disjoint, but remain overlapping at the source-recording level and retain the training timestamp defect.

A broader audit groups original segment paths by their source-recording directory. The processing scripts create these directories from an input recording's stem and place its segments, including separate person crops, within them. Every audited clip resolves through the local mapping. At this level, 229/250 IP videos, 106/112 SRM items (91/97 clips), and 157/200 GRG items (146/185 clips) share a development source. Excluding them leaves only 21 IP videos, 6 SRM items, and 43 GRG items; these small residual sets are not used to establish unseen-recording adaptation performance. We therefore do not claim adaptation performance on unseen recordings.

% Generated by scripts/export_review_tables.py from validation/revision-2026-09-17/reanalysis.json.
\begin{center}
\begin{minipage}{\textwidth}
\centering
\captionof{table}{Adapted minus local-base $\rho_0$ on 219 IP videos after excluding all 31 whose recorded parent clip appears in training or validation. Intervals resample paired videos. Shared source recordings and the training timestamp defect remain; this exclusion does not establish generalization to unseen recordings.}
\label{tab:adaptation-parent-excluded}
\small
\setlength{\tabcolsep}{3pt}
\begin{tabular*}{\textwidth}{@{\extracolsep{\fill}}lrrrr@{}}
\toprule
Run & Joint $\Delta\rho_0$ & 95\% CI & Indep. $\Delta\rho_0$ & 95\% CI \\
\midrule
Intensity mixture, seed 0 & 0.123 & $[-0.003,0.252]$ & 0.142 & $[0.033,0.246]$ \\
Intensity mixture, seed 1 & 0.099 & $[-0.006,0.204]$ & 0.132 & $[0.034,0.230]$ \\
Intensity mixture, seed 2 & 0.067 & $[-0.055,0.192]$ & 0.090 & $[-0.013,0.192]$ \\
Matching only & -0.009 & $[-0.033,0.012]$ & -0.025 & $[-0.077,0.027]$ \\
Multi-task & 0.133 & $[0.018,0.252]$ & 0.159 & $[0.062,0.256]$ \\
Independent only & 0.077 & $[-0.018,0.174]$ & 0.137 & $[0.039,0.234]$ \\
Joint only & 0.140 & $[0.011,0.265]$ & -0.041 & $[-0.151,0.067]$ \\
\bottomrule
\end{tabular*}
\end{minipage}
\end{center}

For SRM, direct clip-ID exclusion removes eight items from seven clips, leaving 104 items on 90 clips. Excluding every known development parent removes 21 items from 20 clips, leaving 91 items on 77 clips. All adapters and the local base are compared on the same retained items; neither subset was selected using model scores. Table~\ref{tab:srm-provenance} reports both sensitivity analyses alongside the main table's original 112-item evaluation.

% Generated by scripts/export_review_tables.py from validation/revision-2026-09-17/reanalysis.json.
\begin{center}
\begin{minipage}{\textwidth}
\centering
\captionof{table}{SRM re-scoring after excluding training and validation overlap. Clip-ID filtering removes 8 items from 7 clips; recorded-parent filtering removes 21 items from 20 clips. Parsed $\Delta$ is adapted minus base accuracy on the base-parseable subset (89 items for Clip ID, 78 for Parent). Parsed-subset McNemar $p=0.804$ for the intensity mixture under parent filtering and 1.000 otherwise. Full-set $p$ compares each adapter with the base on the stated subset.}
\label{tab:srm-provenance}
\small
\setlength{\tabcolsep}{3pt}
\begin{tabular*}{\textwidth}{@{\extracolsep{\fill}}lrrrrrr@{}}
\toprule
Run & Exclusion & $n$ & Acc. & Unparsed & Parsed $\Delta$ & Full $p$ \\
\midrule
Local base & Clip ID & 104 & 0.298 & 15 & 0.000 & --- \\
Local base & Parent & 91 & 0.319 & 13 & 0.000 & --- \\
Intensity mixture, seed 0 & Clip ID & 104 & 0.394 & 0 & 0.011 & 0.087 \\
Intensity mixture, seed 0 & Parent & 91 & 0.429 & 0 & 0.026 & 0.064 \\
Matching only & Clip ID & 104 & 0.394 & 0 & 0.000 & 0.064 \\
Matching only & Parent & 91 & 0.407 & 0 & 0.000 & 0.115 \\
Multi-task & Clip ID & 104 & 0.385 & 0 & -0.011 & 0.150 \\
Multi-task & Parent & 91 & 0.407 & 0 & 0.000 & 0.169 \\
\bottomrule
\end{tabular*}
\end{minipage}
\end{center}

\subsection{Aligned Follow-Up Runs}

\begin{center}
  \begin{minipage}{\textwidth}
  \centering
  \captionof{table}{Original 250-IP/112-SRM adaptation runs and outcomes.}
  \label{tab:followup-runs}
  \small
  \begin{tabular}{p{0.11\textwidth}p{0.26\textwidth}p{0.25\textwidth}p{0.26\textwidth}}
    \toprule
    Run & Supervision and format & Required evaluation & Current status \\
    \midrule
    Intensity mixture & Independent + joint intensity; evaluation-aligned part order; 5 epochs & IP joint and independent, paired base/LoRA under the primary convention; range diagnostics; seeds and compute & Complete; three seeds (Table~\ref{tab:r2-seeds}). Criterion not met. \\
    Format ablation & Single-format ablations of the mixture: independent-only and joint-only & Same evaluation as the mixture, plus per-format collapse attribution & Complete; single seed each (Table~\ref{tab:r5-ablation}). \\
    Matching only & Matching rows only, in the matching prompt format; temporal-alignment labels & Frozen SRM base/LoRA accuracy, Wilson intervals, McNemar test, parse coverage & Complete; single seed (Section~\ref{app:r3r4}). Full-set $+.107$ ($p=.036$); base-parseable-subset $\Delta=+.011$, $p=1.000$. \\
    Multi-task & Mixture of independent intensity, joint intensity, and matching & IP + frozen SRM; per-task changes and interference/collapse diagnostics & Complete; single seed, best checkpoint at $\approx$2.2 epochs (Section~\ref{app:r3r4}). No detected difference from single-task; not an equivalence test. Constants: 153/250. \\
    \bottomrule
  \end{tabular}
  \end{minipage}
\end{center}

Measured compute for the completed runs, all on eight RTX A6000s: the intensity mixture's seed~0 took 5.27 hours of training plus 40 minutes of evaluation; the two seed replicates took 5.26 and 5.29 hours plus 40 minutes of evaluation; and the two format ablations took 0.65 and 2.35 hours plus 40 minutes, for roughly 20.8 hours of wall-clock time across the five adapted runs, or about 166 GPU-hours on the eight-GPU node. The matching-only run took 430 seconds of training (9 optimiser steps on 93 rows) plus 517 seconds to evaluate both SRM runs, and the multi-task run took 5.38 hours of training (455 steps) plus 36 minutes of evaluation; the off-diagonal control evaluations (the intensity mixture on SRM, matching-only on both IP protocols) add roughly one hour, for roughly 28 hours of wall-clock time across the seven adapted runs and controls, or about 224 GPU-hours.

\subsection{Evaluation Cost Ledger}

\begin{center}
  \begin{minipage}{\textwidth}
  \centering
  \captionof{table}{Current evaluation ledger aggregated across completed calls. Dollar values follow the repository's provider accounting or public-list estimates. Tinker calls were grant-billed, and ``0.000'' for local inference means no API charge, not zero energy or hardware cost.}
  \label{tab:evaluation-cost}
  \small
  \setlength{\tabcolsep}{4.5pt}
  \begin{tabular}{llrrrr}
    \toprule
    Serving path & Model / role & Calls & Input tokens & Output tokens & Accounted cost (USD) \\
    \midrule
    Vertex AI & Gemini 3.7 Flash & 3,992 & 59,470,696 & 32,798 & 44.726 \\
    OpenAI API & GPT-5 & 3,033 & 2,583,488 & 61,332 & 3.843 \\
    Tinker & Qwen3.5-9B & 1,887 & 7,307,733 & 21,247 & 4.865 \\
    Tinker & Qwen3.6-27B & 1,842 & 7,107,869 & 16,867 & 13.315 \\
    Tinker & Qwen3.6-35B-A3B & 1,846 & 7,121,241 & 16,964 & 3.868 \\
    Tinker & Qwen3.5-397B-A17B & 1,937 & 7,378,437 & 19,588 & 22.282 \\
    Tinker & Kimi-K2.6 & 1,842 & 10,044,031 & 20,534 & 22.260 \\
    Tinker & DeepSeek-V3.1 judge & 1,391 & 383,651 & 27,820 & 0.768 \\
    Local & Qwen3.5-9B base & 892 & 4,440,405 & 9,096 & 0.000 \\
    Local & Qwen3.5-9B LoRA & 892 & 4,440,405 & 9,096 & 0.000 \\
    \bottomrule
  \end{tabular}
  \end{minipage}
\end{center}

The ledger totals \$44.726 for Gemini, \$3.843 for OpenAI, and \$66.590 across the priced Tinker generators. The legacy DeepSeek ledger omitted a rate; its 383,651 input and 27,820 output tokens imply an additional \$0.768 at the public uncached rates checked on 18 September 2026, rather than zero-cost judging. Local inference has no API charge. The local rows predate the frozen-SRM runs and the intensity-mixture, format-ablation, seed-replicate, matching-only, and multi-task runs, which add hosted-provider calls on the frozen matching set and roughly 224 GPU-hours of local training and evaluation (28 hours of wall-clock time on eight GPUs) at no API charge.

The subsequent SRM reaction-input controls and five-video Qwen timestamp rerun add 338 successful logged calls, estimated at \$0.85 using the recorded rates. This excludes any cancelled in-flight request whose usage was not returned. The subsequent GRG controls, three-condition IP study, and additional timestamp checks add 2,361 successful calls, estimated at \$5.614 (\$1.077 OpenAI and \$4.537 Tinker). These figures exclude requests with no returned usage and remain separate from the historical ledger above.

\subsection{Reproducibility and Release Checklist}

The evaluation repository exposes separate executable entry points for joint and independent ranking, joint matching, the per-option and side-by-side matching settings, rationale generation, rationale judging, scoring, summarization, and cost accounting. Task JSON files preserve canonical item identifiers, source keys, time windows, option order, human references, and consensus fields. Randomized set construction and bootstrap analysis use recorded seeds.

The reported benchmark and follow-up comparisons use a fixed set of task artifacts with canonical item identities. Source-group reconstruction, coverage checks, and all paired comparisons are keyed to these identities.

% Table~\ref{tab:task-versions} identifies the frozen task artifacts used for the reported benchmark and follow-up comparisons. Source-group reconstruction, coverage checks, and all paired comparisons use these same task identities.

% \input{tables/task_versions}

% The source videos cannot be assumed redistributable merely because the originating videos were public.

\section{Limitations}
\label{sec:limitations}

\paragraph{Population and Content Coverage.}
The test population consists of selected online creators. IP contains seven creators from five channels and is dominated by one acquisition source; model-assisted nomination also selects for strong reactions. The source collection's size and domain range do not establish corresponding test coverage. Familiarity opt-outs in GRG further show that interpretation depends on content knowledge.

\paragraph{Measurement Validity.}
Human and model IP references now use matched targets and weights, but differ in response resolution and available audiovisual information. GRG references were authored with stimulus audio, whereas evaluated generators see sparse frames. SRM labels are fixed by temporal alignment and measure recovery of the aligned window rather than a causal attribution judgment. In the GPT-5 input controls, SRM and GRG scores respond to the presence of a reaction but not to its identity, so scores can mix stimulus recognizability with reaction understanding; the text-only judge measures reference consistency rather than complete video factuality.

\paragraph{Privacy, Copyright, and Research Oversight.}
The source videos are publicly posted reaction recordings, but public availability does not eliminate privacy, dignity, copyright, or redistribution concerns. The release pseudonymizes identifying metadata, documents takedown procedures, does not redistribute copyrighted stimulus video, and provides derived annotations. The human annotation work includes research interactions with Prolific participants. The benchmark is intended for evaluating research systems, not for inferring private mental states, diagnosing individuals, or ranking people by emotional expressiveness.

\section{Ethics, Data Governance, and Intended Use}
\label{app:ethics}

The benchmark's human evidence comes from the intensity raters and Prolific rationale annotators reported above. The Prolific studies paid an average of \$16.67 per hour for intensity annotation (median completion time 4.5 hours), \$12.00 per hour for rationale collection (median 1.5 hours), and \$18 per rater (\$14.40 per hour, median 75 minutes) for the judge-anchor study. Institutional ethics review for the human-participant research has been completed. The release pseudonymizes participant and creator identifiers and separates private annotation metadata from benchmark labels.

The source videos are publicly posted, but public availability does not imply permission for unrestricted redistribution. The release distributes derived annotations under CC BY-NC 4.0, refers to video by key without redistributing it, and provides a creator opt-out and takedown process.

\rb{} is intended to evaluate research models that connect observable events with observable reactions under a disclosed protocol. It is not designed to infer private mental states, diagnose emotion or health, determine deception, rank people by emotional expressiveness, or support surveillance or consequential decisions about individuals.

\end{document}